\documentclass[twoside,11pt]{article}
\usepackage{jmlr2e}

\usepackage{amsmath, amssymb} %, amsthm}
\usepackage{graphicx, graphics, epsfig, color}
\usepackage{adjustbox}
\usepackage{subfigure}
\usepackage[utf8]{inputenc}
\usepackage{tikz, pgfplots}
\usepgfplotslibrary{colormaps,fillbetween}
\usepackage{algorithm, algorithmic}
\usepackage{float}
\usepackage{multirow}
\usepackage{cuted, flushend}
\usepackage{midfloat}
\usepackage{bm}
\usepackage{fancyhdr}
\usepackage{enumerate}
\usepackage{siunitx}
\pgfplotsset{compat=1.4}
\usetikzlibrary{backgrounds,spy}
\pgfdeclarelayer{background}
\pgfsetlayers{background,main}

\tikzset{new spy style/.style={spy scope={%
        magnification=5,
        size=1.25cm,
        connect spies,
        every spy on node/.style={
                rectangle,
                draw,
        },
        every spy in node/.style={
                draw,
                rectangle,
                fill=gray!40,
        }
}}}

\usepackage{tabularx}
\makeatletter
\def\hlinewd#1{%
	\noalign{\ifnum0=`}\fi\hrule \@height #1 %
	\futurelet\reserved@a\@xhline}
\makeatother

\newcounter{cassumption}
\newtheorem{assumption}[cassumption]{Assumption}
\def\E{\mathbb{E}}
\def\cov{{\rm Cov}}

\def\RR{\mathbb{R}}

\def\trans{{\sf T}}

\def\asto{ {\overset{\rm a.s.}{\longrightarrow}} }

\def\muctra{\mu_a^{\circ}}
\def\muctrb{\mu_b^{\circ}}

\DeclareMathOperator{\tr}{{\rm tr}}

\newcommand{\RED}{\color[rgb]{0,0,0}}
\newcommand{\BLUE}{\color[rgb]{0,0,0}}

\author{Xiaoyi Mai, Romain Couillet}

\jmlrheading{XX}{2019}{X-X}{XX/XX; Revised XX/XX}{XX/XX}{Mai, Couillet}

\ShortHeadings{High Dimensional Semi-Supervised Graph Regularization}{Mai, Couillet}

\firstpageno{1}

\begin{document}

\title{Consistent Semi-Supervised Graph Regularization\\ for High Dimensional Data}

\author{\name Xiaoyi Mai$^{1}$ \email xiaoyi.mai@l2s.centralesupelec.fr \\ \name Romain Couillet$^{1,2}$ \email romain.couillet@centralesupelec.fr \\
\addr $^1$CentraleSup\'elec, Laboratoire des Signaux et Syst\`emes \\
Universit\'e Paris-Saclay\\
3 rue Joliot Curie, 91192 Gif-Sur-Yvette \\
 \\
$^2$GIPSA-lab, GSTATS DataScience Chair \\
Universit\'e Grenoble--Alpes \\
11 rue des Math\'ematiques, 38400 St Martin d'H\`eres.
}

\editor{XX XX}

\maketitle

%\date{\today}
%
\begin{abstract}
Semi-supervised Laplacian regularization, a standard graph-based approach for learning from both labelled and unlabelled data, was recently demonstrated to have an insignificant high dimensional learning efficiency with respect to unlabelled data \citep{mai2018random}, causing it to be outperformed by its unsupervised counterpart, spectral clustering, given sufficient unlabelled data. Following a detailed discussion on the origin of this inconsistency problem, a novel regularization approach involving centering operation is proposed as solution, supported by both theoretical analysis and empirical results.
\end{abstract}

\section{Introduction}
\label{sec:intro}
Machine learning methods aim to form a mapping from an input data space to an output characterization space (classification labels, regression vectors) by optimally exploiting the information contained in the collected data. Depending on whether the data fed into the learning model are \textit{labelled} or \textit{unlabelled}, the machine learning algorithms are respectively broadly categorized as \textit{supervised} or \textit{unsupervised}. Although the supervised approach has by now occupied a dominant place in real world applications thanks to its high-level accuracy, the cost of labelling process, overly high in comparison to the collection of data, continually compels researchers to develop techniques using unlabelled data with growing interest, as many popular learning tasks of these days, such as image classification, speech recognition and language translation, require enormous training datasets to achieve satisfying results. 

The idea of semi-supervised learning (SSL) comes from the expectation of maximizing the learning performance by combining labelled and unlabelled data \citep{chapelle2009semi}. It is of significant practical value when the cost of supervised learning is too high and the performances of unsupervised approaches is too weak. Despite its natural idea, semi-supervised learning has not reached broad recognition. As a matter of fact, many standard semi-supervised learning techniques were found to be unable to learn effectively from unlabelled data \citep{shahshahani1994effect,cozman2002unlabeled,ben2008does}, thereby hindering the interest for these methods.

\medskip

%%% Ici, faire un petit point historique: les bases du SSL avec le laplacien, puis les méthodes correctives (high order normalization, etc).

\medskip

A first key reason for the underperformance of semi-supervised learning methods lies in the lack of understanding of such approaches, caused by the technical difficulty of a theoretical analysis. Indeed, even the simplest problem formulations, the solutions of which assume an explicit form, involve complicated-to-analyze mathematical objects (such as the resolvent of kernel matrices).

A second important aspect has to do with dimensionality. As most semi-supervised learning techniques are built upon low-dimensional reasonings, they suffer the transition to large dimensional datasets. Indeed, it has been long noticed that learning from data of intrinsically high dimensionality presents some unique problems, for which the term \textit{curse of dimensionality} was coined. One important phenomenon of the curse of dimensionality is known as \textit{distance concentration}, which is the tendency for the distances between high dimensional data vectors to become indistinguishable. This problem has been studied in many works \citep{beyer1999nearest,aggarwal2001surprising,hinneburg2000nearest,francois2007concentration,JMLR:v18:17-151}, providing mathematical characterization of the distance concentration under the conditions of intrinsically high dimensional data. 

Since the strong agreement between geometric proximity and data affinity in low dimensional spaces is the foundation of similarity-based learning techniques, it is then questionable whether these traditional techniques will perform effectively on high dimensional data sets, and many counterintuitive phenomena may occur.

\medskip

{\BLUE The aforementioned tractability and dimensionality difficulties can be tackled at once by exploiting recent advances in random matrix theory to analyze the performance of semi-supervised algorithms. With their weakness understood, it is then possible to propose fundamental corrections for these algorithms. The present article specifically focuses on \textit{semi-supervised graph regularization} approaches \citep{belkin2003using,zhu2003semi,zhou2004learning}, a major subset of semi-supervised learning methods \citep{chapelle2009semi}, often referred to as Laplacian regularizations with their loss functions involving differently normalized Laplacian matrices \citep{avrachenkov2012generalized}.  These semi-supervised learning algorithms of Laplacian regularization are presented in Subsection~\ref{sec:laplacian algorithms}. It was made clear in a recent work of \citet{mai2018random} that among existing Laplacian regularization algorithms, only one (related to the PageRank algorithm) yields reasonable classification results, yet with asymptotically negligible contribution from the unlabelled dataset. This last observation of the inefficiency of Laplacian regularization methods to learn from unlabelled data may cause them to be outperformed by a mere (unsupervised) spectral clustering approach \citep{von2007tutorial} in the same high dimensional settings \citep{COU16}. We refer to Subsection~\ref{sec:previous results} for a summary of the key mathematical results in the previous analysis of \citet{mai2018random}, which motivate the present work.

\medskip

The contributions of the present work start from Section~\ref{sec:regularization with centered similarities}: with the cause for the unlabelled data learning inefficiency of Laplacian regularization identified in Subsection~\ref{sec:problem identification}, a new regularization approach with centered similarities is proposed in Subsection~\ref{sec:proposition of the method} as a cure, followed by a subsection justifying the proposed method from the alternative viewpoint of label propagation. This new regularization method is simple to implement and its effectiveness supported by a rigorous analysis, in addition to heuristic arguments and empirical results which justify its usage in more general data settings. Specifically, the statistical analysis of Section~\ref{sec:analysis}, placed under a high dimensional Gaussian mixture model (as employed in the previous analysis of \citet{mai2018random}, as well as that of \citet{COU16} in the context of spectral clustering), proves the consistency of our proposed high dimensional semi-supervised learning method, with guaranteed performance gains over Laplacian regularization. The theoretical results of Section~\ref{sec:analysis} are validated by simulations in Subsection~\ref{sec:validation}. Broadening the perspective, the discussion in Subsection~\ref{sec:problem identification} suggests that the unlabelled data learning inefficiency of Laplacian regularization is due to the universal distance concentration phenomenon of high dimensional data. The advantage of our centered regularization, proposed as a countermeasure to the problem of distance concentration, should extend beyond the analyzed Gaussian mixture model. This claim is verified in Subsection~\ref{sec:beyond the model} through experimentation on real-world datasets, where we observe that the proposed method tends to produce more marked performance gains over the Laplacian approach when the distance concentration phenomenon is more severe. The discussion is  extended in Section~\ref{sec:further discussion} to include some related graph-based SSL methods. Although not suffering from the unlabelled data learning inefficiency problem like Laplacian regularization, these methods may still have a suboptimal semi-supervised learning performance on high dimensional data as they do not possess the same performance guarantees as our proposed method. This claim is verified in Subsection~\ref{sec:isotropic data} thanks to a recent work of \cite{lelarge2019asymptotic} characterizing the optimal performance on isotropic Gaussian data, upon which the graph-based SSL methods, except the proposed centered regularization approach, are found to yield unsatisfying results. We approach the subject of learning on sparse graphs in Subsection~\ref{sec:sparse graphs}, where the benefit of using the proposed method is justified in terms of computational efficiency and learning performance.
}

\medskip

\textit{Notations:} $1_{n}$ is the column vector of ones of size $n$, $I_{n}$ the $n\times n$ identity matrix. The norm $\|\cdot\|$ is the Euclidean norm for vectors and the operator norm for matrices. We follow the convention to use $o_P(1)$ for a sequence of random variables that convergences to zero in probability. For a random variable $x\equiv x_n$ and $u_n\ge 0$, we write $x=O(u_n)$ if for any $\eta>0$ and $D>0$, we have $n^D{\rm P}(x\ge n^\eta u_n)\to 0$.

\section{Background}
\label{sec:backgroud}

We will begin this section by recalling the basics of graph learning methods, before briefly reviewing the mains results of \citep{mai2018random}, which motivates the proposition of our centered regularization method presented in the subsequent section.

\subsection{Laplacian Regularization Method}
\label{sec:laplacian algorithms}

Consider a set \(\{x_1,\ldots,x_n\}\in\mathbb{R}^{p}\) of $p$-dimensional input vectors belonging to either one of two affinity classes \(\mathcal{C}_1\) or \(\mathcal{C}_2\). In graph-based methods, data samples \(x_1,\ldots,x_n\) are represented by vertices in a graph, upon which a weight matrix \(W\) is computed by
\begin{equation}
\label{eq:definition W}
    W=\{w_{ij}\}_{i,j=1}^n=\left\{h\left(\frac{1}{p}\Vert x_i-x_j\Vert^2\right)\right\}_{i,j=1}^n
\end{equation}
for some decreasing non-negative function \(h\),
so that nearby data vectors \(x_i\), \(x_j\) are connected with a large weight \(w_{ij}\), which can also be seen as a similarity measure between data samples. A typical kernel function for defining \(w_{ij}\) is the radial basis function kernel \(w_{ij}=e^{-\Vert x_i-x_j\Vert^2/t}\). The connectivity of data point \(x_i\) is measured by its degree \(d_i=\sum_{j=1}^n w_{ij}\), the diagonal matrix \(D\in\mathbb{R}^{n\times n}\) having \(d_i\) as its diagonal elements is called the degree matrix.

\medskip

Graph learning approach assumes that data points belonging to the same affinity group are ``close'' in a graph-proximity sense. In other words, if \(f\in\mathbb{R}^n\) is a class signal of data samples \(x_1,\ldots,x_n\), it varies little from \(x_i\) to \(x_j\) when \(w_{ij}\) has a large value. The graph smoothness assumption is usually characterized as minimizing a smoothness penalty term of the form
\begin{equation*}
    \frac{1}{2}\sum_{i,j=1}^n w_{ij}(f_i-f_j)^2=f^\trans L f
\end{equation*}
where \(L=D-W\) is referred to as the Laplacian matrix. Notice that the above loss function is minimized to zero for \(f=1_n\); obviously such constant vector contains no information about data classes. According to this remark, the popular unsupervised graph learning method, spectral clustering, simply consists in finding a unit vector orthogonal to \(1_n\) that minimizes the smoothness penalty term, as formalized below
\begin{gather}
\min_{f\in\mathbb{R}^n} f^\trans L f\nonumber\\
s.t.\quad \Vert f\Vert=1 \quad f^\trans 1_n=0.
\end{gather}
It is easily shown by the spectral properties of Hermitian matrices that the solution to the above optimization is the eigenvector of \(L\) associated to the second smallest eigenvalue. There exist also other variations of the smoothness penalty term involving differently normalized Laplacian matrices, such as the symmetric normalized Laplacian matrix \(L_s=I_n-D^{-\frac{1}{2}}W D^{-\frac{1}{2}}\), and the random walk normalized Laplacian matrix \(L_r=I_n-WD^{-1}\), which is related to the PageRank algorithm. 

\medskip

In the semi-supervised setting, we dispose of \(n_{[l]}\) pairs of labelled points and labels \(\{(x_1, y_1),\ldots,(x_{n_{[l]}},y_{n_{[l]}})\}\) with \(y_i\in\{-1,1\}\) the class label of \(x_i\), and \(n_{[u]}\) unlabelled data \(\{x_{n_{[l]}+1},\ldots,x_n\}\). To incorporate the prior knowledge on the class of labelled data into the class signal \(f\), the semi-supervised graph regularization approach imposes deterministic scores at the labelled points of \(f\), e.g., by letting \(f_i=y_i\) for all \(x_i\) labelled. 
The mathematical formulation of the problem then becomes
\begin{gather}
\label{eq:optimization Laplacian regularization}
\min_{f\in\mathbb{R}^n} f^\trans L f\nonumber\\
s.t.\quad f_i=y_i,\quad 1\leq i \leq n_{[l]}.
\end{gather}
Denoting
\begin{align*}
	f &= \begin{bmatrix}f_{[l]} \\f_{[u]}\end{bmatrix},~L=\begin{bmatrix}L_{[ll]} & L_{[lu]}\\L_{[ul]} & L{[uu]}\end{bmatrix},
\end{align*} the above convex optimization problem with equality constrains on \(f_{[l]}\) is realized by letting the derivative of the loss function with respect to \(f_{[u]}\) equal zero, which gives the following explicit solution
\begin{equation}
\label{eq:solution Laplacian regularization}
    f_{[u]}=-L_{[uu]}^{-1}L_{[ul]}f_{[l]}.
\end{equation}
Finally, the decision step consists in assigning unlabelled sample \(x_i\) to \(\mathcal{C}_1\) (resp., \(\mathcal{C}_2\)) if \(f_i<0\) (resp., \(f_i>0\)). 

The aforementioned method is frequently referred to as Laplacian regularization, for it finds the class scores of unlabelled data \(f_{[u]}\) by regularizing them over the Laplacian matrix along with predefined class signals of labelled data \(f_{[l]}\). It is often observed in practice that using other normalized Laplacian regularizers such as \(f^\trans L_s f\) or \(f^\trans L_r f\) can lead to better classification results. Similarly to the work of \citet{avrachenkov2012generalized}, we define $$L^{(a)}=I-D^{-1-a}WD^a$$ as the \(a\)-normalized Laplacian matrix in order to integrate all these different Laplacian regularization algorithms into a common framework. Replacing \(L\) with \(L^{(a)}\) in \eqref{eq:solution Laplacian regularization} to get
\begin{equation}
\label{eq:solution a-normalized Laplacian regularization}
    f_{[u]}=-\left(L_{[uu]}^{(a)}\right)^{-1}L_{[ul]}^{(a)}f_{[l]},
\end{equation}
we retrieve the solutions of standard Laplacian \(L\), symmetric Laplacian \(L_s\) and random walk Laplacian \(L_r\) respectively at \(a=0\), \(a=-1/2\) and \(a=-1\). 

{\BLUE Note additionally that the matrix \(L_{[uu]}^{(a)}\) is invertible under the trivial condition that the graph represented by \(W\) is fully connected (i.e., with no isolated subgraphs). To show this, note first that, under this condition, we have $$u_{[u]}^{\trans}D_{[u]}^{1+2a}L_{[uu]}^{(a)}u_{[u]}=\sum_{i,j=n_{[l]}+1}^nw_{ij}(d_i^au_i-d_j^au_j)^2+\sum_{i=n_{[l]}+1}^nd_i^{2a}u_i^2\sum_{m=1}^{n_{[l]}}w_{im}>0$$
for any \(u_{[u]}\neq 0_{n_{[u]}}\in\RR^{n_{[u]}}\), as the first term on the right-hand side is strictly positive unless all \(d_i^au_i\) have the same positive value, in which case the second term is strictly positive for there is at least one \(w_{im}>0\). The matrix \(L_{[uu]}^{(a)}\) is therefore positive definite. As shown in the following though, the fully connected condition is not required for the new algorithm proposed in this article to be well defined and to perform as expected.} 
\medskip

Despite being a popular semi-supervised learning approach, Laplacian regularization algorithms are shown by \citet{mai2018random} to have a non-efficient learning capacity for high dimensional unlabelled data, as a direct consequence of the distance concentration phenomenon, hinted at in the introduction. A deeper examination of the results in the work of \citet{mai2018random} allows us to discover that this problem of unlabelled data learning efficiency % of semi-supervised graph regularization 
may in fact be settled through the usage of a centered similarity measure, as opposed to the current convention of non-negative similarities \(w_{ij}\). In the following subsections, we will recall the findings in the analysis of \citet{mai2018random}, then move on to the proposition of the novel corrective algorithm, along with some general remarks explaining the effectiveness of the proposed algorithm, leaving the thorough performance analysis to the next section.

\subsection{High Dimensional Behaviour of Laplacian Regularization}
\label{sec:previous results}

Conforming to the settings employed by \citet{mai2018random}, we adopt the following high dimensional data model for the theoretical discussions in this paper.
\begin{assumption}
\label{ass:data model}
	Data samples $x_1,\ldots,x_n$ are i.i.d.\@ observations from a generative model such that, for \(k\in\{1,2\}\),  \(\mathbb{P}(x_i\in\mathcal{C}_k)=\rho_k\), and
	\begin{align*}
	x_i\in \mathcal C_k \Leftrightarrow x_{i}\sim\mathcal N(\mu_{k},C_{k})
	\end{align*}
	with \(\Vert C_k \Vert=O(1)\), \(\Vert C_k^{-1} \Vert=O(1)\), \(\Vert\mu_2-\mu_1\Vert=O(1)\), \({\rm tr}C_1-{\rm tr}C_2=O(\sqrt{p})\) and  \({\rm tr}(C_1-C_2)^2=O(\sqrt{p})\).
	
	The ratios \(c_0=\frac{n}{p}\), \(c_{[l]}=\frac{n_{[l]}}{p}\) and \(c_{[u]}=\frac{n_{[u]}}{p}\) are  bounded away from zero for arbitrarily large \(p\).
\end{assumption}

Here are some remarks to interpret the conditions imposed on the data means \(\mu_k\) and covariance matrices \(C_k\) in Assumption~\ref{ass:data model}. Firstly, as the discussion is placed under a large dimensional context, we need to ensure that the data vectors do not lie in a low dimensional manifold; the fact that \(\Vert C_k \Vert=O(1)\) along with \(\Vert C_k^{-1} \Vert=O(1)\) guarantees non-negligible variations in \(p\) linearly independent directions. Other conditions controlling the differences between the class statistics \(\Vert\mu_2-\mu_1\Vert=O(1)\), \({\rm tr}C_1-{\rm tr}C_2=O(\sqrt{p})\), and \({\rm tr}(C_1-C_2)^2=O(\sqrt{p})\) are made for the consideration of establishing {\it non-trivial classification} scenarios where the classification of unlabelled data does not become impossible or overly easy at extremely large values of \(p\). %is a bounded function dependent of the data statistics.

The first result concerns the distance concentration of high dimension data. This result is at the core of the reasons why Laplacian-based semi-supervised learning is bound to fail with large dimensional data.
%%%
\begin{proposition}
\label{prop:distance concentration}
Define \(\tau={\rm tr}(C_1+C_2)/p\). Under Assumption~\ref{ass:data model}, we have that, for all \(i,j\in\{1,\ldots,n\}\),
\begin{equation*}
\frac{1}{p}\Vert x_i-x_j \Vert^2=\tau+O(p^{-\frac{1}{2}}).
\end{equation*}
\end{proposition}

The above proposition indicates that in large dimensional spaces, all pairwise distances of data samples converge to the same value, thereby indicating that the presumed connection between {\it proximity and data affinity} is completely disrupted. In such situations, the performance of the Laplacian regularization approach (along with most distance-based classification methods), which normally works well in small dimensions, may be severely affected. Indeed, under some mild smooth conditions on the weight function \(h\), the analysis of \citet{mai2018random} reveals several surprising and critical aspects of the high dimensional behavior of this approach. The first conclusion is that all unlabelled data scores \(f_i\) for \(n_{[l]}+1\leq i\leq n\) tend to have the same signs in the case of unequal class priors (i.e., \(\rho_1\neq\rho_2\)), causing all unlabelled data to be classified in the same class (unless one normalizes the deterministic scores at labelled points so that they are balanced for each class). In accordance with this message, we shall use in the remainder of the article a class-balanced \(f_{[l]}\) defined as below
\begin{equation}
\label{eq:balanced f_l}
    f_{[l]}=\left(I_{n_{[l]}}-\frac{1}{n_{[l]}}1_{n_{[l]}}1_{n_{[l]}}^\trans\right)y_{[l]}
\end{equation}
where \(y_{[l]}\in\mathbb{R}^{n_{[l]}}\) is the label vector composed of \(y_i\) for \(1\leq i\leq n_{[l]}\). 

Nevertheless, even with balanced \(f_{[l]}\) as per \eqref{eq:balanced f_l}, \citep{mai2018random} shows that the aforementioned ``all data affected to the same class'' problem still persists for all Laplacian regularization algorithms under the framework of \(a\)-normalized Laplacian (i.e., for $L^{(a)}=I-D^{-1-a}WD^a$) except for \(a\simeq -1\). This indicates that among all existing Laplacian regularization algorithms proposed in the literature, only the random walk normalized Laplacian regularization yields non-trivial classification results for large dimensional data. We recall in the following theorem the exact statistical characterization of \(f_{[u]}\) produced by the random walk normalized Laplacian regularization, which was firstly presented by \citet{mai2018random}.

\begin{theorem}
\label{th:statistics of fu for Laplacian method}
Let Assumption~\ref{ass:data model} hold, the function \(h\) of \eqref{eq:definition W} be three-times continuously differentiable in a neighborhood of \(\tau\), and the solution \(f_{[u]}\) be given by \eqref{eq:solution a-normalized Laplacian regularization} for \(a=-1\). Then, for \(n_{[l]}+1\leq i\leq n\) (i.e., \(x_i\) unlabelled) and \(x_i\in\mathcal{C}_k\),
\begin{equation*}
    p(c_0/2\rho_1\rho_2c_{[l]})f_i=\tilde{f}_i+o_P(1)\text{, where }\tilde{f}_i\sim\mathcal{N}(m_k,\sigma_k^2)
\end{equation*}
with
\begin{align}
    m_k&=(-1)^k(1-\rho_k)\left[-\frac{2h'(\tau)}{h(\tau)}\Vert \mu_1-\mu_2\Vert^2+\left(\frac{h''(\tau)}{h(\tau)}-\frac{h'(\tau)^2}{h(\tau)^2}\right)\frac{\left({\rm tr}C_1-{\rm tr}C_2\right)^2}{p}\right] \label{eq:means of fu for Laplacian method}\\
    \sigma_k^2&=\frac{4h'(\tau)^2}{h(\tau)^2}\left[(\mu_1-\mu_2)^\trans C_k(\mu_1-\mu_2)+\frac{1}{c_{[l]}}\frac{\sum_{a=1}^2(\rho_a)^{-1}{\rm tr}C_aC_k}{p}\right]\nonumber\\
    &+\left(\frac{h''(\tau)}{h(\tau)}-\frac{h'(\tau)^2}{h(\tau^2)}\right)^2\frac{2{\rm tr} C_k^2\left({\rm tr}C_1-{\rm tr}C_2\right)^2}{p^2}.\label{eq:variances of fu for Laplacian method}
\end{align}
\end{theorem}

Theorem~\ref{th:statistics of fu for Laplacian method} states that the classification scores \(f_i\) for an unlabelled \(x_i\) follows approximately a Gaussian distribution at large values of \(p\), with the mean and variance being explicitly dependent of the data statistics \(\mu_k\), \(C_k\), the class proportions \(\rho_k\), and the ratio of labelled data over dimensionality \(c_{[l]}\). The asymptotic probability of correct classification for unlabelled data is then a direct result of Theorem~\ref{th:statistics of fu for Laplacian method}, and reads
\begin{align}
\label{eq:probability of correct classification}
    \mathcal{P}(x_i\to C_k|x_i\in C_k, i>n_{[l]})=\Phi\left(\sqrt{m_k^2/\sigma_k^2}\right)+o_p(1)
\end{align}
where \(\Phi(u)=\frac1{2\pi}\int_{-\infty}^ue^{-\frac{t^2}2}dt\) is the cumulative distribution function of the standard Gaussian distribution. 

Of utmost importance here is the observation that, while \(m_k^2/\sigma_k^2\) is an increasing function of \(c_{[l]}\), suggesting an effective learning from the labelled set, it is {\it independent of the unlabelled data ratio \(c_{[u]}\)}, which tells us that in the case of high dimensional data, the addition of unlabelled data, even in significant numbers with respect to the dimensionality \(p\), produces negligible performance gain. Motivated by this crucial remark, we propose in this paper a {\it simple and fundamental update} to the classical Laplacian regularization approach, for the purpose of boosting high dimensional learning performance through an enhanced utilization of unlabelled data. The proposed algorithm will be presented and intuitively justified in the next subsection.

\section{Proposed Regularization with Centered Similarities}
\label{sec:regularization with centered similarities}

{\BLUE As will be put forward in Subsection~\ref{sec:problem identification}, we find that the unlabelled data learning efficiency problem of the Laplacian regularization method, revealed by \citet{mai2018random}, is rooted in the concentration of pairwise distances between data vectors of high dimensionality. To counter the disastrous effect of the distance concentration problem, a new regularization approach with centered similarities is proposed in Subsection~\ref{sec:proposition of the method}.  An alternative interpretation of the proposed method from the perspective of label propagation is given in Subsection~\ref{sec:interpretation and properties}, justifying its usage in general scenarios beyond the discussed high dimensional regime.}

\subsection{Problem Identification}
\label{sec:problem identification}

To gain perspective on the cause of inefficient learning from unlabelled data, we will start with a discussion linking the issue to the data high dimensionality. 

Developing \eqref{eq:solution a-normalized Laplacian regularization}, we get
\begin{equation*}
f_{[u]}=L^{(a)-1}_{[uu]}D_{[u]}^{-1-a}W_{[ul]}D_{[l]}^{a}f_{[l]}
\end{equation*}
where
\begin{align*}
W=\begin{bmatrix}W_{[ll]} & W_{[lu]}\\W_{[ul]} & W_{[uu]}\end{bmatrix}\text{ and }D=\begin{bmatrix}D_{[l]} & 0\\0 & D_{[u]}\end{bmatrix}.
\end{align*}

From a graph-signal processing perspective \cite{SHU13}, since \(L^{(a)}_{[uu]}\) is the Laplacian matrix on the subgraph of unlabelled data, and a smooth signal \(s_{[u]}\) on the unlabelled data subgraph typically induces large values for the inverse smoothness penalty \(s_{[u]}^\trans L^{(a)-1}_{[uu]}s_{[u]}\), we may consider the operator \(\mathcal{P}_{u}(s_{[u]})=L^{(a)-1}_{[uu]}s_{[u]}\) as a ``smoothness filter'' strengthening smooth signals on the unlabelled data subgraph. The unlabelled scores \(f_{[u]}\) can be therefore seen as obtained by a two-step procedure:
\begin{enumerate}
	\item propagating the predetermined labelled scores \(f_{[l]}\) through the graph with the \(a\)-normalized weight matrix \(D_{[u]}^{-1-a}W_{[ul]}D_{[l]}^{a}\) through the label propagation operator \(\mathcal{P}_l(f_{[l]})=D_{[u]}^{-1-a}W_{[ul]}D_{[l]}^{a}f_{[l]}\);
	\item passing the received scores at unlabelled points through the smoothness filter \(\mathcal{P}_{u}(s_{[u]})=L^{(a)-1}_{[uu]}s_{[u]}\) to finally get \(f_{[u]}=\mathcal{P}_{u}\left(\mathcal{P}_l(f_{[l]})\right)\).
\end{enumerate}
It is easy to see that the first step is essentially a supervised learning process, whereas the second one allows to capitalize on the global information contained in unlabelled data. However, as a consequence of the distance concentration ``curse'' stated in Proposition~\ref{prop:distance concentration}, the similarities (weights) \(w_{ij}\) between high dimensional data vectors are dominated by the constant value \(h(\tau)\) plus some small fluctuations, which results in the collapse of the smoothness filter: \begin{equation*}
\mathcal{P}_u(s_{[u]})=L^{(a)-1}_{[uu]}s_{[u]}\simeq\left(I_{n_{[u]}}-\frac{1}{n}1_{n_{[u]}}1_{n_{[u]}}^\trans\right)^{-1}s_{[u]}=s_{[u]}+ \frac1{n_{[l]}} (1_{n_{[u]}}^\trans s_{[u]})1_{n_{[u]}},
\end{equation*}
meaning that at large values of \(p\), only the constant signal direction \(1_{n_{[u]}}\) is amplified by the smoothness filter \(\mathcal{P}_u\). 

To understand such behavior of the smoothness filter \(\mathcal{P}_u\), we recall that as mentioned in Subsection~\ref{sec:laplacian algorithms}, constant signals with the same value at all points are always considered to be the most smooth on the graph. This comes from the fact that all weights \(w_{ij}\) have non-negative value, so the smoothness penalty term \(\mathcal{Q}(s)=\sum_{i,j}w_{[ij]}(s_i-s_j)^2\) is minimized at the value of zero if all elements of the signal \(s\) have the same value. Notice also that in perfect situations where the data points in different class subgraphs are connected with zero weights \(w_{ij}\), class indicators (i.e., signals with constant values within class subgraphs which are different for each class) are just as smooth as constant signals for they also minimize the smoothness penalty term to zero. Even though such scenarios almost never happen in real life, it is hoped that the inter-class similarities are sufficiently weak so that the smoothness filter \(\mathcal{P}_u\) is still effective. What is problematic for high dimensional learning is that when the similarities \(w_{ij}\) tend to be indistinguishable due to the distance concentration issue of high dimensional data vectors, constant signals have overwhelming advantages to the point that they become the only direction privileged by the smoothness filter \(\mathcal{P}_u\), with almost no discrimination between all other directions. In consequence, there is nearly no utilization of the global information in high dimensional unlabelled data through Laplacian regularizations.

\smallskip

In view of the above discussion, we shall try to eliminate the dominant advantages of constant signals, in an attempt to render detectable the discrimination between class-structured signals and other noisy directions. As constant signals always have a smoothness penalty of zero, a very easy way to break their optimal smoothness is to introduce negative weights in the graph so that the values of the smoothness regularizer can go below zero. More specifically, in the cases where the intra-class similarities are averagely positive and the inter-class similarities are averagely negative, class-structured signals are bound to have a lower smoothness penalty than constant signals. However, the implementation of such idea using both positive and negative similarities is hindered by the fact that the positivity of the data points degrees \(d_i=\sum_{j=1}^n w_{ij}\) is no longer ensured, and having negative degrees can lead to severely unstable results. Take for instance the label propagation step \(\mathcal{P}_l(f_{[l]})=D_{[u]}^{-1-a}W_{[ul]}D_{[l]}^{a}f_{[l]}\), at an unlabelled point \(x_i\), the sum of the received scores after that step equals to \(d_i^{-1-a}\sum_{j=1}^{n_{[l]}}(w_{ij}d_j^a)f_j\), the sign of which obviously alters if the signs of the degree of that point and those of labelled data change, leading thus to extremely unstable classification results. 

\subsection{Proposition of the Method}
\label{sec:proposition of the method}

To cope with the problem identified above, we propose here the usage of centered similarities \(\hat{w}_{ij}\), for which the positive and negative weights are balanced out at any data point, i.e., for all \(i\in\{1,\ldots,n\}\), \(d_i=\sum_{j=1}^n w_{ij}=0\). Given any similarity matrix \(W\), its centered version \(\hat{W}\) is easily obtained by applying a projection matrix \(P=\left(I_n-\frac{1}{n}1_n1_n^\trans\right)\) on both sides:
\begin{equation*}
    \hat{W}=PWP.
\end{equation*}
As a first advantage, the centering approach allows to remove the degree matrix altogether (for the degrees are exactly zero now) from the updated smoothness penalty
\begin{equation}
\label{eq:smoothness penalty with centered similarities}
    \hat{Q}(s)=\sum_{i,j=1}^n\hat{w}_{ij}(s_i-s_j)^2=-s^\trans\hat{W}s,
\end{equation}
securing thus a stable behavior of graph regularization with both positive and negative weights. 

\smallskip

This being said, a problematic consequence of regularization procedures employing positive and negative weights is that the optimization problem is no longer convex and may have an infinite solution. To deal with this issue, we add a constraint on the norm of the solution. Letting \(f_{[l]}\) be given by \eqref{eq:balanced f_l}, the new optimization problem may now be posed as follows:
\begin{gather}
\min_{f_{[u]}\in\mathbb{R}^{n_{[u]}}}-f^\trans\hat{W}f\nonumber\\
s.t.\Vert f_{[u]}\Vert^2=n_{[u]}e^2
\label{eq:centered kernel ssl}
\end{gather}
for some \(e>0\).

The optimization can be solved by introducing a Lagrange multiplier \(\alpha=\alpha(e)\) to the norm constraint \(\Vert f_{[u]}\Vert^2=n_{[u]}e^2\) and {\BLUE the solution reads
\begin{align}
\label{eq:centered kernel solution}f_{[u]}=\left(\alpha I_{n_{[u]}}-\hat{W}_{[uu]}\right)^{-1}\hat{W}_{[ul]}f_{[l]}
\end{align}
with $\alpha >\Vert \hat{W}_{[uu]}\Vert$ uniquely given by 
\begin{equation}
\label{eq:relation alpha and e}
	\left\Vert \left(\alpha I_{n_{[u]}}-\hat{W}_{[uu]}\right)^{-1}\hat{W}_{[ul]}f_{[l]}\right\Vert^2=n_{[u]}e^2.
\end{equation}

To see that \eqref{eq:centered kernel ssl} is the unique solution to the optimization problem \eqref{eq:centered kernel ssl},  it is useful to remark that, by the properties of convex optimization, \eqref{eq:centered kernel solution} is the unique solution to the unconstrained convex optimization problem $\min_{f_{[u]}} \alpha\Vert f_{[u]}\Vert^2-f^\trans\hat{W}f$ for some $\alpha>\Vert \hat{W}_{[uu]}\Vert$. When Equation~\eqref{eq:relation alpha and e} is satisfied, we get (through a proof by contradiction) that \eqref{eq:centered kernel solution} is the only solution that minimizes \(-f^\trans\hat{W}f\) in the subspace defined by  $\Vert f_{[l]}\Vert^2=n_{[u]}e^2$.}

In practice, \(\alpha\) can be used directly as a hyperparameter for a more convenient implementation. We summarize the method in Algorithm~\ref{alg:centered regularization}.

\begin{algorithm}
	\caption{Semi-Supervised Graph Regularization with Centered Similarities}
\label{alg:centered regularization}
 \begin{algorithmic}[1]
	 \STATE {\bf Input:} \(n_{[l]}\) pairs of labelled points and labels \(\{(x_1, y_1),\ldots,(x_{n_{[l]}},y_{n_{[l]}})\}\) with \(y_i\in\{-1,1\}\) the class label of \(x_i\), and \(n_{[u]}\) unlabelled data \(\{x_{n_{[l]}+1},\ldots,x_n\}\).
	 \STATE {\bf Output:} Classification of unlabelled data \(\{x_{n_{[l]}+1},\ldots,x_n\}\).
	 \STATE Compute the similarity matrix \(W\).
	 \STATE Compute the centered similarity matrix \(\hat{W}=PWP\) with \(P=I_n-\frac{1}{n}1_n1_n^\trans\), and define \(\hat{W}=\begin{bmatrix}\hat{W}_{[ll]} & \hat{W}_{[lu]}\\\hat{W}_{[ul]} & \hat{W}_{[uu]}\end{bmatrix}\).
	 \STATE Set     \(f_{[l]}=\left(I_{n_{[l]}}-\frac{1}{n_{[l]}}1_{n_{[l]}}1_{n_{[l]}}^\trans\right)y_{[l]}\) with \(y_{[l]}\) the vector containing labelled \(y_i\).
	 \STATE Compute the class scores of unlabelled data 
	     \(f_{[u]}=\left(\alpha I_{n_{[u]}}-\hat{W}_{[uu]}\right)^{-1}\hat{W}_{[ul]}f_{[l]}\)
	 for some \(\alpha>\Vert \hat{W}_{[uu]}\Vert\).
	 \STATE Classify unlabelled data \(\{x_{n_{[l]}+1},\ldots,x_n\}\) by the signs of \(f_{[u]}\).
\end{algorithmic}
\end{algorithm}

The proposed algorithm induces almost no extra cost to the classical Laplacian approach, except the addition of the hyperparameter \(\alpha\) controlling the norm of \(f_{[u]}\). The performance analysis in Section~\ref{sec:analysis} will help demonstrate that the existence of this hyperparameter, aside from making the regularization with centered similarities a well-posed problem, allows one to adjust the combination of labelled and unlabelled information in search for an optimal semi-supervised learning performance. {\BLUE As a justification of its usage in a general context (beyond the discussed high dimensional regime), the following subsection provides an alternative interpretation of the proposed method from the perspective of label propagation. }

\subsection{Alternative viewpoint of label propagation}
\label{sec:interpretation and properties}

{\BLUE
% Before the performance analysis of Section~\ref{sec:analysis}, which supports the usage of the proposed method for a consistent high dimensional semi-supervised learning, we provide here some mathematical arguments that serve as more general justification of the proposed method and help explain its empirical superiority in various scenarios beyond the discussed high dimensional regime, as will be shown in the experimentations on real-world data sets in Subsection and the simulations under the stochastic block model in Subsection .  
	
Similarly to Laplacian regularization, the proposed method can be interpreted from the perspective of label propagation \citep{zhu2002learning}. Setting \(f_{[u]}^{(0)}\leftarrow \alpha^{-1}\hat W_{[ul]}f_{[l]}\), we retrieve the solution \eqref{eq:centered kernel solution} of centered regularization at the stationary point $f_{[u]}^{(\infty)}$ of the following iteration: 
\begin{align*}
	f_{[u]}^{(t+1)}\leftarrow \alpha^{-1}\begin{bmatrix}
	0_{n_{[u]}\times n_{[l]}} & I_{n_{[u]}}
	\end{bmatrix}P W P \begin{bmatrix}
	f_{[l]} \\ f_{[u]}^{(t)}
	\end{bmatrix}.
\end{align*}
Denoting $f^{(t)}=[f_{[l]},f_{[u]}^{(t)}]^\trans$, the above process can be seen as propagating the centered score vector \(\hat f^{(t)}= Pf^{(t)}\) through the weight matrix \(W\) and recentering the received scores \(\eta^{(t)}=W\hat f^{(t)}\) before outputting \(f_{[u]}^{(t+1)}\) as the subset of \(\hat\eta^{(t)}=P\eta^{(t)}\) corresponding to the unlabelled points. 

Recall from the discussion in Subsection~\ref{sec:problem identification} that the extremely amplified constant signal \(1_{n_{[u]}}\) in the outcome \(f_{[u]}\) of the Laplacian method is closely related to the ineffective unlabelled data learning problem. In the proposed approach, the constant signal is cancelled thanks to the recentering operations before and after the label propagation over \(W\). The existence of the multiplier \(\alpha^{-1}\)  allows us to magnify the score vector, after its norm was significantly reduced due to the recentering operations.

Since the regularization method with centered similarities can be viewed as a label propagation of the recentered score vector over the original weight matrix \(W\), the proposed method, despite being motivated under the scenario of high dimensional learning, is expected to yield competitive (if not superior) performance even when the original Laplacian approach works well thanks to an informative weight matrix  \(W\). This claim is notably supported by simulations which will be displayed in Subsection~\ref{sec:sparse graphs}, where the proposed method is observed to perform better than the Laplacian regularization (and other graph-based SSL algorithms) on sparse graphs with connections within the same class significantly more frequent than those between different classes.

}

\section{Performance Analysis}% of Proposed Method}
\label{sec:analysis}

The main purpose of this section is to provide mathematical support for its effective high dimensional learning capabilities from not only labelled data but also from unlabelled data, allowing for a theoretically guaranteed performance gain over the classical Laplacian approach (through an enhanced utilization of unlabelled data). The theoretical results also point out that the learning performance of the proposed method has an unlabelled data learning efficiency that is at least as good as spectral clustering, as opposed to Laplacian regularization.

\subsection{Statistical Characterization}
\label{sec:statistical characterization}

We provide here the statistical characterization of unlabelled data scores \(f_{[u]}\) obtained by the proposed algorithm. As the new algorithm will be shown to draw on both labelled and unlabelled data, the complex interactions between these two types of data generate more intricate outcomes than in \citep{mai2018random}. To facilitate the interpretation of the theoretical results without cumbersome notations, we present the theorem here under the homoscedasticity of data vectors, i.e., \(C_1=C_2=C\), without affecting the generality of the conclusions given subsequently. We refer the interested reader to the appendix for an extended version of the theorem along with its proof.

%\begin{proposition}
%	Let Assumption~\ref{ass:data model} hold with \(C_1=C_2=C\), \(h\) be three-times continuously differentiable in a neighborhood of \(\tau\), and \(f_{[u]}\) be the solution of \eqref{eq:centered kernel ssl}. Then, there exists a constant \(\gamma>0\) such that, for \(n_{[l]}+1\leq i\leq n\) (i.e., \(x_i\) unlabelled),
%	\begin{equation*}
%		f_i-\tilde f_i=o_P(1) \text{ where } \tilde f_i\overset{\mathcal{L}}{=}\gamma \beta^\trans \hat x'_i
%	\end{equation*}
%	for some \(\hat x'_i\)  independent of \(\beta\) which follows the same distribution as \(\hat x_i=x_i-\frac{1}{n}\sum_{j=1}^nx_j\), and
%	
%\begin{equation*}
%\Vert\beta-\tilde\beta\Vert=o_P(1) \text{ where } \tilde\beta\overset{\mathcal{L}}{=}\left(I_p-\gamma c_{[u]}C\right)^{-1}X'f
%\end{equation*}
% with \(X'\) an independent copy of the data matrix \(X=[x_1,\ldots,x_n]\).

%\end{proposition}
\medskip

{\BLUE 
	We introduce first two positive functions \(m(\xi)\) and \(\sigma^2(\xi)\) which are crucial for describing the statistical distribution of unlabelled scores:
	
	\begin{align}
	&m(\xi)=\frac{2c_{[l]}\theta(\xi)}{c_{[u]}\big(1-\theta(\xi)\big)}\label{eq:m(xi)}\\
	&\sigma^2(\xi)=\frac{\rho_1\rho_2(2c_{[l]}+m(\xi)c_{[u]})^2s(\xi)+\rho_1\rho_2(4c_{{l}}+m(\xi)^2c_{[u]})q(\xi)}{c_{[u]}\left(c_{[u]}-q(\xi)\right)}\label{eq:sigma2(xi)}
	\end{align}
	where
\begin{align*}
    \theta(\xi)&=\rho_1\rho_2\xi(\mu_1-\mu_2)^\trans\left(I_p-\xi C\right)^{-1}(\mu_1-\mu_2)\\
	q(\xi)&=\xi^2p^{-1}\tr\left[\left(I_p-\xi C\right)^{-1}C\right]^2\\
	s(\xi)&=\rho_1\rho_2\xi^2(\mu_1-\mu_2)^\trans \left(I_p-\xi C\right)^{-1}C\left(I_p-\xi C\right)^{-1}(\mu_1-\mu_2).
\end{align*}
Here the positive functions \(m(\xi)\) and \(\sigma^2(\xi)\) are defined respectively on the domains \((0,\xi_m)\) and \((0,\xi_{\sigma^2})\) with \(\xi_m,\xi_{\sigma^2}>0\) uniquely given  by \(\theta(\xi_m)=1\) and \(q(\xi_{\sigma^2})=c_{[u]}\). Additionally, we define 
\begin{equation}
\xi_{\sup}=\min\{\xi_m,\xi_{\sigma^2}\} \label{eq:xi_sup}.
\end{equation}

\medskip

These definitions may at first glance seem complicated, but it suffices to keep in mind a few key messages to understand the theoretical results and their implications: 
\begin{itemize}
	\item \(\theta(\xi)\), \(q(\xi)\) and \(s(\xi)\)  are all positive and strictly increasing functions for \(\xi\in(0,\xi_{\sup})\); consequently so are \(m(\xi)\) and \(\sigma^2(\xi)\). 
	\item \(\xi_m\) does not depend on \(c_{[l]}\) or \(c_{[u]}\); as for \(\xi_{\sigma^2}\), it is constant with \(c_{[l]}\) but increases as \(c_{[u]}\) increases.
	\item \(\rho_1\rho_2m^2(\xi)+\sigma^2(\xi)\) monotonously increases from zero to infinity as \(\xi\) increases from zero to \(\xi_{\sup}\).
\end{itemize}
The above remarks can be derived directly from the definitions of the involved mathematical objects.

%Here we are interested in the regime where \(m(\xi),\sigma^2(\xi)\) are both positive, which is only achieved under the conditions \(\xi>0\) and \(\theta(\xi),q(\xi)<c_{[u]}\). To specify the range of \(\xi\) for which \(m(\xi),\sigma^2(\xi)\geq 0\), we start by defining \(\xi_m(c_{[u]})\) as the solution of \(\theta(\xi)=c_{[u]}\). Since \(\theta(\xi)\) has well-defined positive value if and only if \(\xi\in[0,\Vert C\Vert)\) and \(\theta(\xi)\) is a strictly increasing function for \(\xi\in[0,\Vert C\Vert)\) , \(\xi_m(c_{[u]})\) has a unique value that is within the interval \([0,\Vert C\Vert)\) and goes up with \(c_{[u]}\). As \(q(\xi)\) is also a strictly increasing function for \(\xi\in[0,\Vert C\Vert)\), we denote by \(\xi_{\sigma^2}(c_{[u]})\) the unique solution of \(q(\xi)=c_{[u]}\) under the constraint \(\xi\in[0,\Vert C\Vert]\). Define 
%\begin{equation}
%	\xi_{\sup}(c_{[u]})=\min\{\xi_m(c_{[u]}),\xi_{\sigma^2}(c_{[u]})\},
%\end{equation}
%we conclude from the previous arguments that \(m(\xi),\sigma^2(\xi)\geq 0\) if and only if \(\xi\in\big[0,\xi_{\sup}(c_{[u]})\big)\).
%
%Remark importantly that as \(\theta(\xi),q(\xi)\) and \(s(\xi)\) are all strictly increasing functions for \(\xi\in[0,\Vert C\Vert)\), it is easy to deduce from \eqref{eq:m(xi)} and \eqref{eq:sigma2(xi)} that \(m(\xi),\sigma^2(\xi)\) are also strictly increasing functions for \(\xi\in\big[0,\xi_{\sup}(c_{[u]})\big)\) since \(\xi_{\sup}(c_{[u]})\leq\Vert C\Vert\). 

\begin{theorem}
\label{th:statistics of fu for centered method}
Let Assumption~\ref{ass:data model} hold with \(C_1=C_2=C\), the function \(h\) of \eqref{eq:definition W} be three-times continuously differentiable in a neighborhood of \(\tau\), \(f_{[u]}\) be the solution of \eqref{eq:centered kernel ssl} with fixed norm \(n_{[u]}e^2\) and with the notations of \(m(\xi)\), \(\sigma^2(\xi)\), \(\xi_{\sup}\) given in \eqref{eq:m(xi)}, \eqref{eq:sigma2(xi)}, \eqref{eq:xi_sup}. Then, for \(n_{[l]}+1\leq i\leq n\) (i.e., \(x_i\) unlabelled) and \(x_i\in\mathcal{C}_k\),

\begin{equation*}
    f_i=\tilde{f}_i+o_P(1)\text{, where }\tilde{f}_i\sim\mathcal{N}((-1)^k(1-\rho_k)\hat m,\hat\sigma^2)
\end{equation*}
with
\begin{equation*}
	\hat m=m(\xi_e),\quad \hat\sigma^2=\sigma^2(\xi_e)
\end{equation*}
for \(\xi_e\in(0,\xi_{\sup})\) uniquely given by \(\rho_1\rho_2m(\xi_e)^2+\sigma^2(\xi_e)=e^2\).
\end{theorem}}

\subsection{Consistent Learning from Labelled and Unlabelled Data}
\label{sec:consistent SSL}
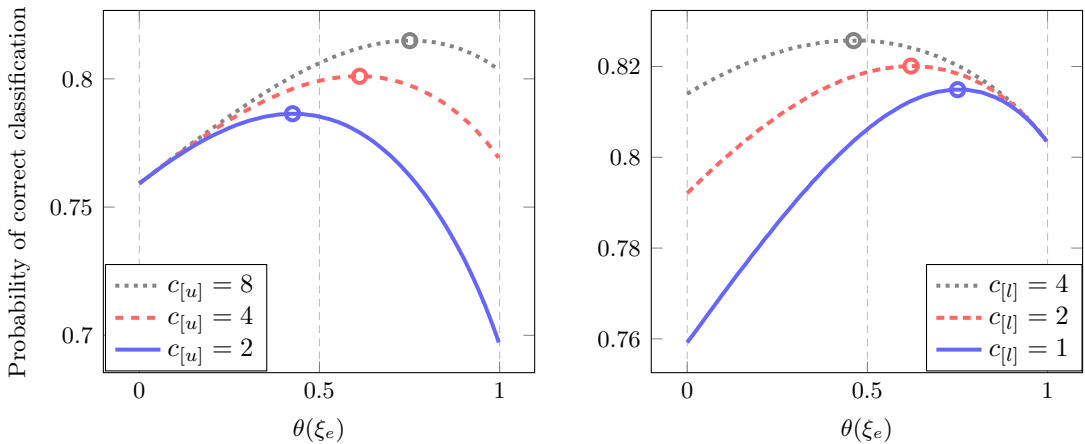
\begin{figure}
	\centering
	\begin{tabular}{cc}
		\begin{tikzpicture}[font=\footnotesize]
		\renewcommand{\axisdefaulttryminticks}{4} 
		\tikzstyle{every major grid}+=[style=densely dashed ] \tikzstyle{every axis y label}+=[yshift=-10pt] 
		\tikzstyle{every axis x label}+=[yshift=5pt]
		\tikzstyle{every axis legend}+=[cells={anchor=west},fill=white,
		at={(0.38,0.3)}, anchor=north east, font=\small]
		\begin{axis}[
		%ybar,
		width=.48\linewidth,
		height=.42\linewidth,
		%xmode=log,
		%log basis x={10},
		grid=major,
		ymajorgrids=false,
		scaled ticks=true,
		xtick={0,0.5,1},
		ylabel={Probability of correct classification},
		xlabel={\(\theta(\xi_e)\)},
		]
		\addplot[dotted,black!80!white!60!,line width=1.5pt] plot coordinates{
		(0.0000,0.7592)(0.0198,0.7613)(0.0392,0.7634)(0.0583,0.7655)(0.0769,0.7675)(0.0952,0.7695)(0.1132,0.7714)(0.1308,0.7733)(0.1482,0.7751)(0.1651,0.7768)(0.1818,0.7786)(0.1982,0.7802)(0.2143,0.7819)(0.2301,0.7834)(0.2456,0.7850)(0.2609,0.7865)(0.2759,0.7879)(0.2906,0.7893)(0.3051,0.7907)(0.3194,0.7920)(0.3334,0.7932)(0.3471,0.7945)(0.3607,0.7956)(0.3740,0.7968)(0.3871,0.7979)(0.4001,0.7989)(0.4128,0.7999)(0.4253,0.8009)(0.4376,0.8018)(0.4497,0.8027)(0.4616,0.8036)(0.4734,0.8044)(0.4850,0.8052)(0.4964,0.8059)(0.5076,0.8067)(0.5187,0.8073)(0.5296,0.8080)(0.5403,0.8086)(0.5509,0.8092)(0.5613,0.8097)(0.5716,0.8102)(0.5818,0.8107)(0.5918,0.8112)(0.6017,0.8116)(0.6114,0.8120)(0.6210,0.8124)(0.6304,0.8127)(0.6398,0.8130)(0.6490,0.8133)(0.6581,0.8136)(0.6670,0.8138)(0.6759,0.8140)(0.6846,0.8142)(0.6933,0.8144)(0.7018,0.8145)(0.7102,0.8146)(0.7185,0.8147)(0.7267,0.8148)(0.7347,0.8149)(0.7427,0.8149)(0.7506,0.8149)(0.7584,0.8149)(0.7661,0.8149)(0.7737,0.8149)(0.7812,0.8148)(0.7886,0.8147)(0.7960,0.8146)(0.8032,0.8145)(0.8104,0.8144)(0.8175,0.8143)(0.8245,0.8141)(0.8314,0.8139)(0.8382,0.8137)(0.8450,0.8135)(0.8517,0.8133)(0.8583,0.8131)(0.8648,0.8128)(0.8713,0.8126)(0.8777,0.8123)(0.8840,0.8120)(0.8902,0.8117)(0.8964,0.8114)(0.9026,0.8110)(0.9086,0.8107)(0.9146,0.8103)(0.9205,0.8100)(0.9264,0.8096)(0.9322,0.8092)(0.9380,0.8088)(0.9436,0.8084)(0.9493,0.8080)(0.9548,0.8075)(0.9604,0.8071)(0.9658,0.8066)(0.9712,0.8061)(0.9766,0.8056)(0.9819,0.8051)(0.9872,0.8046)(0.9924,0.8041)(0.9975,0.8035)
		};
		\addplot[dashed,red!60!white,line width=1.5pt] plot coordinates{
			(0.0000,0.7592)(0.0198,0.7613)(0.0392,0.7634)(0.0583,0.7654)(0.0769,0.7674)(0.0952,0.7693)(0.1132,0.7711)(0.1308,0.7729)(0.1482,0.7746)(0.1651,0.7762)(0.1818,0.7778)(0.1982,0.7793)(0.2143,0.7808)(0.2301,0.7822)(0.2456,0.7835)(0.2609,0.7848)(0.2759,0.7860)(0.2906,0.7872)(0.3051,0.7883)(0.3194,0.7894)(0.3334,0.7904)(0.3471,0.7914)(0.3607,0.7923)(0.3740,0.7931)(0.3871,0.7939)(0.4001,0.7947)(0.4128,0.7954)(0.4253,0.7961)(0.4376,0.7967)(0.4497,0.7973)(0.4616,0.7978)(0.4734,0.7983)(0.4850,0.7987)(0.4964,0.7991)(0.5076,0.7995)(0.5187,0.7998)(0.5296,0.8001)(0.5403,0.8003)(0.5509,0.8005)(0.5613,0.8007)(0.5716,0.8008)(0.5818,0.8009)(0.5918,0.8010)(0.6017,0.8010)(0.6114,0.8010)(0.6210,0.8010)(0.6304,0.8010)(0.6398,0.8009)(0.6490,0.8008)(0.6581,0.8006)(0.6670,0.8005)(0.6759,0.8003)(0.6846,0.8000)(0.6933,0.7998)(0.7018,0.7995)(0.7102,0.7992)(0.7185,0.7989)(0.7267,0.7986)(0.7347,0.7982)(0.7427,0.7978)(0.7506,0.7974)(0.7584,0.7970)(0.7661,0.7966)(0.7737,0.7961)(0.7812,0.7956)(0.7886,0.7951)(0.7960,0.7946)(0.8032,0.7941)(0.8104,0.7935)(0.8175,0.7930)(0.8245,0.7924)(0.8314,0.7918)(0.8382,0.7912)(0.8450,0.7906)(0.8517,0.7899)(0.8583,0.7892)(0.8648,0.7886)(0.8713,0.7879)(0.8777,0.7872)(0.8840,0.7865)(0.8902,0.7857)(0.8964,0.7850)(0.9026,0.7842)(0.9086,0.7835)(0.9146,0.7827)(0.9205,0.7819)(0.9264,0.7811)(0.9322,0.7802)(0.9380,0.7794)(0.9436,0.7785)(0.9493,0.7777)(0.9548,0.7768)(0.9604,0.7759)(0.9658,0.7750)(0.9712,0.7740)(0.9766,0.7731)(0.9819,0.7721)(0.9872,0.7712)(0.9924,0.7702)(0.9975,0.7692)
		};
		\addplot[smooth,blue!60!white,line width=1.5pt] plot coordinates{
	   (0.0000,0.7592)(0.0198,0.7613)(0.0392,0.7633)(0.0583,0.7653)(0.0769,0.7671)(0.0952,0.7689)(0.1132,0.7705)(0.1308,0.7721)(0.1482,0.7736)(0.1651,0.7750)(0.1818,0.7763)(0.1982,0.7775)(0.2143,0.7786)(0.2301,0.7796)(0.2456,0.7806)(0.2609,0.7815)(0.2759,0.7823)(0.2906,0.7830)(0.3051,0.7837)(0.3194,0.7842)(0.3334,0.7847)(0.3471,0.7852)(0.3607,0.7855)(0.3740,0.7858)(0.3871,0.7861)(0.4001,0.7862)(0.4128,0.7863)(0.4253,0.7864)(0.4376,0.7863)(0.4497,0.7863)(0.4616,0.7861)(0.4734,0.7859)(0.4850,0.7857)(0.4964,0.7854)(0.5076,0.7850)(0.5187,0.7846)(0.5296,0.7842)(0.5403,0.7837)(0.5509,0.7832)(0.5613,0.7826)(0.5716,0.7820)(0.5818,0.7813)(0.5918,0.7806)(0.6017,0.7798)(0.6114,0.7791)(0.6210,0.7782)(0.6304,0.7774)(0.6398,0.7765)(0.6490,0.7756)(0.6581,0.7746)(0.6670,0.7736)(0.6759,0.7726)(0.6846,0.7715)(0.6933,0.7705)(0.7018,0.7694)(0.7102,0.7682)(0.7185,0.7671)(0.7267,0.7659)(0.7347,0.7647)(0.7427,0.7634)(0.7506,0.7622)(0.7584,0.7609)(0.7661,0.7596)(0.7737,0.7583)(0.7812,0.7569)(0.7886,0.7556)(0.7960,0.7542)(0.8032,0.7528)(0.8104,0.7513)(0.8175,0.7499)(0.8245,0.7484)(0.8314,0.7469)(0.8382,0.7454)(0.8450,0.7439)(0.8517,0.7424)(0.8583,0.7408)(0.8648,0.7392)(0.8713,0.7376)(0.8777,0.7360)(0.8840,0.7344)(0.8902,0.7327)(0.8964,0.7311)(0.9026,0.7294)(0.9086,0.7277)(0.9146,0.7259)(0.9205,0.7242)(0.9264,0.7224)(0.9322,0.7206)(0.9380,0.7188)(0.9436,0.7170)(0.9493,0.7151)(0.9548,0.7133)(0.9604,0.7114)(0.9658,0.7094)(0.9712,0.7075)(0.9766,0.7055)(0.9819,0.7035)(0.9872,0.7014)(0.9924,0.6994)(0.9975,0.6972)
		
		};
		\addplot[black!80!white!60!,,only marks,mark=o,mark size=2.5pt,line width=1.5pt] plot coordinates{(0.7506,0.8149)};
		\addplot[red!60!white,only marks,mark=o,mark size=2.5pt,line width=1.5pt] plot coordinates{(0.6114,0.8010)};
		\addplot[blue!60!white,only marks,mark=o,mark size=2.5pt,line width=1.5pt] plot coordinates{(0.4253,0.7864)};
		\legend{ {\(c_{[u]}=8\)}, {\(c_{[u]}=4\)},{\(c_{[u]}=2\)}}
		\end{axis}
		\end{tikzpicture} &
		\begin{tikzpicture}[font=\footnotesize]
		\renewcommand{\axisdefaulttryminticks}{4} 
		\tikzstyle{every major grid}+=[style=densely dashed]       
		\tikzstyle{every axis y label}+=[yshift=-10pt] 
		\tikzstyle{every axis x label}+=[yshift=5pt]
		\tikzstyle{every axis legend}+=[cells={anchor=west},fill=white,
		at={(1,0.3)}, anchor=north east, font=\small]
		\begin{axis}[
		%ybar,
		width=.48\linewidth,
		height=.42\linewidth,
		%xmode=log,
		%log basis x={10},
		grid=major,
		ymajorgrids=false,
		scaled ticks=true,
		xtick={0,0.5,1},
		xlabel={\(\theta(\xi_e)\)},
		]
		\addplot[dotted,black!80!white!60!,line width=1.5pt] plot coordinates{
			(0.0000,0.8140)(0.0198,0.8149)(0.0392,0.8158)(0.0583,0.8166)(0.0769,0.8174)(0.0952,0.8181)(0.1132,0.8188)(0.1308,0.8195)(0.1482,0.8201)(0.1651,0.8207)(0.1818,0.8212)(0.1982,0.8217)(0.2143,0.8221)(0.2301,0.8226)(0.2456,0.8230)(0.2609,0.8233)(0.2759,0.8237)(0.2906,0.8240)(0.3051,0.8243)(0.3194,0.8245)(0.3334,0.8247)(0.3471,0.8249)(0.3607,0.8251)(0.3740,0.8253)(0.3871,0.8254)(0.4001,0.8255)(0.4128,0.8256)(0.4253,0.8256)(0.4376,0.8257)(0.4497,0.8257)(0.4616,0.8257)(0.4734,0.8257)(0.4850,0.8257)(0.4964,0.8257)(0.5076,0.8256)(0.5187,0.8255)(0.5296,0.8255)(0.5403,0.8254)(0.5509,0.8252)(0.5613,0.8251)(0.5716,0.8250)(0.5818,0.8248)(0.5918,0.8247)(0.6017,0.8245)(0.6114,0.8243)(0.6210,0.8241)(0.6304,0.8239)(0.6398,0.8237)(0.6490,0.8235)(0.6581,0.8233)(0.6670,0.8230)(0.6759,0.8228)(0.6846,0.8225)(0.6933,0.8223)(0.7018,0.8220)(0.7102,0.8217)(0.7185,0.8214)(0.7267,0.8211)(0.7347,0.8208)(0.7427,0.8205)(0.7506,0.8202)(0.7584,0.8199)(0.7661,0.8196)(0.7737,0.8192)(0.7812,0.8189)(0.7886,0.8185)(0.7960,0.8182)(0.8032,0.8178)(0.8104,0.8175)(0.8175,0.8171)(0.8245,0.8167)(0.8314,0.8164)(0.8382,0.8160)(0.8450,0.8156)(0.8517,0.8152)(0.8583,0.8148)(0.8648,0.8144)(0.8713,0.8140)(0.8777,0.8136)(0.8840,0.8132)(0.8902,0.8127)(0.8964,0.8123)(0.9026,0.8119)(0.9086,0.8114)(0.9146,0.8110)(0.9205,0.8105)(0.9264,0.8101)(0.9322,0.8096)(0.9380,0.8091)(0.9436,0.8087)(0.9493,0.8082)(0.9548,0.8077)(0.9604,0.8072)(0.9658,0.8067)(0.9712,0.8062)(0.9766,0.8057)(0.9819,0.8051)(0.9872,0.8046)(0.9924,0.8041)(0.9975,0.8035)
		};
		\addplot[densely dashed,red!60!white,line width=1.5pt] plot coordinates{
			(0.0000,0.7921)(0.0198,0.7936)(0.0392,0.7951)(0.0583,0.7965)(0.0769,0.7979)(0.0952,0.7992)(0.1132,0.8004)(0.1308,0.8016)(0.1482,0.8028)(0.1651,0.8039)(0.1818,0.8049)(0.1982,0.8060)(0.2143,0.8069)(0.2301,0.8078)(0.2456,0.8087)(0.2609,0.8096)(0.2759,0.8104)(0.2906,0.8111)(0.3051,0.8118)(0.3194,0.8125)(0.3334,0.8132)(0.3471,0.8138)(0.3607,0.8144)(0.3740,0.8149)(0.3871,0.8154)(0.4001,0.8159)(0.4128,0.8163)(0.4253,0.8168)(0.4376,0.8171)(0.4497,0.8175)(0.4616,0.8178)(0.4734,0.8182)(0.4850,0.8184)(0.4964,0.8187)(0.5076,0.8189)(0.5187,0.8191)(0.5296,0.8193)(0.5403,0.8195)(0.5509,0.8196)(0.5613,0.8197)(0.5716,0.8198)(0.5818,0.8199)(0.5918,0.8200)(0.6017,0.8200)(0.6114,0.8201)(0.6210,0.8201)(0.6304,0.8201)(0.6398,0.8200)(0.6490,0.8200)(0.6581,0.8199)(0.6670,0.8199)(0.6759,0.8198)(0.6846,0.8197)(0.6933,0.8196)(0.7018,0.8194)(0.7102,0.8193)(0.7185,0.8192)(0.7267,0.8190)(0.7347,0.8188)(0.7427,0.8186)(0.7506,0.8184)(0.7584,0.8182)(0.7661,0.8180)(0.7737,0.8178)(0.7812,0.8175)(0.7886,0.8173)(0.7960,0.8170)(0.8032,0.8167)(0.8104,0.8164)(0.8175,0.8162)(0.8245,0.8159)(0.8314,0.8155)(0.8382,0.8152)(0.8450,0.8149)(0.8517,0.8146)(0.8583,0.8142)(0.8648,0.8139)(0.8713,0.8135)(0.8777,0.8131)(0.8840,0.8128)(0.8902,0.8124)(0.8964,0.8120)(0.9026,0.8116)(0.9086,0.8112)(0.9146,0.8108)(0.9205,0.8103)(0.9264,0.8099)(0.9322,0.8095)(0.9380,0.8090)(0.9436,0.8086)(0.9493,0.8081)(0.9548,0.8076)(0.9604,0.8071)(0.9658,0.8067)(0.9712,0.8062)(0.9766,0.8056)(0.9819,0.8051)(0.9872,0.8046)(0.9924,0.8041)(0.9975,0.8035)
		};
		\addplot[smooth,blue!60!white,line width=1.5pt] plot coordinates{
			(0.0000,0.7592)(0.0198,0.7613)(0.0392,0.7634)(0.0583,0.7655)(0.0769,0.7675)(0.0952,0.7695)(0.1132,0.7714)(0.1308,0.7733)(0.1482,0.7751)(0.1651,0.7768)(0.1818,0.7786)(0.1982,0.7802)(0.2143,0.7819)(0.2301,0.7834)(0.2456,0.7850)(0.2609,0.7865)(0.2759,0.7879)(0.2906,0.7893)(0.3051,0.7907)(0.3194,0.7920)(0.3334,0.7932)(0.3471,0.7945)(0.3607,0.7956)(0.3740,0.7968)(0.3871,0.7979)(0.4001,0.7989)(0.4128,0.7999)(0.4253,0.8009)(0.4376,0.8018)(0.4497,0.8027)(0.4616,0.8036)(0.4734,0.8044)(0.4850,0.8052)(0.4964,0.8059)(0.5076,0.8067)(0.5187,0.8073)(0.5296,0.8080)(0.5403,0.8086)(0.5509,0.8092)(0.5613,0.8097)(0.5716,0.8102)(0.5818,0.8107)(0.5918,0.8112)(0.6017,0.8116)(0.6114,0.8120)(0.6210,0.8124)(0.6304,0.8127)(0.6398,0.8130)(0.6490,0.8133)(0.6581,0.8136)(0.6670,0.8138)(0.6759,0.8140)(0.6846,0.8142)(0.6933,0.8144)(0.7018,0.8145)(0.7102,0.8146)(0.7185,0.8147)(0.7267,0.8148)(0.7347,0.8149)(0.7427,0.8149)(0.7506,0.8149)(0.7584,0.8149)(0.7661,0.8149)(0.7737,0.8149)(0.7812,0.8148)(0.7886,0.8147)(0.7960,0.8146)(0.8032,0.8145)(0.8104,0.8144)(0.8175,0.8143)(0.8245,0.8141)(0.8314,0.8139)(0.8382,0.8137)(0.8450,0.8135)(0.8517,0.8133)(0.8583,0.8131)(0.8648,0.8128)(0.8713,0.8126)(0.8777,0.8123)(0.8840,0.8120)(0.8902,0.8117)(0.8964,0.8114)(0.9026,0.8110)(0.9086,0.8107)(0.9146,0.8103)(0.9205,0.8100)(0.9264,0.8096)(0.9322,0.8092)(0.9380,0.8088)(0.9436,0.8084)(0.9493,0.8080)(0.9548,0.8075)(0.9604,0.8071)(0.9658,0.8066)(0.9712,0.8061)(0.9766,0.8056)(0.9819,0.8051)(0.9872,0.8046)(0.9924,0.8041)(0.9975,0.8035)
		};
		\addplot[black!80!white!60!,,only marks,mark=o,mark size=2.5pt,line width=1.5pt] plot coordinates{(0.4616,0.8257)};
		\addplot[red!60!white,only marks,mark=o,mark size=2.5pt,line width=1.5pt] plot coordinates{(0.6210,0.8201)};
		\addplot[blue!60!white,only marks,mark=o,mark size=2.5pt,line width=1.5pt] plot coordinates{(0.7506,0.8149)};
		\legend{ {\(c_{[l]}=4\)}, {\(c_{[l]}=2\)},{\(c_{[l]}=1\)}}
		\end{axis}
		\end{tikzpicture}
	\end{tabular}
	\caption{Asymptotic probability of correct classification as \(\theta(\xi_e)\) varies, for \(\rho_1=\rho_2\), \(p=100\), \(\mu_1=-\mu_2=[-1,0,\ldots,0]^\trans\), \(\{C\}_{i,j}=.1^{\vert i-j\vert}\). Left: various \(c_{[u]}\) with \(c_{[l]}=1\). Right: various \(c_{[l]}\) with \(c_{[u]}=8\). Optimal values marked in circle.}
	\label{fig:theta}
\end{figure}

{\BLUE Theorem~\ref{th:statistics of fu for centered method} implies that the performance of the proposed method is controlled by both \(c_{[l]}\) and \(c_{[u]}\) (the number of labelled and unlabelled samples per dimension), as \(m(\xi)\), \(\sigma^2(\xi)\) (given by  \eqref{eq:m(xi)}, \eqref{eq:sigma2(xi)}) are dependent of \(c_{[l]}\) and \(c_{[u]}\). It is however hard to see directly a consistently increasing performance with  both \(c_{[l]}\) and \(c_{[u]}\) from these results. As a first objective of this subsection, we translate the theorem into more interpretable results. 

First, it should be pointed out that with the approach of centered similarities, the norm of the unlabelled data score vector \(f_{[u]}\) can be controlled through the adjustment of the hyperparameter \(e\), as opposed to the Laplacian regularization methods. As will be demonstrated later in this section, the norm of \(f_{[u]}\), or more precisely the norm of its deterministic part \(\mathbb{E}\{f_{[u]}\}\), directly affects how much the learning process relies on the unlabelled (versus labelled) data. With \(\mathbb{E}\{f_{[u]}\}\) given by Theorem~\ref{th:statistics of fu for centered method} for high dimensional data, we indeed note that
\begin{equation*}
\frac{\Vert \mathbb{E}\{f_{[u]}\}\Vert}{\Vert f_{[l]}\Vert+\Vert \mathbb{E}\{f_{[u]}\}\Vert}=\frac{c_{[u]}\hat m}{2c_{[l]}+c_{[u]}\hat m}+o_P(1)=\theta(\xi_e)+o_P(1)
\end{equation*}
as it can be obtained from \eqref{eq:m(xi)} that
\begin{equation*}
\theta(\xi)= \frac{c_{[u]} m(\xi)}{2c_{[l]}+c_{[u]}m(\xi)}.
\end{equation*}

In the following discussion, we shall use the variance over square mean ratio 
\begin{align}
\label{eq:r_ctr}
	r_{{\rm ctr}}\equiv\hat\sigma^2/\hat m^2
\end{align}
as the inverse performance measure for the method of centered regularization (i.e., smaller values of \(r_{{\rm ctr}}\) translate into better classification results for high dimensional data). A reorganization of the results in Theorem~\ref{th:statistics of fu for centered method} leads to the corollary below.

\begin{corollary}
	\label{cor: r_ctr}
	Under the conditions and notations of Theorem~\ref{th:statistics of fu for centered method}, and with \(r_{{\rm ctr}}\) defined in \eqref{eq:r_ctr}, we have
	\begin{equation}
	\label{eq:r_ctr approx}
	\frac{r_{{\rm ctr}}}{\rho_1\rho_2}=\frac{s(\xi_e)}{\theta^2(\xi_e)}+\frac{q(\xi_e)}{\theta^2(\xi_e)}\left[\frac{\theta^2(\xi_e)}{c_{[u]}}\left(1+\frac{r_{{\rm ctr}}}{\rho_1\rho_2}\right)+\frac{\left(1-\theta(\xi_e)\right)^2}{c_{[l]}}\right]
	\end{equation}
	where we recall \(\theta(\xi)= \frac{c_{[u]} m(\xi)}{2c_{[l]}+c_{[u]}m(\xi)}\in(0,1)\).
	
\end{corollary}
 
Equation~\eqref{eq:r_ctr approx} suggests a growing performance with more labelled or unlabelled data, as the last two terms on the right-hand side have respectively \(c_{[u]}\) and \(c_{[l]}\) in their denominators. These two terms are actually quite similar, except for  the pair of \(\theta^2(\xi_e)\) and \(\left[1-\theta(\xi_e)\right]^2\) each associated to one of them, and a factor of \(1+r_{{\rm ctr}}/\rho_1\rho_2\geq 1\) in the term with  \(c_{[u]}\). As said earlier, the quantity \(\theta(\xi_e)=c_{[u]}\hat m/(2c_{[l]}+c_{[u]}\hat m)\in(0,1)\) reflects how much the learning relies on unlabelled data. Indeed, it can be observed from \eqref{eq:r_ctr approx} that \(r_{{\rm ctr}}\) tends to be only dependent of \(c_{[l]}\) (resp., \(c_{[u]}\)) in the limit \(\theta(\xi_e)\to 0\) (resp., \(\theta(\xi_e)\to 1\)). The factor \(1+r_{{\rm ctr}}/\rho_1\rho_2\geq 1\) is related to the fact that unlabelled data are less informative than the labelled ones. According to the definition of  \(r_{{\rm ctr}}\), this factor goes to \(1\) when the scores of unlabelled data tend to deterministic values, indicating an equivalence between labelled and unlabelled data in this extreme scenario.  In a way, the factor of \(1+r_{{\rm ctr}}/\rho_1\rho_2\) quantifies how much labelled samples are more helpful than unlabelled data to the learning process.
 
 \smallskip
 
To demonstrate an effective learning from labelled and unlabelled data, we now show that, for a well-chosen \(e\), \(r_{{\rm ctr}}\) decreases with \(c_{[u]}\) and \(c_{[l]}\). Recall that the expressions of \(\theta(\xi)\), \(q(\xi)\) and \(s(\xi)\)  do not involve \(c_{[u]}\) or \(c_{[l]}\). It is then easy to see that, at some fixed \(\xi_e\), \(r_{{\rm ctr}}>0\) is a strictly decreasing function of both \(c_{[u]}\) and \(c_{[l]}\). Adding to this argument the fact that the attainable range \((0,\xi_{\sup})\) of \(\xi_e\) over \(e>0\)  is independent of \(c_{[l]}\) and only enlarges with greater \(c_{[u]}\) (as can be derived from the definition \eqref{eq:xi_sup} of \(\xi_{\sup}\)), we conclude that the performance of the proposed method consistently benefits from the addition of input data, \textit{whether labelled or unlabelled}, as formally stated in Proposition~\ref{prop:consistent SSL}. These remarks are illustrated in Figure~\ref{fig:theta}, where we plot the probability of correct classification as \(\theta(\xi_e)\) varies from \(0\) to \(1\). 
 \begin{proposition}
 	\label{prop:consistent SSL}
 	Under the conditions and notations of Corollary~\ref{cor: r_ctr}, we have that, for any \( e>0\), there exists a \(e'>0\) such that \(r_{{\rm ctr}}(c_{[l]},c_{[u]},e)>r'_{{\rm ctr}}(c'_{[l]},c'_{[u]},e')\) if \(c'_{[l]}\geq c_{[l]}\), \(c'_{[u]}\geq c_{[u]}\) and \(c'_{[l]}+c'_{[u]}>c_{[l]}+c_{[u]}\).
 \end{proposition}}

 \bigskip
 
{\BLUE Not only is the proposed method of centered regularization able to achieve an effective semi-supervised learning on high dimensional data, it does so with \textit{a labelled data learning  efficiency lower bounded by that of Laplacian regularization} (which is reduced to supervised learning in high dimensions),  and \textit{an unlabelled data learning efficiency lower bounded by that of spectral clustering}, a standard unsupervised learning algorithm on graphs. The focus of  the following  discussion is to establish this second remark, which implies the superiority of centered regularization over the methods of Laplacian regularization and spectral clustering.
 
 \medskip
 
 Recall from Theorem~\ref{th:statistics of fu for Laplacian method} that, similarly to the centered regularization, the random walk normalized Laplacian algorithm (the only one ensuring non-trivial high dimensional classification among existing Laplacian algorithms) gives also \(\tilde f_i\sim\mathcal{N}\left((-1)^k(1-\rho_k)m',\sigma^{\prime 2}\right)\) under the homoscedasticity assumption, for $m'=(2\rho_1\rho_2c_{[l]}/pc_0)(m_2-m_1)$, $\sigma^{\prime }=(2\rho_1\rho_2c_{[l]}/pc_0)\sigma_1=(2\rho_1\rho_2c_{[l]}/pc_0)\sigma_2$ with \(m_k,\sigma_k, k\in\{1,2\}\) given in Theorem~\ref{th:statistics of fu for Laplacian method}. Similarly to the definition of  \(r_{{\rm ctr}}\), we denote 
 \begin{equation}
  \label{eq:r Laplacian}
 	r_{{\rm Lap}}\equiv \sigma^{\prime 2}/m^{\prime 2}.
 \end{equation}
 
 Since \(\theta(\xi_e)\to0\) as \(\xi_e\to 0\) and \(\xi_e\to 0\) as \(e\to 0\), we obtain the following proposition from the results of Theorem~\ref{th:statistics of fu for Laplacian method} and Corollary~\ref{cor: r_ctr}.
\begin{proposition}
	Under the conditions and notations of Theorem~\ref{th:statistics of fu for Laplacian method} and Corollary~\ref{cor: r_ctr}, letting \(r_{{\rm Lap}}\) be defined by \eqref{eq:r Laplacian}, we have that 
	\begin{equation*}
	\lim_{e\to 0} r_{{\rm ctr}}=r_{{\rm Lap}}=\frac{(\mu_1-\mu_2)^\trans C(\mu_1-\mu_2)}{\Vert \mu_1-\mu_2\Vert^4}+\frac{{\rm tr}C^2}{p\Vert \mu_1-\mu_2\Vert^4\rho_1\rho_2c_{[l]}}.
	\end{equation*}
\end{proposition}
We thus remark that \textit{the performance of Laplacian regularization is retrieved by the method proposed in the present article in the limit \(e\to 0\)}.

%Furthermore, it is interesting to note that the performance of Laplacian regularization is actually related to the performance of regularized linear discriminant analysis (LDA), a standard supervised method for Gaussian mixture data under homoscedasticity. The method of regularized LDA classifies a unlabelled data \(x_i\) (not included in the training of LDA) based on the projection \(g_i=\beta^\trans x_i\), where \(\beta=(\lambda I_p+\hat C)^{-1}(\hat\mu_2-\hat\mu_1)\) with \(\hat\mu_1,\hat\mu_2\) the empirical means of the classes estimated from the labelled data, \(\hat C\) the empirical covariance and \(\lambda\) the regularization parameter. Since \(\beta\) is independent of \(x_i\) unlabelled, \(g_i\overset{\mathcal{L}}{\to}\mathcal{N}(\pm m_{{\rm LDA}}/2, \sigma_{{\rm LDA}}^2)\) for some constant \(m_{{\rm LDA}}, \sigma_{{\rm LDA}}>0\), as was demonstrated in the work of \citet{elkhalil2017large}. As \(\hat\mu_1,\hat\mu_2\) are linear combinations  of labelled data vectors, it is easy to show that \(\lim_{\lambda\to +\infty} r_{{\rm LDA}}=r_{{\rm Lap}}\), where \(r_{{\rm LDA}}\equiv \sigma_{{\rm LDA}}^2/m_{{\rm LDA}}^2\).
 
 }

\medskip

After ensuring the superiority of the new regularization method over the original approach, we now proceed to providing further guarantee on its unlabelled data learning efficiency by comparing it to the unsupervised method of spectral clustering.  

Recall that the regular graph smoothness penalty term \(Q(s)\) of a signal \(s\) can be written as \(Q(s)=s^{\trans}Ls\). In an unsupervised learning manner, we shall seek the unit-norm vector that minimizes the smoothness penalty, which is the eigenvector of \(L\) associated with the smallest eigenvalue. However, as \(Q(s)\) reaches its minimum at the clearly non-informative flat vector \(s=1_n\), the sought-for solution is provided instead by the eigenvector associated with the second smallest eigenvalue. In contrast, the updated smoothness penalty term \(\hat{Q}(s)=s^{\trans}\hat{W}s\) with centered similarities does not achieves its minimum for ``flat'' signals, and thus the eigenvector associated with the smallest eigenvalue is here a valid solution. Another important aspect is that spectral clustering based on the unnormalized Laplacian matrix \(L=D-W\) has long been known to behave unstably \citep{von2008consistency}, as opposed to the symmetric normalized Laplacian \(L_s=I_n-D^{-\frac{1}{2}}WD^{-\frac{1}{2}}\), so fair comparison should be made versus \(L_s\) rather than \(L\). 

Let us define \(d_{{\rm inter}}(v)\) as the inter-cluster distance operator that takes as input a real-valued vector \(v\) of dimension \(n\), then returns the distance between the centroids of the clusters formed by the set of points \(\{v_i\vert 1\leq i\leq n, x_i\in\mathcal{C}_k\}\), for \(k\in\{1,2\}\); and \(d_{{\rm intra}}(v)\) be the intra-cluster distance operator that returns the standard deviation within clusters. {\BLUE Namely,
\begin{align*}
&d_{{\rm inter}}(v)=\vert j_1^\trans v/n_1-j_2^\trans v/n_2\vert\\
&d_{{\rm intra}}(v)=\Vert v-(j_1^\trans v/n_1)j_1-(j_2^\trans v/n_2)j_2 \Vert/\sqrt{n}
\end{align*}
where \(j_k\in\mathbb{R}^n\) with \(k\in\{1,2\}\) is the indicator vector of class \(k\) with \([j_k]_i=1\) if \(x_i\in\mathcal{C}_k\), otherwise \([j_k]_i=0\); and \(n_k\) the number of ones in the vector \(j_k\).} As the purpose of clustering analysis is to produce clusters conforming to the intrinsic classes of data points, with low variance within a cluster and large distance between clusters, the following proposition (see the proof in the appendix) shows that the performance of the classical normalized spectral clustering, which has been studied by \citet{couillet2016kernel} under the high dimensional setting, is practically the same as the one with centered similarities on high dimensional data. 

\begin{proposition}\label{prop:spectral clustering}
Under the conditions of Theorem~\ref{th:statistics of fu for centered method}, let \(v_{{\rm Lap}}\) be the eigenvector of \(L_s\) associated with the second smallest eigenvalue, and \(v_{\rm ctr}\) the eigenvector of \(\hat{W}\) associated with the largest eigenvalue.  Then, 
\begin{equation*}
     \frac{d_{{\rm inter}}(v_{{\rm Lap}})}{d_{{\rm intra}}(v_{{\rm Lap}})}=\frac{d_{{\rm inter}}(v_{{\rm ctr}})}{d_{{\rm intra}}(v_{{\rm ctr}})}+o_P(1)
\end{equation*}
for non-trivial clustering with \(d_{{\rm inter}}(v_{{\rm Lap}})/d_{{\rm intra}}(v_{{\rm Lap}}),d_{{\rm inter}}(v_{{\rm ctr}})/d_{{\rm intra}}(v_{{\rm ctr}})= O(1)\) .
\end{proposition}

As explained before, the solution \(f_{[u]}\) of the centered similarities regularization can be expressed as \(f_{[u]}=\big(\alpha I_{n_{[u]}}-\hat{W}_{[uu]}\big)^{-1}\hat{W}_{[ul]}f_{[l]}\) for some \(\alpha>\Vert \hat{W}_{[uu]}\Vert\) {\BLUE (dependent of \(e\) as indicated in \eqref{eq:relation alpha and e})}. Clearly, as \(\alpha\downarrow\Vert \hat{W}_{[uu]}\Vert\), \(f_{[u]}\) tends to align to the eigenvector of \(\hat{W}_{[uu]}\) associated with the largest eigenvalue. {\BLUE Therefore, \textit{the performance of spectral clustering on the unlabelled data subgraph is retrieved at \(e\to+\infty\)}.}

%\smallskip
%
%Relevantly, the work of \citet{couillet2016kernel} also points out that it can happen that the eigenvector of \(L_s\) giving a non-trivial clustering is not the eigenvector associated with the second smallest eigenvalue, which is commonly used in spectral clustering. This remark applies also to the unlabelled data subgraph of \(\hat{W}\): sometimes there is an interest to let \(\alpha\) be smaller than \(\Vert \hat{W}_{[uu]}\Vert\) in order to access the non-trivial eigenvector which is not the one associated with the largest eigenvalue. The practice of relaxing the constraint on \(\alpha\) can also make sense to retrieve several non-trivial eigenvectors in presence of multiple clusters .

\bigskip

{\BLUE In summary of  the discussion in this section, we conclude that the proposed regularization method with centered similarities
\begin{itemize}
    \item recovers the high dimensional performance of Laplacian regularization at \(e\to 0\);
    \item recovers the high dimensional performance of spectral clustering at \(e\to +\infty\);
    \item accomplishes a consistent high dimensional semi-supervised learning for \(e\) appropriately set between the two extremes, thus leading to an increasing performance gain over Laplacian regularization with greater amounts of unlabelled data.
\end{itemize}}

\section{Experimentation}
\label{sec:experimentation}
The objective of this section is to provide empirical evidence to support the proposed regularization method with centered similarities, by comparing it with Laplacian regularization through simulations under and beyond the settings of the theoretical analysis. 

\subsection{Validation on Finite-Size Systems}
\label{sec:validation}
\begin{figure}
	\centering
	\begin{tabular}{cc}
	\begin{tikzpicture}[font=\footnotesize]
	\renewcommand{\axisdefaulttryminticks}{4} 
	\tikzstyle{every major grid}+=[style=densely dashed ] \tikzstyle{every axis y label}+=[yshift=-10pt] 
	\tikzstyle{every axis x label}+=[yshift=5pt]
	\tikzstyle{every axis legend}+=[cells={anchor=west},fill=white,
	at={(1,0.55)}, anchor=north east, font=\footnotesize]
		\begin{axis}[
		%ybar,
		width=.48\linewidth,
		height=.42\linewidth,
		ymin=0.71,
		ymax=0.83,
		ylabel={Accuracy},
		xlabel={\(c_{[u]}\)},
		xmajorgrids=false,
	   ymajorgrids=false,
	   scaled ticks=true,
		]
		\addplot[only marks,mark = *,mark size=1.5pt,color=red!60!white,line width=1.5pt] plot coordinates{
        (2.000000,0.802616)(4.000000,0.811420)(6.000000,0.819297)(8.000000,0.820139)(10.000000,0.822740)
	};
		\addlegendentry{{Centered, empirical}}
		\addplot[dashed,red!60!white,line width=1.5pt] plot coordinates{
		(0,0.792892)(2.000000,0.805220)(4.000000,0.812974)(6.000000,0.818130)(8.000000,0.821754)(10.000000,0.824424)
	};
		\addlegendentry{{Centered, theory}}
		\addplot[only marks,mark = triangle*,mark size=1.5pt,color=blue!60!white,line width=1.5pt] plot coordinates{
        (2.000000,0.791152)(4.000000,0.794140)(6.000000,0.795032)(8.000000,0.791706)(10.000000,0.792760)
	};
		\addlegendentry{{Laplacian, empirical}}
		\addplot[densely dashed,blue!60!white,line width=1.5pt] plot coordinates{
        (0,0.792892)(2.000000,0.792892)(4.000000,0.792892)(6.000000,0.792892)(8.000000,0.792892)(10.000000,0.792892)
	};
		\addlegendentry{{Laplacian, theory}}
		\addplot[only marks,mark = x,mark size=2.5pt,color=black!80!white!60!,line width=1.5pt] plot coordinates{
		(2.000000,0.778060)(4.000000,0.801460)(6.000000,0.810884)(8.000000,0.816627)(10.000000,0.818960)
	};
		\addlegendentry{{Spectral, empirical}}
	\addplot[densely dotted,black!80!white!60!,line width=1.5pt] plot coordinates{
        (0,0.7181)(2.000000,0.780711)(4.000000,0.800988)(6.000000,0.811089)(8.000000,0.817144)(10.000000,0.821179)
	};
		\addlegendentry{{Spectral, theory}}
	\end{axis}
	\end{tikzpicture}&
			\begin{tikzpicture}[font=\footnotesize]
		\renewcommand{\axisdefaulttryminticks}{4} 
		\tikzstyle{every major grid}+=[style=densely dashed ] \tikzstyle{every axis y label}+=[yshift=-10pt] 
		\tikzstyle{every axis x label}+=[yshift=5pt]
		\tikzstyle{every axis legend}+=[cells={anchor=west},fill=white,
		at={(0.3,0.8)}, anchor=north east, font=\small ]
		\begin{axis}[
		%ybar,
		width=.48\linewidth,
		height=.42\linewidth,
		xlabel={\(c_{[u]}\)},
		ymin=0.71,
		ymax=0.83,
		xmajorgrids=false,
	    ymajorgrids=false,
	    scaled ticks=true,
		]
		\addplot[only marks,mark = *,mark size=1.5pt,color=red!60!white,line width=1.5pt] plot coordinates{
		(2.000000,0.805680)(4.000000,0.809382)(6.000000,0.813220)(8.000000,0.816596)(10.000000,0.821040)
	};
		\addplot[only marks,mark = triangle*,mark size=1.5pt,color=blue!60!white,line width=1.5pt] plot coordinates{
        (2.000000,0.795543)(4.000000,0.793017)(6.000000,0.794157)(8.000000,0.788810)(10.000000,0.795580)
	};
		\addplot[only marks,mark = x,mark size=2.5pt,color=black!80!white!60!,line width=1.5pt] plot coordinates{
		(2.000000,0.769640)(4.000000,0.791080)(6.000000,0.800582)(8.000000,0.808135)(10.000000,0.815260)};
	    \addplot[dashed,red!60!white,line width=1.5pt] plot coordinates{
		(0,0.792117)(2.000000,0.804136)(4.000000,0.811644)(6.000000,0.816605)(8.000000,0.820073)(10.000000,0.822615)
	};
		\addplot[densely dashed,blue!60!white,line width=1.5pt] plot coordinates{
        (0,0.792117)(2.000000,0.792117)(4.000000,0.792117)(6.000000,0.792117)(8.000000,0.792117)(10.000000,0.792117)
	};
	\addplot[densely dotted,black!80!white!60!,line width=1.5pt] plot coordinates{
        (0,0.6962)(2.000000,0.768676)(4.000000,0.791767)(6.000000,0.803243)(8.000000,0.810116)(10.000000,0.814695)
	};
	\end{axis}
	\end{tikzpicture}
	\end{tabular}
	\caption{Empirical and theoretical accuracy as a function of \(c_{[u]}\) with \(c_{[l]}=2\), \(\rho_1=\rho_2\), \(p=100\),  \(-\mu_1=\mu_2=[-1,0,\ldots,0]^\trans\), \(C=I_p\) (left) or \(\{C\}_{i,j}=.1^{\vert i-j\vert}\) (right). Graph constructed with \(w_{ij}=e^{-\Vert x_i-x_j\Vert^2/p}\). Averaged over \(50000/n_{[u]}\) iterations.}
	\label{fig:Gaussian data}
\end{figure}
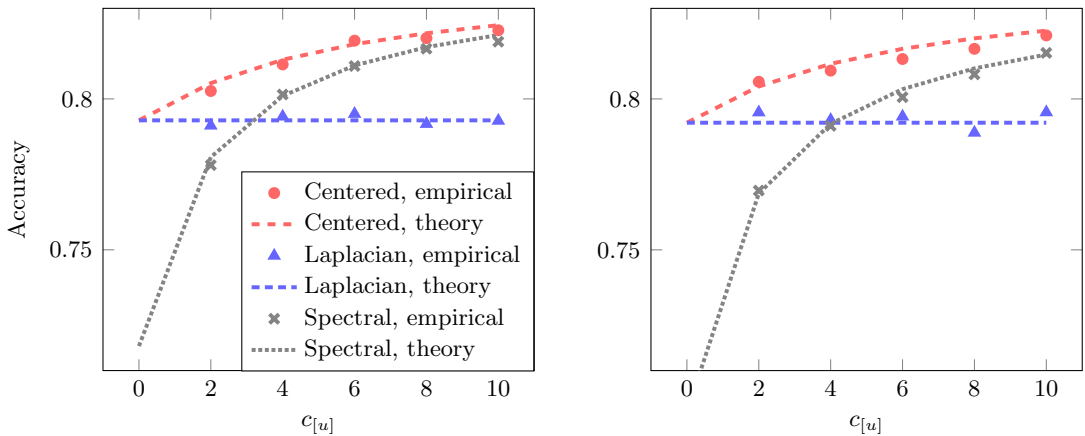

We first validate the asymptotic results of the above section on finite data sets of relatively small sizes (\(n,p\sim 100\)). Recall from Section~\ref{sec:analysis} that the asymptotic performance of Laplacian regularization and spectral clustering are recovered by centered regularization at extreme values of the hyperparameter $\theta$. In other words, the high dimensional accuracies of Laplacian regularization and spectral clustering are given by Equation~\eqref{eq:r centered method} of Theorem~\ref{th:statistics of fu for centered method}, respectively in the limit \(\theta=0\) and \(\theta=+\infty\) (when spectral clustering yields non-trivial solutions); this is how the theoretical values of both methods are computed in Figure~\ref{fig:Gaussian data}. The finite-sample results are given for the best (oracle) choice of the hyperparameter \(a\) in the generalized Laplacian matrix \(L^{(a)}=I-D^{-1-a}WD^a\) for Laplacian regularization and spectral clustering, and for the optimal (oracle) choice of the hyperparameter \(\alpha\) for centered regularization. 

Under a non-trivial Gaussian mixture model setting (see caption) with $p=100$, Figure~\ref{fig:Gaussian data} demonstrates a sharp prediction of the average empirical performance by the asymptotic analysis. As revealed by the theoretical results, the Laplacian regularization fails to learn effectively from unlabelled data, causing it to be outperformed by the purely unsupervised spectral clustering approach (for which the labelled data are treated as unlabelled ones) for sufficiently numerous unlabelled data. The performance curve of the proposed centered approach, on the other hand, is consistently above that of spectral clustering, with a growing advantage over Laplacian regularization as the number of unlabelled data increases. 

\smallskip

Figure~\ref{fig:Gaussian data} also interestingly shows that the unsupervised performance of spectral clustering is noticeably reduced when the covariance matrix of the data distribution changes from the identity matrix to a slightly disrupted model (here for \(\{C\}_{i,j}=.1^{\vert i-j\vert}\)). On the contrary, the Laplacian regularization, the high dimensional performance of which relies essentially on labelled data, is barely affected. This is explained by the different impacts labelled and unlabelled data have on the learning process, which can be understood from the theoretical results of the above section.

\subsection{Beyond the Model Assumptions}
\label{sec:beyond the model}

{\BLUE 
\begin{figure}
	\centering
	\begin{tabular}{cc}
		Digits \((3,5)\)&Digits \((7,8,9)\)\\
		\begin{tikzpicture}[font=\footnotesize]
		\renewcommand{\axisdefaulttryminticks}{4} 
		%\pgfplotsset{every major grid/.append style={densely dashed}}       
		\pgfplotsset{every axis legend/.append style={cells={anchor=west},fill=white, at={(1,1)}, anchor=north east, font=\footnotesize}}
		\begin{axis}[
		width=.45\linewidth,
		height=.37\linewidth,
		bar width=4pt,
		grid=major,
		ymajorgrids=false,
		xmax=2.3,
		ytick={0,0.1},
		ylabel={Relative frequency},
		xlabel={Normalized pairwise distances},
		]
		\addplot+[ybar,mark=none,draw=white,fill=orange!70!,area legend,opacity=0.8] coordinates{
			(0.042000,0.000000)(0.126000,0.000051)(0.210000,0.000692)(0.294000,0.003614)(0.378000,0.011413)(0.462000,0.025817)(0.546000,0.046513)(0.630000,0.070628)(0.714000,0.093853)(0.798000,0.111307)(0.882000,0.119430)(0.966000,0.117601)(1.050000,0.106826)(1.134000,0.089880)(1.218000,0.070215)(1.302000,0.050943)(1.386000,0.034321)(1.470000,0.021517)(1.554000,0.012530)(1.638000,0.006783)(1.722000,0.003379)(1.806000,0.001565)(1.890000,0.000678)(1.974000,0.000277)(2.058000,0.000113)(2.142000,0.000040)(2.226000,0.000011)
		};
		\addlegendentry{{Intra-class}}
		\addplot+[ybar,mark=none,draw=white,fill=green!80!black!80!,area legend,opacity=0.7] coordinates{
			(0.121000,0.000000)(0.203000,0.000002)(0.285000,0.000048)(0.367000,0.000502)(0.449000,0.002779)(0.531000,0.009820)(0.613000,0.024827)(0.695000,0.048944)(0.777000,0.079258)(0.859000,0.108736)(0.941000,0.129299)(1.023000,0.135327)(1.105000,0.127080)(1.187000,0.107567)(1.269000,0.082886)(1.351000,0.058453)(1.433000,0.037963)(1.515000,0.022594)(1.597000,0.012451)(1.679000,0.006374)(1.761000,0.002994)(1.843000,0.001283)(1.925000,0.000512)(2.007000,0.000195)(2.089000,0.000071)(2.171000,0.000024)
		};
		\addlegendentry{{Inter-class}}
		\end{axis}
		\end{tikzpicture}&	\begin{tikzpicture}[font=\footnotesize]
		\renewcommand{\axisdefaulttryminticks}{4} 
		%\pgfplotsset{every major grid/.append style={densely dashed}}       
		\pgfplotsset{every axis legend/.append style={cells={anchor=west},fill=white, at={(1,1)}, anchor=north east, font=\footnotesize}}
		\begin{axis}[
		width=.45\linewidth,
		height=.37\linewidth,
		bar width=4pt,
		grid=major,
		ymajorgrids=false,
		xmax=2.3,
		ytick={0,0.1},
		xlabel={Normalized pairwise distances},
		]
		\addplot+[ybar,mark=none,draw=white,fill=orange!70!,area legend,opacity=0.8] coordinates{
			%(0.042000,0.000004)(0.126000,0.000290)(0.210000,0.002476)(0.294000,0.009149)(0.378000,0.021459)(0.462000,0.038487)(0.546000,0.058330)(0.630000,0.078614)(0.714000,0.095910)(0.798000,0.107523)(0.882000,0.111318)(0.966000,0.107076)(1.050000,0.096262)(1.134000,0.081042)(1.218000,0.064043)(1.302000,0.047430)(1.386000,0.032870)(1.470000,0.021192)(1.554000,0.012668)(1.638000,0.006959)(1.722000,0.003550)(1.806000,0.001739)(1.890000,0.000858)(1.974000,0.000414)(2.058000,0.000200)(2.142000,0.000086)(2.226000,0.000036)(2.310000,0.000012)
			(0.042000,0.000013)(0.126000,0.000626)(0.210000,0.004195)(0.294000,0.013217)(0.378000,0.027954)(0.462000,0.046375)(0.546000,0.066549)(0.630000,0.085823)(0.714000,0.101422)(0.798000,0.110730)(0.882000,0.111779)(0.966000,0.104513)(1.050000,0.090937)(1.134000,0.073974)(1.218000,0.056358)(1.302000,0.040302)(1.386000,0.027065)(1.470000,0.017115)(1.554000,0.010108)(1.638000,0.005531)(1.722000,0.002829)(1.806000,0.001376)(1.890000,0.000662)(1.974000,0.000306)(2.058000,0.000145)(2.142000,0.000060)(2.226000,0.000025)(2.310000,0.000008)
			
		};
		\addplot+[ybar,mark=none,draw=white,fill=green!80!black!80!,area legend,opacity=0.7] coordinates{
			%(0.203000,0.000001)(0.289000,0.000028)(0.375000,0.000299)(0.461000,0.001908)(0.547000,0.008100)(0.633000,0.024130)(0.719000,0.053269)(0.805000,0.090172)(0.891000,0.121472)(0.977000,0.137458)(1.063000,0.135905)(1.149000,0.120660)(1.235000,0.098076)(1.321000,0.074387)(1.407000,0.052659)(1.493000,0.034917)(1.579000,0.021477)(1.665000,0.012239)(1.751000,0.006466)(1.837000,0.003271)(1.923000,0.001647)(2.009000,0.000786)(2.095000,0.000369)(2.181000,0.000177)(2.267000,0.000080)(2.353000,0.000033)(2.439000,0.000013)
			(0.125500,0.000002)(0.216500,0.000101)(0.307500,0.001050)(0.398500,0.004723)(0.489500,0.013097)(0.580500,0.027843)(0.671500,0.050626)(0.762500,0.080710)(0.853500,0.110792)(0.944500,0.130832)(1.035500,0.135434)(1.126500,0.124363)(1.217500,0.103409)(1.308500,0.078808)(1.399500,0.055391)(1.490500,0.036056)(1.581500,0.021779)(1.672500,0.012363)(1.763500,0.006620)(1.854500,0.003296)(1.945500,0.001501)(2.036500,0.000650)(2.127500,0.000292)(2.218500,0.000142)(2.309500,0.000074)(2.400500,0.000032)(2.491500,0.000010)
			
		};
		\end{axis}
		\end{tikzpicture}\\	
		\begin{tikzpicture}[font=\footnotesize]
		\renewcommand{\axisdefaulttryminticks}{4}   
		\pgfplotsset{every axis legend/.append style={cells={anchor=west},fill=white, at={(1,0.23)}, anchor=north east, font=\scriptsize}}
		\begin{axis}[
		height=0.38\linewidth,
		width=0.45\linewidth,
		grid=major,
		ymajorgrids=false,
		scaled ticks=true,
		xtick={100,300,500},
		xlabel={\(n_{[u]}\)},
		ylabel={Accuracy},
		]
		\addplot[mark = triangle*,mark size=1.5pt,color=blue!60!white,line width=1.5pt] plot coordinates{
			(90.000000,0.830522)(190.000000,0.849742)(290.000000,0.866607)(390.000000,0.872595)(490.000000,0.881186)
		};
		\addlegendentry{{Lapalcian regularization}}
		\addplot[mark = *,mark size=1.5pt,color=red!60!white,line width=1.5pt] plot coordinates{
			(90.000000,0.848656)(190.000000,0.868389)(290.000000,0.878217)(390.000000,0.888067)(490.000000,0.895031)
		};
		\addlegendentry{{Centered regularization}}
		\addplot[name path = lap_cdfminus, draw = none] plot coordinates{
			(90.000000,0.824499)(190.000000,0.845269)(290.000000,0.862558)(390.000000,0.868735)(490.000000,0.877443)
		};
		\addplot[name path = lap_cdfplus, draw = none] plot coordinates{
			(90.000000,0.836545)(190.000000,0.854215)(290.000000,0.870656)(390.000000,0.876454)(490.000000,0.884928)
		};
		\addplot [color = blue!20, draw = none] fill between[of = lap_cdfplus and lap_cdfminus];
		\addplot[name path = centered_cdfminus, draw = none] plot coordinates{
			(90.000000,0.843513)(190.000000,0.864398)(290.000000,0.874537)(390.000000,0.884664)(490.000000,0.891818)
		};
		\addplot[name path = centered_cdfplus, draw = none] plot coordinates{
			(90.000000,0.853798)(190.000000,0.872381)(290.000000,0.881898)(390.000000,0.891470)(490.000000,0.898244)
		};
		\addplot [color = red!20, draw = none] fill between[of = centered_cdfplus and centered_cdfminus];
		\end{axis}
		\end{tikzpicture}&
		\begin{tikzpicture}[font=\footnotesize]
		\renewcommand{\axisdefaulttryminticks}{4}   
		\pgfplotsset{every axis legend/.append style={cells={anchor=west},fill=white, at={(0.99,0.2)}, anchor=north east, font=\scriptsize}}
		\begin{axis}[
		height=0.38\linewidth,
		width=0.45\linewidth,
		grid=major,
		ymajorgrids=false,
		scaled ticks=true,
		xlabel={\(n_{[u]}\)},
		]
		\addplot[mark = triangle*,mark size=1.5pt,color=blue!60!white,line width=1.5pt] plot coordinates{
			(75.0000,0.8048)(150.0000,0.8239)(300.0000,0.8463)(450.0000,0.8583)(600.0000,0.8692)
		};
		\addplot[mark = *,mark size=1.5pt,color=red!60!white,line width=1.5pt] plot coordinates{
			(75.0000,0.8143)(150.0000,0.8298)(300.0000,0.8532)(450.0000,0.8668)(600.0000,0.8750)
		};
		\addplot[name path = lap_cdfminus, draw = none] plot coordinates{
			(75.000000,0.799118)(150.000000,0.819162)(300.000000,0.842220)(450.000000,0.854496)(600.000000,0.865292)
		};
		\addplot[name path = lap_cdfplus, draw = none] plot coordinates{
			(75.000000,0.810535)(150.000000,0.828665)(300.000000,0.850313)(450.000000,0.862055)(600.000000,0.873045)
		};
		\addplot [color = blue!20, draw = none] fill between[of = lap_cdfplus and lap_cdfminus];
		\addplot[name path = centered_cdfminus, draw = none] plot coordinates{
			(75.000000,0.809451)(150.000000,0.825079)(300.000000,0.849221)(450.000000,0.862926)(600.000000,0.871215)
		};
		\addplot[name path = centered_cdfplus, draw = none] plot coordinates{
			(75.000000,0.819109)(150.000000,0.834588)(300.000000,0.857186)(450.000000,0.870728)(600.000000,0.878708)
		};
		\addplot [color = red!20, draw = none] fill between[of = centered_cdfplus and centered_cdfminus];
		\end{axis}
		\end{tikzpicture}
	\end{tabular}
	\caption{Top: distribution of normalized pairwise distances \(\Vert x_i-x_j\Vert^2/\bar{\delta}\) (\(i\neq j\)) with \(\bar{\delta}\) the average of \(\Vert x_i-x_j\Vert^2\) for MNIST data. Bottom: average accuracy as a function of \(n_{[u]}\)
		with \(n_{[l]}=15\) (left) or \(n_{[l]}=10\) (right), computed over 1000 random realizations with \(99\%\) confidence intervals represented by shaded regions.}
	\label{fig:MNIST}
	\vspace{-0.2cm}
\end{figure}
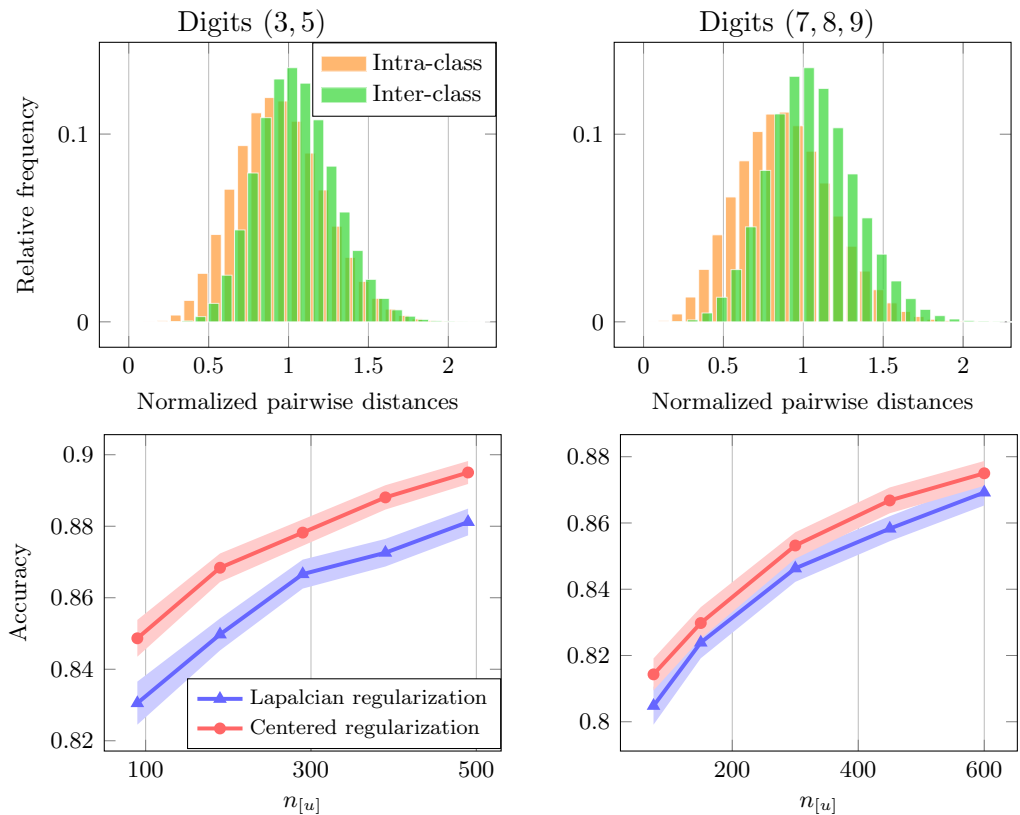

While the performance analysis of this article is placed under the Gaussianity of data vectors, we expect the proposed method to exhibit its advantage of non-negligible unlabelled data learning over the Laplacian approach in a broader context of high dimensional learning. Indeed, as discussed in Subsection~\ref{sec:problem identification}, the key element causing the unlabelled data learning inefficiency of Laplacian regularization is the negligible distinction between inter-class and intra-class similarities, induced by the \textit{distance concentration} of high dimensional data. It is important to understand that this concentration phenomenon is essentially \textit{irrespective of the Gaussianity of the data}. Proposition~\ref{prop:distance concentration} can indeed be generalized to a wider statistical model by a mere law of large numbers; this is the case for instance of all high dimensional data vectors \(x_i\) of the form \(x_i=\mu_k+C_k^{\frac{1}{2}}z_i\), for \(k\in\{1,2\}\), where \(\mu_k\in\mathbb{R}^p\), \(C_k\in\mathbb{R}^{p\times p}\) are means and covariance matrices as specified in Assumption~\ref{ass:data model} and \(z_i\in\mathbb{R}^p\) any random vector of independent elements with zero mean, unit variance and bounded fourth order moment. Beyond this model of $z_i$ with independent entries, the recent work \citep{louart2018concentration} strongly suggests that Proposition~\ref{prop:distance concentration} remains valid for the wider class of \emph{concentrated vectors} $x_i$ (i.e., satisfying a concentration of measure phenomenon \citep{LED05}), including in particular generative models of the type $x_i=F(z_i)$ for $z_i\sim\mathcal{N}(0,I_p)$ and $F:\RR^p\to\RR^p$ any $1$-Lipschitz mapping (for instance, artificial images produced by generative adversarial networks \citep{goodfellow2014generative}).

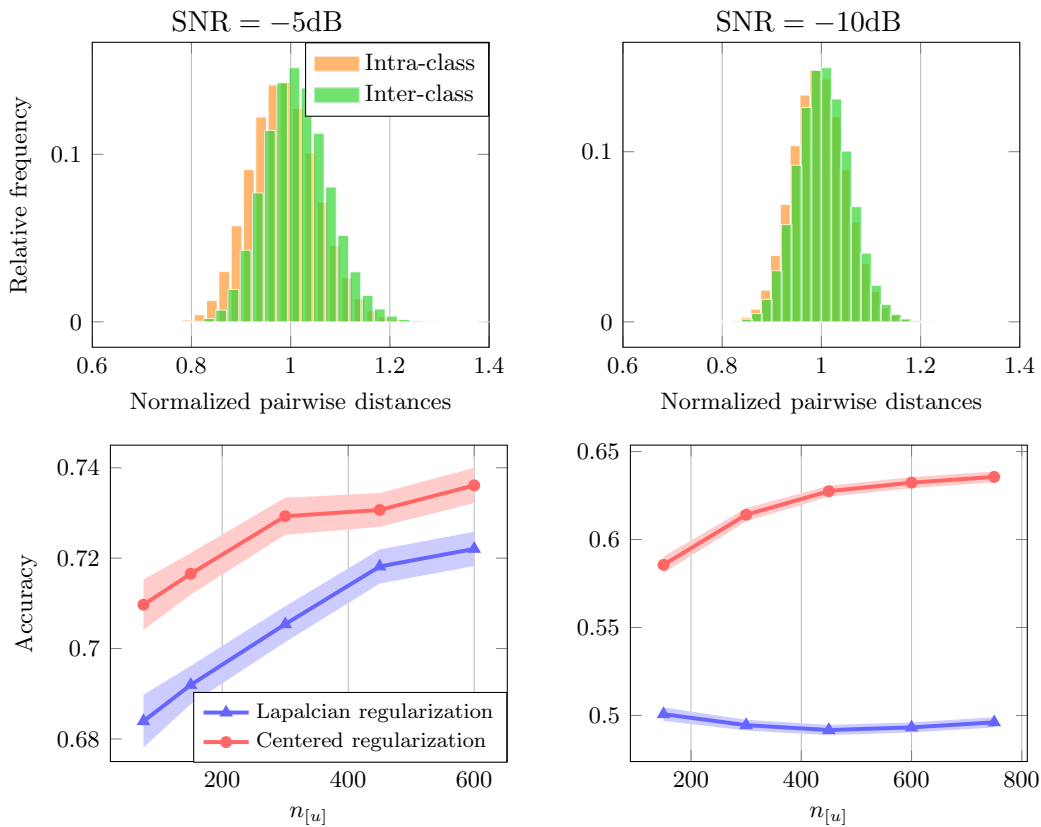
\begin{figure}
	\centering
	\begin{tabular}{cc}
		\({\rm SNR}=-5{\rm dB}\)&\({\rm SNR}=-10{\rm dB}\)\\	
		
		\begin{tikzpicture}[font=\footnotesize]
		\renewcommand{\axisdefaulttryminticks}{4} 
		%\pgfplotsset{every major grid/.append style={densely dashed}}       
		\pgfplotsset{every axis legend/.append style={cells={anchor=west},fill=white, at={(1,1)}, anchor=north east, font=\footnotesize}}
		\begin{axis}[
		width=.45\linewidth,
		height=.37\linewidth,
		bar width=4pt,
		grid=major,
		ymajorgrids=false,
		ytick={0,0.1},
		xmax=1.4,
		xmin=0.6,
		ylabel={Relative frequency},
		xlabel={Normalized pairwise distances},
		]
		\addplot+[ybar,mark=none,draw=white,fill=orange!70!,area legend,opacity=0.8] coordinates{
			(0.692400,0.000000)(0.717200,0.000002)(0.742000,0.000025)(0.766800,0.000197)(0.791600,0.001075)(0.816400,0.004236)(0.841200,0.012705)(0.866000,0.030008)(0.890800,0.057375)(0.915600,0.090890)(0.940400,0.122270)(0.965200,0.141430)(0.990000,0.143020)(1.014800,0.127417)(1.039600,0.100767)(1.064400,0.071410)(1.089200,0.045506)(1.114000,0.026312)(1.138800,0.013855)(1.163600,0.006644)(1.188400,0.002944)(1.213200,0.001197)(1.238000,0.000463)(1.262800,0.000164)(1.287600,0.000059)(1.312400,0.000018)(1.337200,0.000005)(1.362000,0.000002)(1.386800,0.000000)(1.411600,0.000000)
		};
		\addlegendentry{{Intra-class}}
		\addplot+[ybar,mark=none,draw=white,fill=green!80!black!80!,area legend,opacity=0.7] coordinates{
			(0.712300,0.000000)(0.736900,0.000001)(0.761500,0.000008)(0.786100,0.000070)(0.810700,0.000427)(0.835300,0.001979)(0.859900,0.006972)(0.884500,0.019344)(0.909100,0.042718)(0.933700,0.076973)(0.958300,0.114226)(0.982900,0.142614)(1.007500,0.151687)(1.032100,0.139542)(1.056700,0.112462)(1.081300,0.080421)(1.105900,0.051578)(1.130500,0.029855)(1.155100,0.015792)(1.179700,0.007657)(1.204300,0.003410)(1.228900,0.001419)(1.253500,0.000542)(1.278100,0.000207)(1.302700,0.000069)(1.327300,0.000021)(1.351900,0.000006)(1.376500,0.000002)(1.401100,0.000000)(1.425700,0.000000)
		};
		\addlegendentry{{Inter-class}}
		\end{axis}
		\end{tikzpicture}&
		\begin{tikzpicture}[font=\footnotesize]
		\renewcommand{\axisdefaulttryminticks}{4} 
		%\pgfplotsset{every major grid/.append style={densely dashed}}       
		\pgfplotsset{every axis legend/.append style={cells={anchor=west},fill=white, at={(0.98,0.98)}, anchor=north east, font=\footnotesize}}
		\begin{axis}[
		width=.45\linewidth,
		height=.37\linewidth,
		bar width=4pt,
		grid=major,
		ymajorgrids=false,
		ytick={0,0.1},
		xmax=1.4,
		xmin=0.6,
		xlabel={Normalized pairwise distances},
		]
		\addplot+[ybar,mark=none,draw=white,fill=orange!70!,area legend,opacity=0.8] coordinates{
			(0.729900,0.000000)(0.749700,0.000000)(0.769500,0.000004)(0.789300,0.000028)(0.809100,0.000154)(0.828900,0.000675)(0.848700,0.002476)(0.868500,0.007475)(0.888300,0.018670)(0.908100,0.039045)(0.927900,0.069024)(0.947700,0.103569)(0.967500,0.133281)(0.987300,0.148088)(1.007100,0.142751)(1.026900,0.120530)(1.046700,0.089346)(1.066500,0.058672)(1.086300,0.034247)(1.106100,0.017904)(1.125900,0.008396)(1.145700,0.003580)(1.165500,0.001389)(1.185300,0.000481)(1.205100,0.000153)(1.224900,0.000046)(1.244700,0.000013)(1.264500,0.000003)(1.284300,0.000001)(1.304100,0.000000)
		};
		\addplot+[ybar,mark=none,draw=white,fill=green!80!black!80!,area legend,opacity=0.7] coordinates{
			(0.730000,0.000000)(0.750000,0.000000)(0.770000,0.000001)(0.790000,0.000011)(0.810000,0.000073)(0.830000,0.000357)(0.850000,0.001455)(0.870000,0.004833)(0.890000,0.013160)(0.910000,0.030004)(0.930000,0.057129)(0.950000,0.092055)(0.970000,0.126010)(0.990000,0.147774)(1.010000,0.149366)(1.030000,0.130951)(1.050000,0.100337)(1.070000,0.067652)(1.090000,0.040369)(1.110000,0.021423)(1.130000,0.010165)(1.150000,0.004348)(1.170000,0.001678)(1.190000,0.000591)(1.210000,0.000186)(1.230000,0.000054)(1.250000,0.000015)(1.270000,0.000004)(1.290000,0.000001)(1.310000,0.000000)
		};
		\end{axis}
		\end{tikzpicture}\\
		\begin{tikzpicture}[font=\footnotesize]
		\renewcommand{\axisdefaulttryminticks}{4}   
		\pgfplotsset{every axis legend/.append style={cells={anchor=west},fill=white, at={(1,0.22)}, anchor=north east, font=\scriptsize}}
		\begin{axis}[
		height=0.38\linewidth,
		width=0.45\linewidth,
		ymax=0.745,
		ymin=0.675,
		scaled ticks=true,
		grid=major,
		ymajorgrids=false,
		ylabel={Accuracy},
		xlabel={\(n_{[u]}\)},
		]
		\addplot[mark = triangle*,mark size=1.5pt,color=blue!60!white,line width=1.5pt] plot coordinates{
			(75.000000,0.683933)(150.000000,0.691947)(300.000000,0.705390)(450.000000,0.718156)(600.000000,0.722057)
		};
		\addlegendentry{{Lapalcian regularization}}
		\addplot[mark = *,mark size=1.5pt,color=red!60!white,line width=1.5pt] plot coordinates{
			(75.000000,0.709707)(150.000000,0.716540)(300.000000,0.729283)(450.000000,0.730658)(600.000000,0.736100)
		};
		\addlegendentry{{Centered regularization}}
		\addplot[name path = lap_cdfminus, draw = none] plot coordinates{
			(75.000000,0.678094)(150.000000,0.687721)(300.000000,0.701432)(450.000000,0.714384)(600.000000,0.718257)
		};
		\addplot[name path = lap_cdfplus, draw = none] plot coordinates{
			(75.000000,0.689773)(150.000000,0.696173)(300.000000,0.709348)(450.000000,0.721927)(600.000000,0.725856)
		};
		\addplot [color = blue!20, draw = none] fill between[of = lap_cdfplus and lap_cdfminus];
		\addplot[name path = centered_cdfminus, draw = none] plot coordinates{
			(75.000000,0.704133)(150.000000,0.711939)(300.000000,0.725199)(450.000000,0.726934)(600.000000,0.732203)
		};
		\addplot[name path = centered_cdfplus, draw = none] plot coordinates{
			(75.000000,0.715281)(150.000000,0.721141)(300.000000,0.733368)(450.000000,0.734381)(600.000000,0.739997)
		};
		\addplot [color = red!20, draw = none] fill between[of = centered_cdfplus and centered_cdfminus];
		\end{axis}
		\end{tikzpicture}&	 
		\begin{tikzpicture}[font=\footnotesize]
		\renewcommand{\axisdefaulttryminticks}{4}   
		\pgfplotsset{every axis legend/.append style={cells={anchor=west},fill=white, at={(0.99,0.42)}, anchor=north east, font=\scriptsize}}
		\begin{axis}[
		height=0.38\linewidth,
		width=0.45\linewidth,
		scaled ticks=true,
		grid=major,
		ymajorgrids=false,
		xlabel={\(n_{[u]}\)},
		]
		\addplot[mark = triangle*,mark size=1.5pt,color=blue!60!white,line width=1.5pt] plot coordinates{
			(150.000000,0.500727)(300.000000,0.494487)(450.000000,0.491707)(600.000000,0.493205)(750.000000,0.496107)
		};
		\addplot[mark = *,mark size=1.5pt,color=red!60!white,line width=1.5pt] plot coordinates{
			(150.000000,0.585600)(300.000000,0.614087)(450.000000,0.627471)(600.000000,0.632333)(750.000000,0.635585)
		};
		\addplot[name path = lap_cdfminus, draw = none] plot coordinates{
			(150.000000,0.496897)(300.000000,0.491453)(450.000000,0.488791)(600.000000,0.490354)(750.000000,0.493201)
		};
		\addplot[name path = lap_cdfplus, draw = none] plot coordinates{
			(150.000000,0.504556)(300.000000,0.497520)(450.000000,0.494622)(600.000000,0.496056)(750.000000,0.499012)
		};
		\addplot [color = blue!20, draw = none] fill between[of = lap_cdfplus and lap_cdfminus];
		\addplot[name path = centered_cdfminus, draw = none] plot coordinates{
			(150.000000,0.580929)(300.000000,0.610365)(450.000000,0.624361)(600.000000,0.629196)(750.000000,0.632506)
		};
		\addplot[name path = centered_cdfplus, draw = none] plot coordinates{
			(150.000000,0.590271)(300.000000,0.617809)(450.000000,0.630582)(600.000000,0.635471)(750.000000,0.638664)
		};
		\addplot [color = red!20, draw = none] fill between[of = centered_cdfplus and centered_cdfminus];
		\end{axis}
		\end{tikzpicture}
	\end{tabular}
	\caption{Top: distribution of normalized pairwise distances \(\Vert x_i-x_j\Vert^2/\bar{\delta}\) (\(i\neq j\)) with \(\bar{\delta}\) the average of \(\Vert x_i-x_j\Vert^2\) for noisy MNIST data (7,8,9).
		%{\color{red} with additive noise (left: $5$dBm noise, right: $10$dBm noise)}. Xiaoyi: why $5$dBm and $10$dBm?
		Bottom: average accuracy as a function of \(n_{[u]}\)
		with \(n_{[l]}=15\), computed over 1000 random realizations with \(99\%\) confidence intervals represented by shaded regions.}
	\label{fig:MNIST noised}
	\vspace{-0.2cm}
\end{figure}

\smallskip

The main objective of this subsection is to provide an actual sense of how the Laplacian regularization approach and the proposed method behave under \textit{different levels of distance concentration}. We first give here, as a real-life example, simulations on datasets from the standard MNIST database of handwritten digits \citep{lecun1998mnist}, which are depicted in Figures~\ref{fig:MNIST}--\ref{fig:MNIST noised}.}

\smallskip

For a fair comparison of Laplacian and centered regularizations, the results displayed here are obtained on their respective best performing graphs, selected among the \(k-\)nearest neighbors graphs (which were observed to yield very competitive performance on MNIST data) with various numbers of neighbors \(k=\{2^1,\ldots,2^q\}\), for \(q\) the largest integer such that \(2^q<n\). The hyperparameters of the Laplacian and centered regularization approaches are set optimally within the admissible range.\footnote{Specifically, the hyperparameter \(a\) of Laplacian regularization is searched among the values from \(-2\) to \(0\) with a step of \(0.02\), and the hyperparameter \(\alpha\) of centered regularization within the grid \(\alpha=(1+10^t)\Vert \hat{W}_{[uu]}\Vert\) where \(t\) varies from \(-3\) to \(3\) with a step of \(0.1\). The results outside these ranges are observed to be non-competitive.} It worth pointing out that the popular KNN graphs, constructed by letting \(w_{ij}=1\) if data points \(x_i\) or \(x_j\) is among the \(k\) nearest (\(k\) being the parameter to be set beforehand) to the other data point, and \(w_{ij}=0\) if not, are not covered by the present analytic framework. Our study only deals with graphs where \(w_{ij}\) is exclusively determined by the distance between \(x_i\) and \(x_j\), while in the KNN graphs, \(w_{ij}\) is dependent of all pairwise distances of the whole data sets. Nonetheless, KNN graphs evidently suffer the same problem of distance concentration, for they are still based on the distances between data points. It is thus natural to expect that the proposed centering procedure may also be advantageous on KNN graphs.

Figure~\ref{fig:MNIST} shows that high classification accuracy is easily obtained on MNIST data, even with the classical Laplacian approach. However, it exhibits an lower learning efficiency compared to the proposed method. {\BLUE We also find that the benefit of the proposed algorithm is more perceptible on the binary classification task displayed on the left side of Figure~\ref{fig:MNIST}  than the multiclassification task on the right side, for which the difference between inter-class and intra-class distances is more apparent. This suggests that the advantage of the proposed method is more related to a subtle distinction between inter-class and intra-class distances than to the number of classes.}

{\BLUE As further evidence, Figure~\ref{fig:MNIST noised} presents situations where the learning problem becomes more challenging in the presence of additive noise. Understandably, the distance concentration phenomenon is more acute in this noise-corrupted setting, causing more subtle distinction between inter-class and intra-class distances. As a result, the performance gain generated by the proposed method should be more significant, according to our discussion at the beginning of this subsection. This is corroborated by Figure~\ref{fig:MNIST noised}, where larger performance gains are observed for the muticlassification task on the right side of Figure~\ref{fig:MNIST}. Moreover, on the right display of Figure~\ref{fig:MNIST noised}, where the similarity information is seriously disrupted by the additive noise, we observe the anticipated saturation effect when increasing $n_{[u]}$ for Laplacian regularization, in contrast to the growing performance of the proposed approach. This suggests, in conclusion, that regularization with centered similarities has a competitive, if not superior, performance in various situations, and yields particularly significant performance gains when the distinction between intra-class and inter-class similarities is quite subtle.}

\smallskip

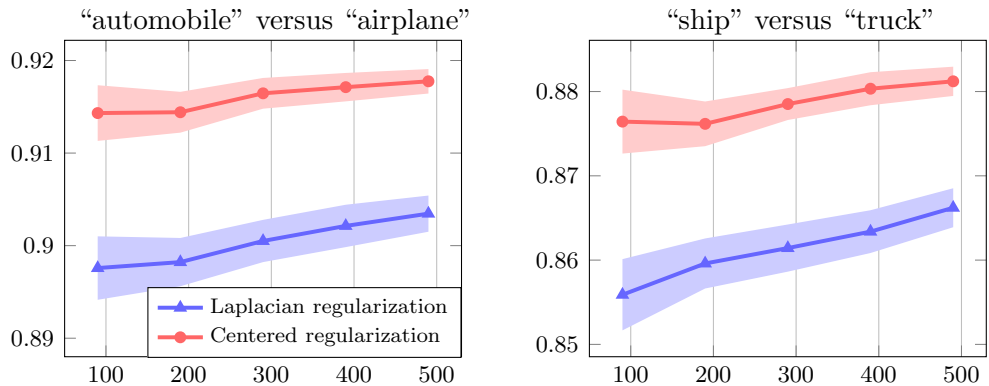
\begin{figure}
	\centering
	\begin{tabular}{cc}
	\hspace*{1cm} ``automobile'' versus ``airplane'' &\hspace*{1cm} ``ship'' versus ``truck'' \\
		\begin{tikzpicture}[font=\footnotesize]
	\renewcommand{\axisdefaulttryminticks}{4}   
	\pgfplotsset{every axis legend/.append style={cells={anchor=west},fill=white, at={(1,0.22)}, anchor=north east, font=\scriptsize}}
	\begin{axis}[
	height=0.38\linewidth,
	width=0.45\linewidth,
	scaled ticks=true,
	ymin=0.888,
	grid=major,
	ymajorgrids=false,
	]
	\addplot[mark = triangle*,mark size=1.5pt,color=blue!60!white,line width=1.5pt] plot coordinates{
    (90.000000,0.897578)(190.000000,0.898221)(290.000000,0.900503)(390.000000,0.902131)(490.000000,0.903457)
	};
	\addlegendentry{{Laplacian regularization}}
     \addplot[mark = *,mark size=1.5pt,color=red!60!white,line width=1.5pt] plot coordinates{
    (90.000000,0.914322)(190.000000,0.914411)(290.000000,0.916455)(390.000000,0.917121)(490.000000,0.917753)
	};
	\addlegendentry{{Centered regularization}}
	 \addplot[name path = lap_cdfminus, draw = none] plot coordinates{
    (90.000000,0.894151)(190.000000,0.895614)(290.000000,0.898222)(390.000000,0.899843)(490.000000,0.901503)
    };
    \addplot[name path = lap_cdfplus, draw = none] plot coordinates{
    (90.000000,0.901004)(190.000000,0.900828)(290.000000,0.902784)(390.000000,0.904419)(490.000000,0.905411)
    };
     \addplot [color = blue!20, draw = none] fill between[of = lap_cdfplus and lap_cdfminus];
    \addplot[name path = centered_cdfminus, draw = none] plot coordinates{
    (90.000000,0.911315)(190.000000,0.912207)(290.000000,0.914795)(390.000000,0.915586)(490.000000,0.916432)
    };
    \addplot[name path = centered_cdfplus, draw = none] plot coordinates{
    (90.000000,0.917329)(190.000000,0.916615)(290.000000,0.918115)(390.000000,0.918655)(490.000000,0.919074)
    };
    \addplot [color = red!20, draw = none] fill between[of = centered_cdfplus and centered_cdfminus];
	\end{axis}
	\end{tikzpicture}&	\begin{tikzpicture}[font=\footnotesize]
	\renewcommand{\axisdefaulttryminticks}{4}   
	\pgfplotsset{every axis legend/.append style={cells={anchor=west},fill=white, at={(0.99,0.42)}, anchor=north east, font=\scriptsize}}
	\begin{axis}[
	height=0.38\linewidth,
		width=0.45\linewidth,
	scaled ticks=true,
	grid=major,
	ymajorgrids=false,
	]
	\addplot[mark = triangle*,mark size=1.5pt,color=blue!60!white,line width=1.5pt] plot coordinates{
    (90.000000,0.855889)(190.000000,0.859600)(290.000000,0.861428)(390.000000,0.863379)(490.000000,0.866212)
	};
	\addplot[mark = *,mark size=1.5pt,color=red!60!white,line width=1.5pt] plot coordinates{
    (90.000000,0.876433)(190.000000,0.876168)(290.000000,0.878517)(390.000000,0.880336)(490.000000,0.881214)
	};
	 \addplot[name path = lap_cdfminus, draw = none] plot coordinates{
    (90.000000,0.851664)(190.000000,0.856631)(290.000000,0.858631)(390.000000,0.860842)(490.000000,0.863901)  
    };
    \addplot[name path = lap_cdfplus, draw = none] plot coordinates{
    (90.000000,0.860114)(190.000000,0.862569)(290.000000,0.864225)(390.000000,0.865917)(490.000000,0.868524)
    };
     \addplot [color = blue!20, draw = none] fill between[of = lap_cdfplus and lap_cdfminus];
    \addplot[name path = centered_cdfminus, draw = none] plot coordinates{
    (90.000000,0.872641)(190.000000,0.873517)(290.000000,0.876625)(390.000000,0.878376)(490.000000,0.879479)
    };
    \addplot[name path = centered_cdfplus, draw = none] plot coordinates{
    (90.000000,0.880226)(190.000000,0.878820)(290.000000,0.880409)(390.000000,0.882296)(490.000000,0.882950)
    };
    \addplot [color = red!20, draw = none] fill between[of = centered_cdfplus and centered_cdfminus];
	\end{axis}
	\end{tikzpicture}
	\end{tabular}
\caption{Average accuracy on two-class Cifar10 data as a function of \(n_{[u]}\)
with \(n_{[l]}=10\), computed over 1000 random realizations with \(99\%\) confidence intervals represented by shaded regions.}
\label{fig:Cifar}
\end{figure}

{\BLUE To further test the proposed method on challenging real-world datasets, we also compare} the Laplacian and centered similarities methods on the popular Cifar10 database \citep{krizhevsky2014cifar}. To obtain meaningful results, the data went through a feature extraction step using the standard pre-trained ResNet-50 network \citep{he2016deep}. Other experimental settings are the same as for the above MNIST data. The simulations are reported in Figure~\ref{fig:Cifar}, {\BLUE where the findings support again the use of the proposed method.}

\section{Further Discussion}
\label{sec:further discussion}

{\BLUE As further discussion, we start this section by presenting other graph-based semi-supervised learning methods and explaining how they relate to the regularization approaches investigated in this paper. To  evaluate the ability of these SSL methods to exploit optimally the information in partially labelled data sets, we use the recent results of \cite{lelarge2019asymptotic} as a reference point, where the best achievable semi-supervised learning performance on high dimensional Gaussian mixture data with identity covariance matrices was characterized. The proposed centered regularization method is found to have a remarkable advantage over other graph-based SSL methods for reaching the optimal performance. We test also on sparse graphs generated from the stochastic block model. These simulations, where we can control the informativeness of the local geometry of the graph, will provide additional empirical support for the proposed method from another perspective.}

\subsection{Related Methods}
\label{sec:related methods}

{\BLUE As multiple times emphasized, the focus of the article is to promote the usage of centered similarities in graph regularization for semi-supervised learning. This fundamental idea can also be embedded into more involved graph regularization methods, such as iterated approaches. In parallel to the graph regularization methods, there also exists an alternative approach which uses the spectral information of the graph matrix instead of optimizing the graph smoothness. We briefly discuss these related methods here.}

\subsubsection{Higher Order Regularization}

{\BLUE 
The method of semi-supervised Laplacian regularization can also be problematic outside the high dimensional regime discussed in this article.  The earlier work of \citet{nadler2009semi} showed that unlabelled data scores \(f_i\) concentrate around the same value (i.e., \(f_i=c+o(1)\) for some constant \(c\)) in the limit where the number of unlabelled samples is exceedingly large compared to that of labelled ones (i.e., \(n_{[u]}/n_{[l]}\to\infty\)).  The follow-up works \cite{alamgir2011phase,zhou2011semi,bridle2013p,kyng2015algorithms,el2016asymptotic} advocated the usage of higher order regularization techniques to address the problem of flat scores under the same setting of \cite{nadler2009semi}. Among these techniques, the  method of iterated Laplacian regularization, which consists in using the powers of Laplacian matrices for constructing high order regularizers \(f^\trans L^m f\) of graph smoothness, tends to highly competitive classification results \citep{zhou2011semi}.

\smallskip

While bringing into light this important phenomenon of `flat' unlabelled data scores, the analysis of \cite{nadler2009semi}, unlike that of \cite{mai2018random}, did not clarify why non-trivial classification is still empirically observed to be achievable and how the classification performance is affected. Remarkably, the analysis of \citet{mai2018random} also pointed out that, in high dimensions, the phenomenon of flat unlabelled data scores occurs even when the number of unlabelled samples is comparable to that of labelled ones. As can be easily deduced from our study, the problem of flat unlabelled scores is addressed by the centered regularization method in the more challenging setting of high dimensional learning. In terms of performance guarantees, as high order regularization techniques include the basic Laplacian regularization as a special case, they are guaranteed to perform no worse than Laplacian algorithms. However, it is not clear how they compare to the unsupervised performance of spectral clustering. Finally, it should be emphasized that the use of centered similarities is not in conflict with the approach of high order regularization. Future studies can be envisioned to further improve the performance by combining these two ideas.

}

\subsubsection{Eigenvector-Based Method}

{\BLUE Aside from graph regularization methods, another popular graph-based semi-supervised approach exists which takes advantage of the spectral information of Laplacian matrices \citep{belkin2003using,belkin2004semi}. Rather than regularizing \(f\) over the graph, this method computes first the eigenmap of Laplacian matrices, then uses a certain number \(s\) of eigenvectors \(E=[e_1,\ldots,e_s]\) associated with the smallest eigenvalues to build a linear subspace and search within this space for an \(f\) which minimizes \(\Vert f_{[l]}-y_{[l]}\Vert\). By the method of least squares, \(f=Ea\) with \(a=(E_{[l]}^\trans E_{[l]})^{-1}E_{[l]}^\trans y_{[l]} \).  
		
As an advantage of using the spectral information, this eigenvector-based method is guaranteed to achieve at least the performance of spectral clustering, as opposed to the Laplacian regularization approach. On the other hand, the regularization approach does not have a performance which depends crucially on how well the class signal is captured by a small number of eigenvectors, as it uses the graph matrix as a whole. Another benefit of the graph regularization approach is that it can be easily incorporated into other algorithms involving optimization as an additional term in the loss function (e.g., Laplacian SVMs). With our proposed algorithm using centered similarities, a consistent learning of unlabelled data, related to the performance of spectral clustering, can also be achieved by the graph regularization approach. Moreover, the proposed method has a theoretically-proven efficient learning of labelled data which is absent in the eigenvector-based method. }

\subsection{Optimal Performance on Isotropic Gaussian Data of High Dimensionality}
\label{sec:isotropic data}

{\BLUE A very recent work of \citet{lelarge2019asymptotic} has established the optimal performance of semi-supervised learning on a high dimensional Gaussian mixture data model  \(\mathcal{N}(\pm\mu,I_p)\), with identity covariance matrices.\footnote{To the authors' knowledge, more general results (e.g., with arbitrary covariance matrices) are currently out-of-reach.} In this work, a method of Bayesian estimation is identified as the one achieving the optimal performance. However, as pointed out by the authors, this method is computationally expensive except on fully labelled datasets and approximations are needed for practical usage.

By comparing the results of this work with our performance analysis in Section~\ref{sec:analysis}, we find that the method of centered regularization achieves an optimal performance on fully labelled datasets and a nearly optimal one on partially labelled sets.\footnote{We refer to Appendix~\ref{app:optimal performance on isotropic data} for some theoretical details.} Numerical results are given in Figure~\ref{fig:optimal performance on isotropic data}, where the classification accuracy of the centered regularization method, computed from Theorem~\ref{th:statistics of fu for centered method} and maximized over the hyperparameter \(e\), is observed to be extremely close to the optimal performance provided by \citet{lelarge2019asymptotic}. Hence, the centered regularization method can be used as a computationally efficient alternative to the Bayesian approach which yields the best achievable performance. In contrast, other graph-based semi-supervised learning algorithms are much less effective in reaching the optimal performance, as can be observed from Figure~\ref{fig: graph-based SSL algorithms on isotropic Gaussian mixture data}.

We remark also that the iterated Laplacian regularization appears to be comparably less efficient in exploiting unlabelled data and so is the eigenvector-based method in learning from labelled data. As can be observed in Figure~\ref{fig: graph-based SSL algorithms on isotropic Gaussian mixture data}, the iterated Laplacian regularization falls notably short of approaching the optimal performance when the value of \(m\) yielding the highest accuracy is further away from \(1\) (scenarios corresponding to the blue curves in the figure). Since we retrieve the standard Laplacian regularization at \(m=1\), which gives the optimal performance in the absence of unlabelled data, the performance gain yielded by the iterated Laplacian regularization over the Laplacian method is mainly brought by the utilization of unlabelled data at higher \(m\). However, as demonstrated in Figure~\ref{fig: graph-based SSL algorithms on isotropic Gaussian mixture data}, the utilization of unlabelled data at higher \(m\) is unsatisfactory in allowing the method to reach the optimal semi-supervised learning performance. Since the eigenvector-based approach is reduced to the purely unsupervised method of spectral clustering at \(s=1\), the same remark can be made with respect to its labelled data learning efficiency.

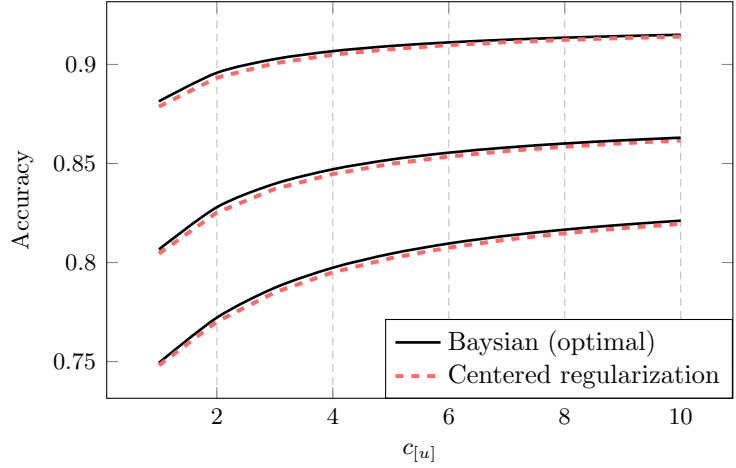
\begin{figure}
	\centering
	 \begin{tikzpicture}[font=\footnotesize]
		\renewcommand{\axisdefaulttryminticks}{4} 
		\tikzstyle{every major grid}+=[style=densely dashed ] \tikzstyle{every axis y label}+=[yshift=-10pt] 
		\tikzstyle{every axis x label}+=[yshift=5pt]
		\tikzstyle{every axis legend}+=[cells={anchor=west},fill=white,
		at={(1,0.2)}, anchor=north east, font=\small]
		\begin{axis}[
		%ybar,
		width=.65\linewidth,
		height=.45\linewidth,
		%xmode=log,
		%log basis x={10},
		grid=major,
		ymajorgrids=false,
		scaled ticks=true,
		ylabel={Accuracy},
		xlabel={\(c_{[u]}\)},
		]
		\addplot[smooth,black,line width=1pt] plot coordinates{
		(1.000000,0.881362)(2.000000,0.895790)(3.000000,0.902703)(4.000000,0.906702)(5.000000,0.909298)(6.000000,0.911117)(7.000000,0.912461)(8.000000,0.913493)(9.000000,0.914310)(10.000000,0.914974)
		};
	   \addlegendentry{{Baysian (optimal)}}
	   	\addplot[dashed,red!60!white,line width=1.5pt] plot coordinates{
	   	(1.000000,0.804518)(2.000000,0.825262)(3.000000,0.837193)(4.000000,0.844716)(5.000000,0.849839)(6.000000,0.853537)(7.000000,0.856325)(8.000000,0.858499)(9.000000,0.860242)(10.000000,0.861668)
	   };
   	\addlegendentry{{Centered regularization}}
		\addplot[smooth,black,line width=1pt] plot coordinates{
		(1.000000,0.806616)(2.000000,0.827935)(3.000000,0.839784)(4.000000,0.847082)(5.000000,0.851974)(6.000000,0.855465)(7.000000,0.858076)(8.000000,0.860098)(9.000000,0.861711)(10.000000,0.863027)
		   };
		\addplot[smooth,black,line width=1pt] plot coordinates{
		(1.000000,0.749337)(2.000000,0.772213)(3.000000,0.787292)(4.000000,0.797387)(5.000000,0.804448)(6.000000,0.809616)(7.000000,0.813540)(8.000000,0.816618)(9.000000,0.819089)(10.000000,0.821117)
        };
		\addplot[dashed,red!60!white,line width=1.5pt] plot coordinates{
		(1.000000,0.878752)(2.000000,0.893328)(3.000000,0.900615)(4.000000,0.904929)(5.000000,0.907768)(6.000000,0.909775)(7.000000,0.911268)(8.000000,0.912420)(9.000000,0.913337)(10.000000,0.914083)	
		};
	
	 \addplot[dashed,red!60!white,line width=1.5pt] plot coordinates{
	    (1.000000,0.748169)(2.000000,0.770210)(3.000000,0.785003)(4.000000,0.795083)(5.000000,0.802245)(6.000000,0.807546)(7.000000,0.811610)(8.000000,0.814816)(9.000000,0.817407)(10.000000,0.819541)
        };

		\end{axis}
		\end{tikzpicture} 
	\caption{Asymptotic accuracy on isotropic Gaussian mixture data. Performance curves as a function of \(c_{[u]}\) with \(c_{[l]}=1/2\), for (from top to bottom) \(\Vert\mu\Vert^2=2,4/3,\text{or } 1\). }
	\label{fig:optimal performance on isotropic data}
\end{figure}

\begin{figure}
	\centering
	\begin{tabular}{ccc}
	$\quad\quad\quad$Centered & Iterated & Eigenvector-based\\
		\begin{tikzpicture}[font=\footnotesize]
	\renewcommand{\axisdefaulttryminticks}{4} 
	\tikzstyle{every major grid}+=[style=densely dashed ] \tikzstyle{every axis y label}+=[yshift=-10pt] 
	\tikzstyle{every axis x label}+=[yshift=5pt]
	\tikzstyle{every axis legend}+=[cells={anchor=west},fill=white,
	at={(1,0.3)}, anchor=north east, font=\scriptsize]
	\begin{axis}[
	%ybar,
	width=.33\linewidth,
	height=.3\linewidth,
	ymin=0.74,
	ymax=0.82,
	tick pos=left,
	ymajorgrids=false,
	scaled ticks=true,
	ylabel={Accuracy},
	]
	
	\addplot[smooth,color=blue!60!white,line width=1pt] plot coordinates{
		
	(-3.00,0.7910)(-2.95,0.7911)(-2.90,0.7911)(-2.85,0.7912)(-2.80,0.7913)(-2.75,0.7914)(-2.70,0.7914)(-2.65,0.7915)(-2.60,0.7916)(-2.55,0.7917)(-2.50,0.7918)(-2.45,0.7919)(-2.40,0.7921)(-2.35,0.7922)(-2.30,0.7924)(-2.25,0.7926)(-2.20,0.7929)(-2.15,0.7932)(-2.10,0.7934)(-2.05,0.7937)(-2.00,0.7940)(-1.95,0.7944)(-1.90,0.7948)(-1.85,0.7951)(-1.80,0.7956)(-1.75,0.7961)(-1.70,0.7967)(-1.65,0.7973)(-1.60,0.7980)(-1.55,0.7985)(-1.50,0.7993)(-1.45,0.8000)(-1.40,0.8008)(-1.35,0.8015)(-1.30,0.8023)(-1.25,0.8032)(-1.20,0.8040)(-1.15,0.8045)(-1.10,0.8050)(-1.05,0.8056)(-1.00,0.8060)(-0.95,0.8064)(-0.90,0.8066)(-0.85,0.8068)(-0.80,0.8069)(-0.75,0.8067)(-0.70,0.8062)(-0.65,0.8053)(-0.60,0.8043)(-0.55,0.8034)(-0.50,0.8022)(-0.45,0.8009)(-0.40,0.7993)(-0.35,0.7977)(-0.30,0.7960)(-0.25,0.7942)(-0.20,0.7924)(-0.15,0.7904)(-0.10,0.7884)(-0.05,0.7864)(0.00,0.7845)(0.05,0.7827)(0.10,0.7809)(0.15,0.7791)(0.20,0.7773)(0.25,0.7756)(0.30,0.7742)(0.35,0.7727)(0.40,0.7714)(0.45,0.7703)(0.50,0.7692)(0.55,0.7682)(0.60,0.7672)(0.65,0.7663)(0.70,0.7656)(0.75,0.7649)(0.80,0.7643)(0.85,0.7637)(0.90,0.7631)(0.95,0.7626)(1.00,0.7622)(1.05,0.7618)(1.10,0.7615)(1.15,0.7612)(1.20,0.7610)(1.25,0.7607)(1.30,0.7605)(1.35,0.7602)(1.40,0.7600)(1.45,0.7599)(1.50,0.7598)(1.55,0.7596)(1.60,0.7595)(1.65,0.7595)(1.70,0.7593)(1.75,0.7592)(1.80,0.7592)(1.85,0.7591)(1.90,0.7590)(1.95,0.7590)(2.00,0.7590)(2.05,0.7589)(2.10,0.7589)(2.15,0.7589)(2.20,0.7588)(2.25,0.7588)(2.30,0.7588)(2.35,0.7588)(2.40,0.7588)(2.45,0.7587)(2.50,0.7588)(2.55,0.7587)(2.60,0.7587)(2.65,0.7587)(2.70,0.7587)(2.75,0.7587)(2.80,0.7587)(2.85,0.7586)(2.90,0.7586)(2.95,0.7586)(3.00,0.7586)
	
	};
\addlegendentry{{\(c_{[l]}=1\)}}
	\addplot[densely dashed,color=green!80!black!90!,line width=1pt] plot coordinates{
		
	(-3.00,0.7568)(-2.95,0.7570)(-2.90,0.7573)(-2.85,0.7575)(-2.80,0.7576)(-2.75,0.7578)(-2.70,0.7581)(-2.65,0.7584)(-2.60,0.7587)(-2.55,0.7590)(-2.50,0.7594)(-2.45,0.7599)(-2.40,0.7603)(-2.35,0.7608)(-2.30,0.7614)(-2.25,0.7620)(-2.20,0.7625)(-2.15,0.7632)(-2.10,0.7642)(-2.05,0.7652)(-2.00,0.7661)(-1.95,0.7669)(-1.90,0.7680)(-1.85,0.7693)(-1.80,0.7705)(-1.75,0.7717)(-1.70,0.7733)(-1.65,0.7748)(-1.60,0.7763)(-1.55,0.7780)(-1.50,0.7798)(-1.45,0.7816)(-1.40,0.7835)(-1.35,0.7855)(-1.30,0.7873)(-1.25,0.7892)(-1.20,0.7912)(-1.15,0.7933)(-1.10,0.7953)(-1.05,0.7975)(-1.00,0.7996)(-0.95,0.8016)(-0.90,0.8034)(-0.85,0.8051)(-0.80,0.8067)(-0.75,0.8084)(-0.70,0.8101)(-0.65,0.8112)(-0.60,0.8126)(-0.55,0.8136)(-0.50,0.8145)(-0.45,0.8154)(-0.40,0.8160)(-0.35,0.8163)(-0.30,0.8166)(-0.25,0.8166)(-0.20,0.8168)(-0.15,0.8166)(-0.10,0.8163)(-0.05,0.8161)(0.00,0.8158)(0.05,0.8152)(0.10,0.8146)(0.15,0.8142)(0.20,0.8138)(0.25,0.8134)(0.30,0.8129)(0.35,0.8125)(0.40,0.8121)(0.45,0.8117)(0.50,0.8112)(0.55,0.8107)(0.60,0.8105)(0.65,0.8102)(0.70,0.8098)(0.75,0.8095)(0.80,0.8092)(0.85,0.8090)(0.90,0.8088)(0.95,0.8085)(1.00,0.8083)(1.05,0.8081)(1.10,0.8080)(1.15,0.8079)(1.20,0.8078)(1.25,0.8076)(1.30,0.8074)(1.35,0.8073)(1.40,0.8072)(1.45,0.8072)(1.50,0.8071)(1.55,0.8070)(1.60,0.8070)(1.65,0.8070)(1.70,0.8069)(1.75,0.8068)(1.80,0.8068)(1.85,0.8068)(1.90,0.8068)(1.95,0.8067)(2.00,0.8067)(2.05,0.8067)(2.10,0.8067)(2.15,0.8066)(2.20,0.8066)(2.25,0.8066)(2.30,0.8066)(2.35,0.8066)(2.40,0.8066)(2.45,0.8066)(2.50,0.8066)(2.55,0.8066)(2.60,0.8066)(2.65,0.8066)(2.70,0.8066)(2.75,0.8066)(2.80,0.8066)(2.85,0.8066)(2.90,0.8066)(2.95,0.8066)(3.00,0.8066)
		
	};
\addlegendentry{{\(c_{[l]}=3\)}}
\addplot[blue!60!white,mark=o,mark size=2pt,line width=1pt] plot coordinates{(-0.80,0.8069)};
\addplot[green!80!black!90!,only marks,mark=o,mark size=2pt,line width=1pt] plot coordinates{(-0.20,0.8168)};
\addplot[black,mark=x,mark size=2pt,line width=1pt] plot coordinates{(-0.80,0.8099)};
\addplot[black,only marks,mark=x,mark size=2pt,line width=1pt] plot coordinates{(-0.20,0.8170)};

	\end{axis}
	\end{tikzpicture} &\begin{tikzpicture}[font=\footnotesize]
	\renewcommand{\axisdefaulttryminticks}{4} 
	\tikzstyle{every major grid}+=[style=densely dashed ] \tikzstyle{every axis y label}+=[yshift=-10pt] 
	\tikzstyle{every axis x label}+=[yshift=5pt]
	\tikzstyle{every axis legend}+=[cells={anchor=west},fill=white,
	at={(1,0.2)}, anchor=north east, font=\small]
	\begin{axis}[
	%ybar,
	width=.33\linewidth,
	height=.3\linewidth,
	ymin=0.74,
	ymax=0.82,
	ymajorgrids=false,
	tick pos=left,
	scaled ticks=true,
	yticklabels={},
	ylabel={},
	]
	\addplot[smooth,color=blue!60!white,line width=1pt] plot coordinates{
		
		(1,0.7595)(2,0.7599)(3,0.7604)(4,0.7609)(5,0.7613)(6,0.7618)(7,0.7621)(8,0.7626)(9,0.7630)(10,0.7634)(11,0.7638)(12,0.7643)(13,0.7647)(14,0.7651)(15,0.7655)(16,0.7659)(17,0.7663)(18,0.7666)(19,0.7671)(20,0.7675)(21,0.7679)(22,0.7683)(23,0.7688)(24,0.7692)(25,0.7695)(26,0.7699)(27,0.7703)(28,0.7706)(29,0.7709)(30,0.7713)(31,0.7717)(32,0.7721)(33,0.7724)(34,0.7727)(35,0.7730)(36,0.7734)(37,0.7738)(38,0.7742)(39,0.7745)(40,0.7748)(41,0.7751)(42,0.7754)(43,0.7758)(44,0.7761)(45,0.7764)(46,0.7766)(47,0.7770)(48,0.7773)(49,0.7775)(50,0.7777)(51,0.7780)(52,0.7783)(53,0.7786)(54,0.7789)(55,0.7792)(56,0.7794)(57,0.7797)(58,0.7799)(59,0.7802)(60,0.7804)(61,0.7807)(62,0.7809)(63,0.7811)(64,0.7813)(65,0.7815)(66,0.7817)(67,0.7819)(68,0.7820)(69,0.7822)(70,0.7824)(71,0.7826)(72,0.7827)(73,0.7830)(74,0.7832)(75,0.7834)(76,0.7835)(77,0.7837)(78,0.7838)(79,0.7839)(80,0.7840)(81,0.7841)(82,0.7843)(83,0.7844)(84,0.7845)(85,0.7846)(86,0.7847)(87,0.7848)(88,0.7849)(89,0.7850)(90,0.7851)(91,0.7853)(92,0.7854)(93,0.7855)(94,0.7855)(95,0.7856)(96,0.7857)(97,0.7858)(98,0.7858)(99,0.7859)(100,0.7859)(101,0.7860)(102,0.7861)(103,0.7861)(104,0.7861)(105,0.7862)(106,0.7862)(107,0.7862)(108,0.7862)(109,0.7862)(110,0.7862)(111,0.7863)(112,0.7863)(113,0.7863)(114,0.7862)(115,0.7862)(116,0.7862)(117,0.7862)(118,0.7862)(119,0.7862)(120,0.7861)(121,0.7861)(122,0.7861)(123,0.7860)(124,0.7860)(125,0.7859)(126,0.7858)(127,0.7858)(128,0.7858)(129,0.7857)(130,0.7856)(131,0.7855)(132,0.7854)(133,0.7853)(134,0.7852)(135,0.7851)(136,0.7850)(137,0.7848)(138,0.7847)(139,0.7846)(140,0.7845)(141,0.7843)(142,0.7842)(143,0.7841)(144,0.7839)(145,0.7837)(146,0.7836)(147,0.7835)(148,0.7834)(149,0.7833)(150,0.7832)(151,0.7831)(152,0.7829)(153,0.7827)(154,0.7826)(155,0.7824)(156,0.7822)(157,0.7821)(158,0.7819)(159,0.7817)(160,0.7815)(161,0.7812)(162,0.7811)(163,0.7809)(164,0.7807)(165,0.7804)(166,0.7803)(167,0.7801)(168,0.7799)(169,0.7797)(170,0.7796)(171,0.7793)(172,0.7792)(173,0.7790)(174,0.7788)(175,0.7785)(176,0.7783)(177,0.7782)(178,0.7779)(179,0.7777)(180,0.7774)(181,0.7772)(182,0.7770)(183,0.7767)(184,0.7765)(185,0.7763)(186,0.7759)(187,0.7757)(188,0.7755)(189,0.7752)(190,0.7749)(191,0.7746)(192,0.7743)(193,0.7740)(194,0.7738)(195,0.7735)(196,0.7732)(197,0.7729)(198,0.7726)(199,0.7724)(200,0.7721)
	};
	\addplot[densely dashed,color=green!80!black!90!,line width=1pt] plot coordinates{
		
		(1,0.8070)(2,0.8071)(3,0.8072)(4,0.8073)(5,0.8075)(6,0.8076)(7,0.8077)(8,0.8079)(9,0.8079)(10,0.8081)(11,0.8082)(12,0.8083)(13,0.8084)(14,0.8086)(15,0.8087)(16,0.8089)(17,0.8090)(18,0.8091)(19,0.8092)(20,0.8093)(21,0.8093)(22,0.8094)(23,0.8095)(24,0.8096)(25,0.8096)(26,0.8097)(27,0.8099)(28,0.8099)(29,0.8101)(30,0.8101)(31,0.8102)(32,0.8102)(33,0.8102)(34,0.8102)(35,0.8102)(36,0.8102)(37,0.8103)(38,0.8104)(39,0.8104)(40,0.8104)(41,0.8104)(42,0.8105)(43,0.8106)(44,0.8105)(45,0.8105)(46,0.8105)(47,0.8106)(48,0.8105)(49,0.8105)(50,0.8106)(51,0.8106)(52,0.8107)(53,0.8107)(54,0.8107)(55,0.8106)(56,0.8106)(57,0.8106)(58,0.8106)(59,0.8106)(60,0.8106)(61,0.8106)(62,0.8105)(63,0.8105)(64,0.8105)(65,0.8104)(66,0.8104)(67,0.8104)(68,0.8104)(69,0.8104)(70,0.8103)(71,0.8103)(72,0.8102)(73,0.8102)(74,0.8101)(75,0.8101)(76,0.8100)(77,0.8100)(78,0.8100)(79,0.8100)(80,0.8099)(81,0.8099)(82,0.8099)(83,0.8098)(84,0.8098)(85,0.8097)(86,0.8097)(87,0.8096)(88,0.8096)(89,0.8095)(90,0.8094)(91,0.8094)(92,0.8093)(93,0.8093)(94,0.8092)(95,0.8091)(96,0.8090)(97,0.8090)(98,0.8089)(99,0.8087)(100,0.8086)(101,0.8086)(102,0.8085)(103,0.8084)(104,0.8083)(105,0.8083)(106,0.8082)(107,0.8080)(108,0.8079)(109,0.8078)(110,0.8076)(111,0.8075)(112,0.8074)(113,0.8073)(114,0.8072)(115,0.8071)(116,0.8070)(117,0.8069)(118,0.8068)(119,0.8066)(120,0.8065)(121,0.8064)(122,0.8062)(123,0.8061)(124,0.8060)(125,0.8059)(126,0.8058)(127,0.8057)(128,0.8057)(129,0.8055)(130,0.8054)(131,0.8053)(132,0.8052)(133,0.8050)(134,0.8049)(135,0.8048)(136,0.8046)(137,0.8045)(138,0.8044)(139,0.8043)(140,0.8042)(141,0.8040)(142,0.8039)(143,0.8038)(144,0.8036)(145,0.8035)(146,0.8034)(147,0.8033)(148,0.8031)(149,0.8030)(150,0.8029)(151,0.8028)(152,0.8027)(153,0.8025)(154,0.8024)(155,0.8023)(156,0.8022)(157,0.8021)(158,0.8019)(159,0.8018)(160,0.8016)(161,0.8015)(162,0.8014)(163,0.8012)(164,0.8011)(165,0.8009)(166,0.8007)(167,0.8006)(168,0.8004)(169,0.8002)(170,0.8001)(171,0.7999)(172,0.7997)(173,0.7996)(174,0.7995)(175,0.7993)(176,0.7992)(177,0.7990)(178,0.7989)(179,0.7987)(180,0.7985)(181,0.7984)(182,0.7982)(183,0.7981)(184,0.7979)(185,0.7977)(186,0.7976)(187,0.7974)(188,0.7972)(189,0.7971)(190,0.7968)(191,0.7967)(192,0.7966)(193,0.7964)(194,0.7962)(195,0.7961)(196,0.7959)(197,0.7958)(198,0.7957)(199,0.7955)(200,0.7954)
		
	};

\addplot[blue!60!white,mark=o,mark size=2pt,line width=1pt] plot coordinates{(111,0.7863)};
\addplot[green!80!black!90!,only marks,mark=o,mark size=2pt,line width=1pt] plot coordinates{(52,0.8107)};
\addplot[black,mark=x,mark size=2pt,line width=1pt] plot coordinates{(111,0.8099)};
\addplot[black,only marks,mark=x,mark size=2pt,line width=1pt] plot coordinates{(52,0.8170)};

	\end{axis}
	\end{tikzpicture} &\begin{tikzpicture}[font=\footnotesize]
	\renewcommand{\axisdefaulttryminticks}{4} 
	\tikzstyle{every major grid}+=[style=densely dashed ] \tikzstyle{every axis y label}+=[yshift=-10pt] 
	\tikzstyle{every axis x label}+=[yshift=5pt]
	\tikzstyle{every axis legend}+=[cells={anchor=west},fill=white,
	at={(1,0.2)}, anchor=north east, font=\small]
	\begin{axis}[
	%ybar,
	width=.33\linewidth,
	height=.3\linewidth,
	ymin=0.74,
	ymax=0.82,
	tick pos=left,
	ymajorgrids=false,
	scaled ticks=true,
	ylabel={},
	yticklabels={},
	]
	\addplot[smooth,color=blue!60!white,line width=1pt] plot coordinates{
	(1,0.7996)(2,0.7995)(3,0.7990)(4,0.7985)(5,0.7979)(6,0.7972)(7,0.7967)(8,0.7962)(9,0.7956)(10,0.7946)(11,0.7938)(12,0.7932)(13,0.7924)(14,0.7919)(15,0.7910)(16,0.7902)(17,0.7896)(18,0.7887)(19,0.7880)(20,0.7873)(21,0.7862)(22,0.7857)(23,0.7849)(24,0.7838)(25,0.7831)(26,0.7821)(27,0.7813)(28,0.7805)(29,0.7798)(30,0.7784)(31,0.7777)(32,0.7770)(33,0.7762)(34,0.7756)(35,0.7747)(36,0.7739)(37,0.7732)(38,0.7723)(39,0.7715)(40,0.7705)(41,0.7692)(42,0.7681)(43,0.7675)(44,0.7666)(45,0.7653)(46,0.7645)(47,0.7635)(48,0.7625)(49,0.7615)(50,0.7609)(51,0.7596)(52,0.7589)(53,0.7578)(54,0.7569)(55,0.7564)(56,0.7557)(57,0.7546)(58,0.7536)(59,0.7527)(60,0.7515)(61,0.7508)(62,0.7496)(63,0.7487)(64,0.7476)(65,0.7465)(66,0.7457)(67,0.7450)(68,0.7437)(69,0.7427)(70,0.7415)(71,0.7408)(72,0.7394)(73,0.7387)(74,0.7376)(75,0.7367)(76,0.7359)(77,0.7350)(78,0.7340)(79,0.7330)(80,0.7321)(81,0.7312)(82,0.7299)(83,0.7289)(84,0.7281)(85,0.7272)(86,0.7260)(87,0.7256)(88,0.7245)(89,0.7231)(90,0.7220)(91,0.7210)(92,0.7200)(93,0.7185)(94,0.7178)(95,0.7162)(96,0.7150)(97,0.7141)(98,0.7129)(99,0.7123)(100,0.7108)(101,0.7095)(102,0.7082)(103,0.7076)(104,0.7059)(105,0.7050)(106,0.7040)(107,0.7029)(108,0.7012)(109,0.6999)(110,0.6984)(111,0.6974)(112,0.6960)(113,0.6947)(114,0.6938)(115,0.6928)(116,0.6919)(117,0.6910)(118,0.6900)(119,0.6887)(120,0.6875)(121,0.6864)(122,0.6851)(123,0.6838)(124,0.6829)(125,0.6816)(126,0.6800)(127,0.6784)(128,0.6774)(129,0.6769)(130,0.6755)(131,0.6742)(132,0.6733)(133,0.6723)(134,0.6709)(135,0.6693)(136,0.6675)(137,0.6660)(138,0.6644)(139,0.6631)(140,0.6618)(141,0.6599)(142,0.6589)(143,0.6578)(144,0.6559)(145,0.6544)(146,0.6532)(147,0.6516)(148,0.6502)(149,0.6485)(150,0.6475)(151,0.6458)(152,0.6438)(153,0.6423)(154,0.6407)(155,0.6393)(156,0.6379)(157,0.6357)(158,0.6342)(159,0.6332)(160,0.6315)(161,0.6299)(162,0.6285)(163,0.6268)(164,0.6251)(165,0.6236)(166,0.6217)(167,0.6199)(168,0.6174)(169,0.6157)(170,0.6136)(171,0.6113)(172,0.6093)(173,0.6076)(174,0.6067)(175,0.6048)(176,0.6024)(177,0.5999)(178,0.5976)(179,0.5955)(180,0.5929)(181,0.5901)(182,0.5886)(183,0.5862)(184,0.5838)(185,0.5810)(186,0.5767)(187,0.5741)(188,0.5714)(189,0.5680)(190,0.5652)(191,0.5624)(192,0.5591)(193,0.5563)(194,0.5518)(195,0.5476)(196,0.5420)(197,0.5385)(198,0.5327)(199,0.5253)(200,0.5175)
		
	};
	\addplot[densely dashed,color=green!80!black!90!,line width=1pt] plot coordinates{
	(1,0.8026)(2,0.8032)(3,0.8036)(4,0.8039)(5,0.8043)(6,0.8044)(7,0.8044)(8,0.8043)(9,0.8040)(10,0.8045)(11,0.8045)(12,0.8045)(13,0.8048)(14,0.8047)(15,0.8047)(16,0.8046)(17,0.8044)(18,0.8047)(19,0.8047)(20,0.8047)(21,0.8043)(22,0.8042)(23,0.8037)(24,0.8041)(25,0.8040)(26,0.8041)(27,0.8039)(28,0.8041)(29,0.8042)(30,0.8042)(31,0.8040)(32,0.8039)(33,0.8038)(34,0.8039)(35,0.8038)(36,0.8038)(37,0.8033)(38,0.8031)(39,0.8029)(40,0.8030)(41,0.8030)(42,0.8026)(43,0.8027)(44,0.8022)(45,0.8023)(46,0.8020)(47,0.8023)(48,0.8020)(49,0.8017)(50,0.8017)(51,0.8014)(52,0.8014)(53,0.8012)(54,0.8011)(55,0.8009)(56,0.8011)(57,0.8010)(58,0.8005)(59,0.8004)(60,0.7999)(61,0.8000)(62,0.7999)(63,0.7998)(64,0.7998)(65,0.7999)(66,0.7995)(67,0.7993)(68,0.7987)(69,0.7989)(70,0.7987)(71,0.7985)(72,0.7980)(73,0.7978)(74,0.7975)(75,0.7975)(76,0.7974)(77,0.7972)(78,0.7971)(79,0.7968)(80,0.7971)(81,0.7969)(82,0.7963)(83,0.7963)(84,0.7959)(85,0.7957)(86,0.7955)(87,0.7953)(88,0.7950)(89,0.7950)(90,0.7948)(91,0.7944)(92,0.7940)(93,0.7939)(94,0.7936)(95,0.7932)(96,0.7931)(97,0.7928)(98,0.7926)(99,0.7920)(100,0.7920)(101,0.7917)(102,0.7913)(103,0.7911)(104,0.7909)(105,0.7908)(106,0.7904)(107,0.7901)(108,0.7898)(109,0.7896)(110,0.7897)(111,0.7894)(112,0.7889)(113,0.7885)(114,0.7882)(115,0.7876)(116,0.7874)(117,0.7872)(118,0.7872)(119,0.7867)(120,0.7864)(121,0.7862)(122,0.7859)(123,0.7857)(124,0.7856)(125,0.7853)(126,0.7853)(127,0.7850)(128,0.7845)(129,0.7838)(130,0.7837)(131,0.7833)(132,0.7831)(133,0.7830)(134,0.7826)(135,0.7822)(136,0.7818)(137,0.7815)(138,0.7809)(139,0.7810)(140,0.7806)(141,0.7803)(142,0.7800)(143,0.7797)(144,0.7798)(145,0.7794)(146,0.7791)(147,0.7785)(148,0.7783)(149,0.7783)(150,0.7784)(151,0.7777)(152,0.7776)(153,0.7769)(154,0.7765)(155,0.7766)(156,0.7762)(157,0.7757)(158,0.7757)(159,0.7756)(160,0.7751)(161,0.7751)(162,0.7747)(163,0.7745)(164,0.7745)(165,0.7740)(166,0.7738)(167,0.7730)(168,0.7728)(169,0.7723)(170,0.7719)(171,0.7714)(172,0.7710)(173,0.7704)(174,0.7698)(175,0.7696)(176,0.7694)(177,0.7694)(178,0.7692)(179,0.7689)(180,0.7687)(181,0.7683)(182,0.7675)(183,0.7674)(184,0.7671)(185,0.7668)(186,0.7665)(187,0.7664)(188,0.7658)(189,0.7656)(190,0.7651)(191,0.7646)(192,0.7645)(193,0.7640)(194,0.7638)(195,0.7634)(196,0.7632)(197,0.7630)(198,0.7626)(199,0.7621)(200,0.7614)(201,0.7610)(202,0.7607)(203,0.7603)(204,0.7598)(205,0.7593)(206,0.7585)(207,0.7579)(208,0.7575)(209,0.7571)(210,0.7566)(211,0.7562)(212,0.7556)(213,0.7554)(214,0.7551)(215,0.7545)(216,0.7543)(217,0.7539)(218,0.7534)(219,0.7529)(220,0.7527)(221,0.7523)(222,0.7519)(223,0.7512)(224,0.7507)(225,0.7503)(226,0.7500)(227,0.7495)(228,0.7489)(229,0.7484)(230,0.7481)(231,0.7478)(232,0.7473)(233,0.7469)(234,0.7460)(235,0.7457)(236,0.7454)(237,0.7449)(238,0.7447)(239,0.7442)(240,0.7439)(241,0.7434)(242,0.7430)(243,0.7424)(244,0.7418)(245,0.7414)(246,0.7409)(247,0.7405)(248,0.7402)(249,0.7397)(250,0.7394)(251,0.7389)(252,0.7384)(253,0.7378)(254,0.7376)(255,0.7370)(256,0.7367)(257,0.7364)(258,0.7359)(259,0.7355)(260,0.7351)(261,0.7346)(262,0.7341)(263,0.7336)(264,0.7330)(265,0.7325)(266,0.7321)(267,0.7316)(268,0.7313)(269,0.7307)(270,0.7304)(271,0.7300)(272,0.7299)(273,0.7296)(274,0.7289)(275,0.7284)(276,0.7279)(277,0.7275)(278,0.7265)(279,0.7264)(280,0.7256)(281,0.7249)(282,0.7242)(283,0.7238)(284,0.7234)(285,0.7230)(286,0.7226)(287,0.7221)(288,0.7217)(289,0.7212)(290,0.7209)(291,0.7205)(292,0.7196)(293,0.7193)(294,0.7191)(295,0.7185)(296,0.7180)(297,0.7175)(298,0.7168)(299,0.7161)(300,0.7157)(301,0.7150)(302,0.7143)(303,0.7138)(304,0.7134)(305,0.7131)(306,0.7125)(307,0.7121)(308,0.7117)(309,0.7115)(310,0.7107)(311,0.7102)(312,0.7100)(313,0.7093)(314,0.7086)(315,0.7086)(316,0.7080)(317,0.7076)(318,0.7072)(319,0.7066)(320,0.7062)(321,0.7058)(322,0.7053)(323,0.7049)(324,0.7044)(325,0.7040)(326,0.7036)(327,0.7033)(328,0.7028)(329,0.7027)(330,0.7020)(331,0.7009)(332,0.7004)(333,0.6999)(334,0.6999)(335,0.6995)(336,0.6992)(337,0.6988)(338,0.6978)(339,0.6970)(340,0.6966)(341,0.6962)(342,0.6958)(343,0.6955)(344,0.6947)(345,0.6944)(346,0.6942)(347,0.6937)(348,0.6929)(349,0.6926)(350,0.6920)(351,0.6920)(352,0.6914)(353,0.6907)(354,0.6902)(355,0.6897)(356,0.6892)(357,0.6886)(358,0.6886)(359,0.6878)(360,0.6871)(361,0.6871)(362,0.6862)(363,0.6860)(364,0.6861)(365,0.6853)(366,0.6847)(367,0.6844)(368,0.6835)(369,0.6834)(370,0.6829)(371,0.6824)(372,0.6819)(373,0.6815)(374,0.6808)(375,0.6800)(376,0.6798)(377,0.6796)(378,0.6788)(379,0.6780)(380,0.6777)(381,0.6772)(382,0.6771)(383,0.6765)(384,0.6764)(385,0.6759)(386,0.6749)(387,0.6746)(388,0.6740)(389,0.6736)(390,0.6730)(391,0.6722)(392,0.6719)(393,0.6714)(394,0.6708)(395,0.6703)(396,0.6696)(397,0.6693)(398,0.6685)(399,0.6675)(400,0.6671)(401,0.6665)(402,0.6661)(403,0.6658)(404,0.6648)(405,0.6646)(406,0.6640)(407,0.6635)(408,0.6630)(409,0.6627)(410,0.6625)(411,0.6617)(412,0.6616)(413,0.6613)(414,0.6610)(415,0.6603)(416,0.6600)(417,0.6595)(418,0.6588)(419,0.6583)(420,0.6576)(421,0.6577)(422,0.6567)(423,0.6562)(424,0.6558)(425,0.6550)(426,0.6543)(427,0.6542)(428,0.6531)(429,0.6526)(430,0.6522)(431,0.6520)(432,0.6516)(433,0.6509)(434,0.6501)(435,0.6493)(436,0.6490)(437,0.6488)(438,0.6478)(439,0.6474)(440,0.6468)(441,0.6464)(442,0.6456)(443,0.6452)(444,0.6447)(445,0.6442)(446,0.6434)(447,0.6429)(448,0.6426)(449,0.6420)(450,0.6416)(451,0.6413)(452,0.6408)(453,0.6401)(454,0.6389)(455,0.6383)(456,0.6378)(457,0.6373)(458,0.6368)(459,0.6359)(460,0.6359)(461,0.6351)(462,0.6352)(463,0.6342)(464,0.6338)(465,0.6330)(466,0.6326)(467,0.6321)(468,0.6317)(469,0.6313)(470,0.6307)(471,0.6298)(472,0.6290)(473,0.6284)(474,0.6283)(475,0.6278)(476,0.6267)(477,0.6266)(478,0.6258)(479,0.6251)(480,0.6244)(481,0.6238)(482,0.6238)(483,0.6232)(484,0.6222)(485,0.6217)(486,0.6210)(487,0.6200)(488,0.6196)(489,0.6192)(490,0.6187)(491,0.6179)(492,0.6177)(493,0.6165)(494,0.6155)(495,0.6155)(496,0.6152)(497,0.6147)(498,0.6144)(499,0.6135)(500,0.6138)(501,0.6127)(502,0.6124)(503,0.6120)(504,0.6110)(505,0.6102)(506,0.6091)(507,0.6085)(508,0.6078)(509,0.6075)(510,0.6069)(511,0.6067)(512,0.6056)(513,0.6048)(514,0.6040)(515,0.6032)(516,0.6030)(517,0.6025)(518,0.6018)(519,0.6013)(520,0.6013)(521,0.6004)(522,0.5995)(523,0.5986)(524,0.5984)(525,0.5978)(526,0.5973)(527,0.5956)(528,0.5956)(529,0.5948)(530,0.5936)(531,0.5929)(532,0.5917)(533,0.5912)(534,0.5911)(535,0.5904)(536,0.5895)(537,0.5890)(538,0.5882)(539,0.5873)(540,0.5867)(541,0.5860)(542,0.5850)(543,0.5841)(544,0.5835)(545,0.5823)(546,0.5813)(547,0.5811)(548,0.5801)(549,0.5795)(550,0.5786)(551,0.5777)(552,0.5772)(553,0.5755)(554,0.5753)(555,0.5743)(556,0.5744)(557,0.5731)(558,0.5715)(559,0.5704)(560,0.5698)(561,0.5687)(562,0.5683)(563,0.5672)(564,0.5657)(565,0.5643)(566,0.5638)(567,0.5624)(568,0.5616)(569,0.5608)(570,0.5597)(571,0.5581)(572,0.5571)(573,0.5559)(574,0.5552)(575,0.5546)(576,0.5533)(577,0.5520)(578,0.5500)(579,0.5497)(580,0.5489)(581,0.5479)(582,0.5468)(583,0.5457)(584,0.5444)(585,0.5416)(586,0.5406)(587,0.5399)(588,0.5385)(589,0.5367)(590,0.5350)(591,0.5336)(592,0.5323)(593,0.5304)(594,0.5289)(595,0.5261)(596,0.5248)(597,0.5215)(598,0.5188)(599,0.5131)(600,0.5096)
		
	};
	
	\addplot[blue!60!white,mark=o,mark size=2pt,line width=1pt] plot coordinates{(1,0.7996)};
	\addplot[green!80!black!90!,only marks,mark=o,mark size=2pt,line width=1pt] plot coordinates{(13,0.8048)};
	\addplot[black,mark=x,mark size=2pt,line width=1pt] plot coordinates{(1,0.8099)};
	\addplot[black,only marks,mark=x,mark size=2pt,line width=1pt] plot coordinates{(13,0.8170)};
	
	\end{axis}
	\end{tikzpicture}\\
	\begin{tikzpicture}[font=\footnotesize]
	\renewcommand{\axisdefaulttryminticks}{4} 
	\tikzstyle{every major grid}+=[style=densely dashed ] \tikzstyle{every axis y label}+=[yshift=-10pt] 
	\tikzstyle{every axis x label}+=[yshift=5pt]
	\tikzstyle{every axis legend}+=[cells={anchor=west},fill=white,
	at={(1,0.2)}, anchor=north east, font=\scriptsize]
	\begin{axis}[
	%ybar,
	width=.33\linewidth,
	height=.3\linewidth,
	ymin=0.65,
	ymax=0.72,
	ymajorgrids=false,
	scaled ticks=true,
	tick pos=left,
	ylabel={Accuracy},
	]
	\addplot[smooth,color=blue!60!white,line width=1pt] plot coordinates{
	(-3.00,0.6197)(-2.95,0.6207)(-2.90,0.6218)(-2.85,0.6229)(-2.80,0.6241)(-2.75,0.6253)(-2.70,0.6266)(-2.65,0.6281)(-2.60,0.6297)(-2.55,0.6312)(-2.50,0.6330)(-2.45,0.6349)(-2.40,0.6367)(-2.35,0.6388)(-2.30,0.6408)(-2.25,0.6428)(-2.20,0.6451)(-2.15,0.6473)(-2.10,0.6495)(-2.05,0.6519)(-2.00,0.6543)(-1.95,0.6566)(-1.90,0.6589)(-1.85,0.6613)(-1.80,0.6636)(-1.75,0.6661)(-1.70,0.6684)(-1.65,0.6707)(-1.60,0.6729)(-1.55,0.6749)(-1.50,0.6768)(-1.45,0.6787)(-1.40,0.6805)(-1.35,0.6819)(-1.30,0.6833)(-1.25,0.6845)(-1.20,0.6857)(-1.15,0.6869)(-1.10,0.6876)(-1.05,0.6882)(-1.00,0.6884)(-0.95,0.6882)(-0.90,0.6884)(-0.85,0.6884)(-0.80,0.6881)(-0.75,0.6871)(-0.70,0.6866)(-0.65,0.6857)(-0.60,0.6848)(-0.55,0.6840)(-0.50,0.6826)(-0.45,0.6815)(-0.40,0.6804)(-0.35,0.6792)(-0.30,0.6780)(-0.25,0.6768)(-0.20,0.6754)(-0.15,0.6743)(-0.10,0.6731)(-0.05,0.6718)(0.00,0.6707)(0.05,0.6698)(0.10,0.6688)(0.15,0.6679)(0.20,0.6668)(0.25,0.6660)(0.30,0.6653)(0.35,0.6646)(0.40,0.6638)(0.45,0.6632)(0.50,0.6626)(0.55,0.6620)(0.60,0.6615)(0.65,0.6611)(0.70,0.6607)(0.75,0.6604)(0.80,0.6601)(0.85,0.6598)(0.90,0.6595)(0.95,0.6594)(1.00,0.6591)(1.05,0.6590)(1.10,0.6588)(1.15,0.6586)(1.20,0.6585)(1.25,0.6584)(1.30,0.6582)(1.35,0.6581)(1.40,0.6580)(1.45,0.6578)(1.50,0.6578)(1.55,0.6577)(1.60,0.6576)(1.65,0.6576)(1.70,0.6575)(1.75,0.6575)(1.80,0.6575)(1.85,0.6574)(1.90,0.6574)(1.95,0.6573)(2.00,0.6573)(2.05,0.6573)(2.10,0.6572)(2.15,0.6572)(2.20,0.6572)(2.25,0.6572)(2.30,0.6572)(2.35,0.6572)(2.40,0.6572)(2.45,0.6572)(2.50,0.6572)(2.55,0.6571)(2.60,0.6571)(2.65,0.6571)(2.70,0.6571)(2.75,0.6571)(2.80,0.6571)(2.85,0.6571)(2.90,0.6571)(2.95,0.6571)(3.00,0.6571)

	};
	\addplot[densely dashed,color=green!80!black!90!,line width=1pt] plot coordinates{
      (-3.00,0.5875)(-2.95,0.5893)(-2.90,0.5910)(-2.85,0.5929)(-2.80,0.5948)(-2.75,0.5966)(-2.70,0.5986)(-2.65,0.6010)(-2.60,0.6034)(-2.55,0.6059)(-2.50,0.6084)(-2.45,0.6112)(-2.40,0.6144)(-2.35,0.6172)(-2.30,0.6202)(-2.25,0.6232)(-2.20,0.6263)(-2.15,0.6295)(-2.10,0.6331)(-2.05,0.6364)(-2.00,0.6399)(-1.95,0.6434)(-1.90,0.6471)(-1.85,0.6503)(-1.80,0.6540)(-1.75,0.6578)(-1.70,0.6611)(-1.65,0.6647)(-1.60,0.6683)(-1.55,0.6716)(-1.50,0.6746)(-1.45,0.6781)(-1.40,0.6811)(-1.35,0.6841)(-1.30,0.6870)(-1.25,0.6899)(-1.20,0.6923)(-1.15,0.6950)(-1.10,0.6976)(-1.05,0.6997)(-1.00,0.7019)(-0.95,0.7037)(-0.90,0.7056)(-0.85,0.7072)(-0.80,0.7089)(-0.75,0.7106)(-0.70,0.7119)(-0.65,0.7128)(-0.60,0.7137)(-0.55,0.7144)(-0.50,0.7147)(-0.45,0.7150)(-0.40,0.7150)(-0.35,0.7153)(-0.30,0.7155)(-0.25,0.7152)(-0.20,0.7151)(-0.15,0.7149)(-0.10,0.7146)(-0.05,0.7142)(0.00,0.7141)(0.05,0.7137)(0.10,0.7133)(0.15,0.7130)(0.20,0.7127)(0.25,0.7122)(0.30,0.7119)(0.35,0.7115)(0.40,0.7112)(0.45,0.7110)(0.50,0.7108)(0.55,0.7103)(0.60,0.7101)(0.65,0.7099)(0.70,0.7097)(0.75,0.7095)(0.80,0.7092)(0.85,0.7091)(0.90,0.7089)(0.95,0.7088)(1.00,0.7087)(1.05,0.7086)(1.10,0.7085)(1.15,0.7083)(1.20,0.7082)(1.25,0.7082)(1.30,0.7081)(1.35,0.7080)(1.40,0.7080)(1.45,0.7080)(1.50,0.7079)(1.55,0.7079)(1.60,0.7079)(1.65,0.7079)(1.70,0.7079)(1.75,0.7079)(1.80,0.7079)(1.85,0.7078)(1.90,0.7078)(1.95,0.7078)(2.00,0.7078)(2.05,0.7078)(2.10,0.7077)(2.15,0.7077)(2.20,0.7077)(2.25,0.7077)(2.30,0.7077)(2.35,0.7077)(2.40,0.7077)(2.45,0.7077)(2.50,0.7077)(2.55,0.7077)(2.60,0.7077)(2.65,0.7077)(2.70,0.7077)(2.75,0.7077)(2.80,0.7077)(2.85,0.7077)(2.90,0.7077)(2.95,0.7077)(3.00,0.7077)
	
};
\addplot[blue!60!white,mark=o,mark size=2pt,line width=1pt] plot coordinates{(-0.85,0.6884)};
\addplot[green!80!black!90!,only marks,mark=o,mark size=2pt,line width=1pt] plot coordinates{(-0.30,0.7155)};
\addplot[black,mark=x,mark size=2pt,line width=1pt] plot coordinates{(-0.85,0.6920)};
\addplot[black,only marks,mark=x,mark size=2pt,line width=1pt] plot coordinates{(-0.30,0.7161)};

    %\legend{ {}, {Empirical}}
	\end{axis}
	\end{tikzpicture} &\begin{tikzpicture}[font=\footnotesize]
	\renewcommand{\axisdefaulttryminticks}{4} 
	\tikzstyle{every major grid}+=[style=densely dashed ] \tikzstyle{every axis y label}+=[yshift=-10pt] 
	\tikzstyle{every axis x label}+=[yshift=5pt]
	\tikzstyle{every axis legend}+=[cells={anchor=west},fill=white,
	at={(1,0.2)}, anchor=north east, font=\small]
	\begin{axis}[
	%ybar,
	width=.33\linewidth,
	height=.3\linewidth,
	ymin=0.65,
	ymax=0.72,
	%xmode=log,
	%log basis x={10},
	%grid=major,
	tick pos=left,
	ymajorgrids=false,
	scaled ticks=true,
	yticklabels={},
	ylabel={},
	]
	\addplot[smooth,color=blue!60!white,line width=1pt] plot coordinates{
		(1,0.6581)(2,0.6582)(3,0.6585)(4,0.6587)(5,0.6589)(6,0.6591)(7,0.6593)(8,0.6594)(9,0.6596)(10,0.6598)(11,0.6599)(12,0.6601)(13,0.6603)(14,0.6605)(15,0.6607)(16,0.6610)(17,0.6612)(18,0.6614)(19,0.6614)(20,0.6616)(21,0.6618)(22,0.6619)(23,0.6621)(24,0.6623)(25,0.6626)(26,0.6628)(27,0.6630)(28,0.6632)(29,0.6634)(30,0.6635)(31,0.6638)(32,0.6639)(33,0.6641)(34,0.6642)(35,0.6644)(36,0.6646)(37,0.6646)(38,0.6649)(39,0.6650)(40,0.6651)(41,0.6653)(42,0.6654)(43,0.6656)(44,0.6657)(45,0.6659)(46,0.6661)(47,0.6663)(48,0.6664)(49,0.6665)(50,0.6667)(51,0.6668)(52,0.6669)(53,0.6670)(54,0.6672)(55,0.6673)(56,0.6675)(57,0.6675)(58,0.6677)(59,0.6678)(60,0.6679)(61,0.6681)(62,0.6682)(63,0.6683)(64,0.6684)(65,0.6685)(66,0.6685)(67,0.6686)(68,0.6687)(69,0.6688)(70,0.6689)(71,0.6690)(72,0.6691)(73,0.6692)(74,0.6692)(75,0.6693)(76,0.6694)(77,0.6694)(78,0.6695)(79,0.6697)(80,0.6698)(81,0.6699)(82,0.6699)(83,0.6701)(84,0.6702)(85,0.6702)(86,0.6703)(87,0.6703)(88,0.6705)(89,0.6706)(90,0.6706)(91,0.6707)(92,0.6707)(93,0.6707)(94,0.6707)(95,0.6707)(96,0.6707)(97,0.6707)(98,0.6707)(99,0.6707)(100,0.6708)(101,0.6708)(102,0.6709)(103,0.6709)(104,0.6709)(105,0.6709)(106,0.6709)(107,0.6709)(108,0.6709)(109,0.6709)(110,0.6708)(111,0.6709)(112,0.6709)(113,0.6709)(114,0.6709)(115,0.6709)(116,0.6709)(117,0.6709)(118,0.6709)(119,0.6709)(120,0.6709)(121,0.6709)(122,0.6708)(123,0.6707)(124,0.6708)(125,0.6708)(126,0.6707)(127,0.6707)(128,0.6707)(129,0.6707)(130,0.6706)(131,0.6706)(132,0.6705)(133,0.6705)(134,0.6705)(135,0.6705)(136,0.6705)(137,0.6704)(138,0.6703)(139,0.6702)(140,0.6701)(141,0.6700)(142,0.6698)(143,0.6698)(144,0.6697)(145,0.6697)(146,0.6697)(147,0.6696)(148,0.6696)(149,0.6696)(150,0.6695)(151,0.6694)(152,0.6693)(153,0.6693)(154,0.6693)(155,0.6693)(156,0.6692)(157,0.6692)(158,0.6691)(159,0.6690)(160,0.6689)(161,0.6687)(162,0.6685)(163,0.6684)(164,0.6683)(165,0.6682)(166,0.6681)(167,0.6680)(168,0.6679)(169,0.6678)(170,0.6676)(171,0.6675)(172,0.6673)(173,0.6672)(174,0.6671)(175,0.6671)(176,0.6669)(177,0.6668)(178,0.6667)(179,0.6665)(180,0.6664)(181,0.6663)(182,0.6661)(183,0.6659)(184,0.6658)(185,0.6657)(186,0.6656)(187,0.6654)(188,0.6653)(189,0.6652)(190,0.6651)(191,0.6649)(192,0.6647)(193,0.6646)(194,0.6644)(195,0.6642)(196,0.6641)(197,0.6639)(198,0.6637)(199,0.6635)(200,0.6634)
		
	};
	\addplot[densely dashed,color=green!80!black!90!,line width=1pt] plot coordinates{
	  (1,0.7083)(2,0.7084)(3,0.7085)(4,0.7086)(5,0.7087)(6,0.7087)(7,0.7089)(8,0.7089)(9,0.7090)(10,0.7090)(11,0.7090)(12,0.7091)(13,0.7092)(14,0.7093)(15,0.7093)(16,0.7094)(17,0.7094)(18,0.7095)(19,0.7096)(20,0.7096)(21,0.7097)(22,0.7098)(23,0.7098)(24,0.7099)(25,0.7100)(26,0.7100)(27,0.7099)(28,0.7100)(29,0.7100)(30,0.7100)(31,0.7101)(32,0.7102)(33,0.7102)(34,0.7102)(35,0.7102)(36,0.7103)(37,0.7104)(38,0.7104)(39,0.7104)(40,0.7105)(41,0.7105)(42,0.7105)(43,0.7105)(44,0.7105)(45,0.7105)(46,0.7105)(47,0.7105)(48,0.7105)(49,0.7105)(50,0.7105)(51,0.7105)(52,0.7105)(53,0.7105)(54,0.7105)(55,0.7105)(56,0.7105)(57,0.7104)(58,0.7104)(59,0.7104)(60,0.7103)(61,0.7103)(62,0.7102)(63,0.7102)(64,0.7102)(65,0.7102)(66,0.7102)(67,0.7102)(68,0.7102)(69,0.7102)(70,0.7102)(71,0.7101)(72,0.7101)(73,0.7100)(74,0.7100)(75,0.7100)(76,0.7100)(77,0.7099)(78,0.7098)(79,0.7097)(80,0.7097)(81,0.7096)(82,0.7095)(83,0.7094)(84,0.7093)(85,0.7093)(86,0.7092)(87,0.7092)(88,0.7091)(89,0.7090)(90,0.7089)(91,0.7089)(92,0.7090)(93,0.7089)(94,0.7088)(95,0.7087)(96,0.7087)(97,0.7087)(98,0.7086)(99,0.7085)(100,0.7084)(101,0.7083)(102,0.7082)(103,0.7081)(104,0.7081)(105,0.7080)(106,0.7080)(107,0.7080)(108,0.7079)(109,0.7078)(110,0.7077)(111,0.7076)(112,0.7076)(113,0.7075)(114,0.7075)(115,0.7074)(116,0.7073)(117,0.7073)(118,0.7071)(119,0.7071)(120,0.7070)(121,0.7069)(122,0.7068)(123,0.7068)(124,0.7067)(125,0.7067)(126,0.7067)(127,0.7066)(128,0.7064)(129,0.7062)(130,0.7062)(131,0.7061)(132,0.7060)(133,0.7059)(134,0.7058)(135,0.7058)(136,0.7057)(137,0.7056)(138,0.7054)(139,0.7054)(140,0.7053)(141,0.7052)(142,0.7051)(143,0.7050)(144,0.7050)(145,0.7049)(146,0.7048)(147,0.7047)(148,0.7046)(149,0.7045)(150,0.7044)(151,0.7042)(152,0.7041)(153,0.7040)(154,0.7038)(155,0.7037)(156,0.7036)(157,0.7035)(158,0.7035)(159,0.7033)(160,0.7032)(161,0.7030)(162,0.7029)(163,0.7028)(164,0.7027)(165,0.7025)(166,0.7024)(167,0.7022)(168,0.7021)(169,0.7019)(170,0.7019)(171,0.7018)(172,0.7017)(173,0.7016)(174,0.7014)(175,0.7013)(176,0.7012)(177,0.7011)(178,0.7010)(179,0.7009)(180,0.7008)(181,0.7007)(182,0.7006)(183,0.7004)(184,0.7003)(185,0.7002)(186,0.7001)(187,0.7001)(188,0.6999)(189,0.6999)(190,0.6997)(191,0.6996)(192,0.6994)(193,0.6993)(194,0.6992)(195,0.6992)(196,0.6990)(197,0.6988)(198,0.6987)(199,0.6986)(200,0.6985)
	
};

\addplot[blue!60!white,mark=o,mark size=2pt,line width=1pt] plot coordinates{(102,0.6709)};
\addplot[green!80!black!90!,only marks,mark=o,mark size=2pt,line width=1pt] plot coordinates{(9,0.7090)};
\addplot[black,mark=x,mark size=2pt,line width=1pt] plot coordinates{(102,0.6920)};
\addplot[black,only marks,mark=x,mark size=2pt,line width=1pt] plot coordinates{(9,0.7161)};
	\end{axis}
	\end{tikzpicture} &\begin{tikzpicture}[font=\footnotesize]
	\renewcommand{\axisdefaulttryminticks}{4} 
	\tikzstyle{every major grid}+=[style=densely dashed ] \tikzstyle{every axis y label}+=[yshift=-10pt] 
	\tikzstyle{every axis x label}+=[yshift=5pt]
	\tikzstyle{every axis legend}+=[cells={anchor=west},fill=white,
	at={(1,0.2)}, anchor=north east, font=\small]
	\begin{axis}[
	%ybar,
	width=.33\linewidth,
	height=.3\linewidth,
    ymin=0.65,
	ymax=0.72,
	%xmode=log,
	%log basis x={10},
	%grid=major,
	ymajorgrids=false,
	tick pos=left,
	scaled ticks=true,
	ylabel={},
	yticklabels={},
	]
		\addplot[smooth,color=blue!60!white,line width=1pt] plot coordinates{
	(1,0.6306)(2,0.6470)(3,0.6536)(4,0.6578)(5,0.6607)(6,0.6622)(7,0.6633)(8,0.6641)(9,0.6650)(10,0.6655)(11,0.6658)(12,0.6654)(13,0.6662)(14,0.6662)(15,0.6664)(16,0.6664)(17,0.6667)(18,0.6664)(19,0.6667)(20,0.6667)(21,0.6665)(22,0.6661)(23,0.6663)(24,0.6665)(25,0.6661)(26,0.6654)(27,0.6649)(28,0.6647)(29,0.6642)(30,0.6643)(31,0.6636)(32,0.6631)(33,0.6629)(34,0.6626)(35,0.6620)(36,0.6618)(37,0.6620)(38,0.6617)(39,0.6607)(40,0.6602)(41,0.6597)(42,0.6590)(43,0.6586)(44,0.6583)(45,0.6572)(46,0.6568)(47,0.6560)(48,0.6556)(49,0.6547)(50,0.6547)(51,0.6537)(52,0.6534)(53,0.6530)(54,0.6529)(55,0.6521)(56,0.6516)(57,0.6514)(58,0.6510)(59,0.6502)(60,0.6502)(61,0.6495)(62,0.6489)(63,0.6488)(64,0.6481)(65,0.6475)(66,0.6470)(67,0.6467)(68,0.6467)(69,0.6460)(70,0.6455)(71,0.6444)(72,0.6440)(73,0.6432)(74,0.6425)(75,0.6418)(76,0.6409)(77,0.6405)(78,0.6404)(79,0.6392)(80,0.6388)(81,0.6383)(82,0.6374)(83,0.6369)(84,0.6363)(85,0.6357)(86,0.6345)(87,0.6341)(88,0.6331)(89,0.6326)(90,0.6320)(91,0.6314)(92,0.6304)(93,0.6299)(94,0.6289)(95,0.6285)(96,0.6278)(97,0.6275)(98,0.6263)(99,0.6258)(100,0.6248)(101,0.6240)(102,0.6233)(103,0.6225)(104,0.6222)(105,0.6217)(106,0.6205)(107,0.6201)(108,0.6193)(109,0.6185)(110,0.6177)(111,0.6169)(112,0.6164)(113,0.6157)(114,0.6149)(115,0.6139)(116,0.6131)(117,0.6126)(118,0.6114)(119,0.6105)(120,0.6092)(121,0.6088)(122,0.6077)(123,0.6074)(124,0.6069)(125,0.6062)(126,0.6049)(127,0.6040)(128,0.6028)(129,0.6020)(130,0.6016)(131,0.6004)(132,0.5998)(133,0.5988)(134,0.5976)(135,0.5967)(136,0.5959)(137,0.5949)(138,0.5945)(139,0.5936)(140,0.5928)(141,0.5912)(142,0.5906)(143,0.5892)(144,0.5883)(145,0.5879)(146,0.5869)(147,0.5858)(148,0.5850)(149,0.5843)(150,0.5838)(151,0.5831)(152,0.5822)(153,0.5813)(154,0.5802)(155,0.5792)(156,0.5785)(157,0.5773)(158,0.5764)(159,0.5747)(160,0.5740)(161,0.5733)(162,0.5722)(163,0.5710)(164,0.5696)(165,0.5686)(166,0.5678)(167,0.5669)(168,0.5659)(169,0.5651)(170,0.5640)(171,0.5627)(172,0.5619)(173,0.5608)(174,0.5594)(175,0.5583)(176,0.5570)(177,0.5559)(178,0.5543)(179,0.5528)(180,0.5524)(181,0.5515)(182,0.5500)(183,0.5488)(184,0.5473)(185,0.5454)(186,0.5443)(187,0.5422)(188,0.5407)(189,0.5386)(190,0.5367)(191,0.5353)(192,0.5336)(193,0.5311)(194,0.5290)(195,0.5267)(196,0.5251)(197,0.5223)(198,0.5189)(199,0.5144)(200,0.5095)
		
	};
	\addplot[densely dashed,color=green!80!black!90!,line width=1pt] plot coordinates{
(1,0.6327)(2,0.6508)(3,0.6583)(4,0.6625)(5,0.6665)(6,0.6699)(7,0.6729)(8,0.6756)(9,0.6768)(10,0.6786)(11,0.6803)(12,0.6813)(13,0.6825)(14,0.6843)(15,0.6847)(16,0.6860)(17,0.6871)(18,0.6872)(19,0.6878)(20,0.6889)(21,0.6898)(22,0.6908)(23,0.6914)(24,0.6916)(25,0.6922)(26,0.6929)(27,0.6932)(28,0.6939)(29,0.6941)(30,0.6942)(31,0.6952)(32,0.6949)(33,0.6954)(34,0.6959)(35,0.6964)(36,0.6965)(37,0.6968)(38,0.6974)(39,0.6972)(40,0.6975)(41,0.6973)(42,0.6974)(43,0.6977)(44,0.6979)(45,0.6981)(46,0.6982)(47,0.6980)(48,0.6981)(49,0.6981)(50,0.6979)(51,0.6978)(52,0.6981)(53,0.6983)(54,0.6980)(55,0.6977)(56,0.6981)(57,0.6982)(58,0.6983)(59,0.6984)(60,0.6987)(61,0.6985)(62,0.6989)(63,0.6992)(64,0.6994)(65,0.6990)(66,0.6989)(67,0.6987)(68,0.6987)(69,0.6987)(70,0.6981)(71,0.6979)(72,0.6983)(73,0.6982)(74,0.6976)(75,0.6978)(76,0.6980)(77,0.6979)(78,0.6975)(79,0.6973)(80,0.6973)(81,0.6970)(82,0.6967)(83,0.6964)(84,0.6967)(85,0.6965)(86,0.6963)(87,0.6963)(88,0.6962)(89,0.6964)(90,0.6961)(91,0.6961)(92,0.6958)(93,0.6961)(94,0.6955)(95,0.6955)(96,0.6957)(97,0.6958)(98,0.6957)(99,0.6957)(100,0.6955)(101,0.6953)(102,0.6957)(103,0.6957)(104,0.6954)(105,0.6954)(106,0.6954)(107,0.6955)(108,0.6953)(109,0.6951)(110,0.6952)(111,0.6949)(112,0.6939)(113,0.6937)(114,0.6935)(115,0.6931)(116,0.6931)(117,0.6931)(118,0.6929)(119,0.6929)(120,0.6927)(121,0.6924)(122,0.6921)(123,0.6918)(124,0.6916)(125,0.6917)(126,0.6911)(127,0.6908)(128,0.6910)(129,0.6910)(130,0.6907)(131,0.6907)(132,0.6903)(133,0.6900)(134,0.6900)(135,0.6897)(136,0.6896)(137,0.6896)(138,0.6892)(139,0.6887)(140,0.6885)(141,0.6883)(142,0.6880)(143,0.6877)(144,0.6879)(145,0.6878)(146,0.6875)(147,0.6870)(148,0.6870)(149,0.6864)(150,0.6866)(151,0.6865)(152,0.6861)(153,0.6857)(154,0.6855)(155,0.6851)(156,0.6849)(157,0.6848)(158,0.6845)(159,0.6840)(160,0.6842)(161,0.6840)(162,0.6839)(163,0.6836)(164,0.6833)(165,0.6828)(166,0.6825)(167,0.6822)(168,0.6818)(169,0.6818)(170,0.6810)(171,0.6809)(172,0.6806)(173,0.6809)(174,0.6804)(175,0.6801)(176,0.6797)(177,0.6794)(178,0.6789)(179,0.6787)(180,0.6782)(181,0.6780)(182,0.6774)(183,0.6769)(184,0.6767)(185,0.6767)(186,0.6766)(187,0.6767)(188,0.6763)(189,0.6761)(190,0.6756)(191,0.6756)(192,0.6759)(193,0.6754)(194,0.6751)(195,0.6751)(196,0.6745)(197,0.6743)(198,0.6741)(199,0.6734)(200,0.6732)(201,0.6726)(202,0.6724)(203,0.6719)(204,0.6714)(205,0.6708)(206,0.6705)(207,0.6705)(208,0.6702)(209,0.6699)(210,0.6692)(211,0.6691)(212,0.6691)(213,0.6686)(214,0.6681)(215,0.6676)(216,0.6672)(217,0.6669)(218,0.6664)(219,0.6661)(220,0.6655)(221,0.6649)(222,0.6647)(223,0.6646)(224,0.6641)(225,0.6640)(226,0.6633)(227,0.6630)(228,0.6625)(229,0.6623)(230,0.6617)(231,0.6614)(232,0.6613)(233,0.6610)(234,0.6605)(235,0.6603)(236,0.6600)(237,0.6597)(238,0.6595)(239,0.6590)(240,0.6583)(241,0.6578)(242,0.6572)(243,0.6569)(244,0.6568)(245,0.6563)(246,0.6561)(247,0.6557)(248,0.6555)(249,0.6551)(250,0.6549)(251,0.6548)(252,0.6549)(253,0.6544)(254,0.6539)(255,0.6534)(256,0.6528)(257,0.6527)(258,0.6523)(259,0.6519)(260,0.6517)(261,0.6513)(262,0.6510)(263,0.6507)(264,0.6500)(265,0.6499)(266,0.6496)(267,0.6495)(268,0.6488)(269,0.6487)(270,0.6487)(271,0.6480)(272,0.6472)(273,0.6472)(274,0.6472)(275,0.6470)(276,0.6463)(277,0.6457)(278,0.6453)(279,0.6446)(280,0.6445)(281,0.6443)(282,0.6441)(283,0.6434)(284,0.6430)(285,0.6429)(286,0.6425)(287,0.6425)(288,0.6418)(289,0.6419)(290,0.6414)(291,0.6410)(292,0.6404)(293,0.6406)(294,0.6401)(295,0.6401)(296,0.6398)(297,0.6392)(298,0.6390)(299,0.6381)(300,0.6382)(301,0.6376)(302,0.6372)(303,0.6371)(304,0.6371)(305,0.6365)(306,0.6363)(307,0.6357)(308,0.6355)(309,0.6353)(310,0.6353)(311,0.6351)(312,0.6343)(313,0.6338)(314,0.6338)(315,0.6333)(316,0.6329)(317,0.6328)(318,0.6325)(319,0.6321)(320,0.6321)(321,0.6316)(322,0.6313)(323,0.6309)(324,0.6305)(325,0.6299)(326,0.6297)(327,0.6294)(328,0.6294)(329,0.6293)(330,0.6286)(331,0.6283)(332,0.6282)(333,0.6273)(334,0.6269)(335,0.6271)(336,0.6266)(337,0.6265)(338,0.6261)(339,0.6256)(340,0.6255)(341,0.6248)(342,0.6241)(343,0.6236)(344,0.6235)(345,0.6235)(346,0.6229)(347,0.6226)(348,0.6224)(349,0.6219)(350,0.6217)(351,0.6213)(352,0.6209)(353,0.6208)(354,0.6201)(355,0.6198)(356,0.6196)(357,0.6193)(358,0.6192)(359,0.6188)(360,0.6182)(361,0.6178)(362,0.6173)(363,0.6172)(364,0.6166)(365,0.6164)(366,0.6163)(367,0.6153)(368,0.6151)(369,0.6147)(370,0.6146)(371,0.6147)(372,0.6143)(373,0.6139)(374,0.6136)(375,0.6133)(376,0.6129)(377,0.6127)(378,0.6125)(379,0.6119)(380,0.6119)(381,0.6116)(382,0.6112)(383,0.6110)(384,0.6112)(385,0.6107)(386,0.6102)(387,0.6096)(388,0.6093)(389,0.6097)(390,0.6092)(391,0.6086)(392,0.6083)(393,0.6080)(394,0.6075)(395,0.6078)(396,0.6072)(397,0.6065)(398,0.6061)(399,0.6056)(400,0.6052)(401,0.6049)(402,0.6047)(403,0.6042)(404,0.6037)(405,0.6030)(406,0.6024)(407,0.6027)(408,0.6023)(409,0.6022)(410,0.6018)(411,0.6010)(412,0.6009)(413,0.6010)(414,0.6013)(415,0.6005)(416,0.5999)(417,0.5992)(418,0.5990)(419,0.5987)(420,0.5984)(421,0.5979)(422,0.5979)(423,0.5978)(424,0.5975)(425,0.5968)(426,0.5964)(427,0.5956)(428,0.5946)(429,0.5940)(430,0.5940)(431,0.5928)(432,0.5926)(433,0.5925)(434,0.5913)(435,0.5913)(436,0.5913)(437,0.5909)(438,0.5908)(439,0.5905)(440,0.5902)(441,0.5903)(442,0.5899)(443,0.5893)(444,0.5891)(445,0.5891)(446,0.5885)(447,0.5885)(448,0.5877)(449,0.5878)(450,0.5868)(451,0.5867)(452,0.5858)(453,0.5859)(454,0.5856)(455,0.5853)(456,0.5848)(457,0.5843)(458,0.5844)(459,0.5841)(460,0.5837)(461,0.5833)(462,0.5825)(463,0.5823)(464,0.5826)(465,0.5824)(466,0.5822)(467,0.5817)(468,0.5816)(469,0.5814)(470,0.5802)(471,0.5799)(472,0.5796)(473,0.5796)(474,0.5792)(475,0.5789)(476,0.5783)(477,0.5776)(478,0.5772)(479,0.5764)(480,0.5761)(481,0.5756)(482,0.5750)(483,0.5743)(484,0.5741)(485,0.5734)(486,0.5736)(487,0.5733)(488,0.5732)(489,0.5729)(490,0.5723)(491,0.5716)(492,0.5711)(493,0.5710)(494,0.5708)(495,0.5702)(496,0.5703)(497,0.5697)(498,0.5693)(499,0.5690)(500,0.5688)(501,0.5680)(502,0.5673)(503,0.5671)(504,0.5665)(505,0.5664)(506,0.5661)(507,0.5659)(508,0.5653)(509,0.5654)(510,0.5649)(511,0.5648)(512,0.5644)(513,0.5640)(514,0.5641)(515,0.5636)(516,0.5631)(517,0.5631)(518,0.5625)(519,0.5617)(520,0.5616)(521,0.5610)(522,0.5607)(523,0.5611)(524,0.5603)(525,0.5594)(526,0.5587)(527,0.5584)(528,0.5578)(529,0.5578)(530,0.5572)(531,0.5572)(532,0.5572)(533,0.5569)(534,0.5563)(535,0.5558)(536,0.5551)(537,0.5551)(538,0.5546)(539,0.5536)(540,0.5528)(541,0.5524)(542,0.5521)(543,0.5519)(544,0.5516)(545,0.5509)(546,0.5505)(547,0.5506)(548,0.5498)(549,0.5494)(550,0.5489)(551,0.5479)(552,0.5474)(553,0.5471)(554,0.5464)(555,0.5458)(556,0.5453)(557,0.5442)(558,0.5440)(559,0.5430)(560,0.5429)(561,0.5419)(562,0.5416)(563,0.5410)(564,0.5401)(565,0.5396)(566,0.5391)(567,0.5386)(568,0.5385)(569,0.5381)(570,0.5371)(571,0.5361)(572,0.5351)(573,0.5343)(574,0.5342)(575,0.5336)(576,0.5329)(577,0.5324)(578,0.5321)(579,0.5311)(580,0.5306)(581,0.5305)(582,0.5287)(583,0.5276)(584,0.5264)(585,0.5256)(586,0.5247)(587,0.5236)(588,0.5224)(589,0.5211)(590,0.5203)(591,0.5190)(592,0.5188)(593,0.5178)(594,0.5161)(595,0.5153)(596,0.5132)(597,0.5127)(598,0.5106)(599,0.5072)(600,0.5049)
	
};
\addplot[blue!60!white,mark=o,mark size=2pt,line width=1pt] plot coordinates{(19,0.6667)};
\addplot[green!80!black!90!,only marks,mark=o,mark size=2pt,line width=1pt] plot coordinates{(53,0.6983)};
\addplot[black,mark=x,mark size=2pt,line width=1pt] plot coordinates{(19,0.6920)};
\addplot[black,only marks,mark=x,mark size=2pt,line width=1pt] plot coordinates{(53,0.7161)};
	
	\end{axis}
	\end{tikzpicture}\\
	\(\quad\quad\quad\quad\){\scriptsize \(\log_{10}(\alpha/\Vert\hat W_{[uu]}-1\Vert)\)}&\(\quad\){\scriptsize \(m\)}&\(\quad\){\scriptsize \(s\)}
\end{tabular}
	\caption{Empirical accuracy of graph-based SSL algorithms at different values of hyperparameters for isotropic Gaussian mixture data with \(p=60\),\(n=360\) and \(\Vert\mu\Vert^2=1\) (bottom) or \(\Vert\mu\Vert^2=2\) (top). Averaged over \(1000\) realizations. Best empirical value marked in circle and the asymptotic optimum in cross.}
	\label{fig: graph-based SSL algorithms on isotropic Gaussian mixture data}
\end{figure}
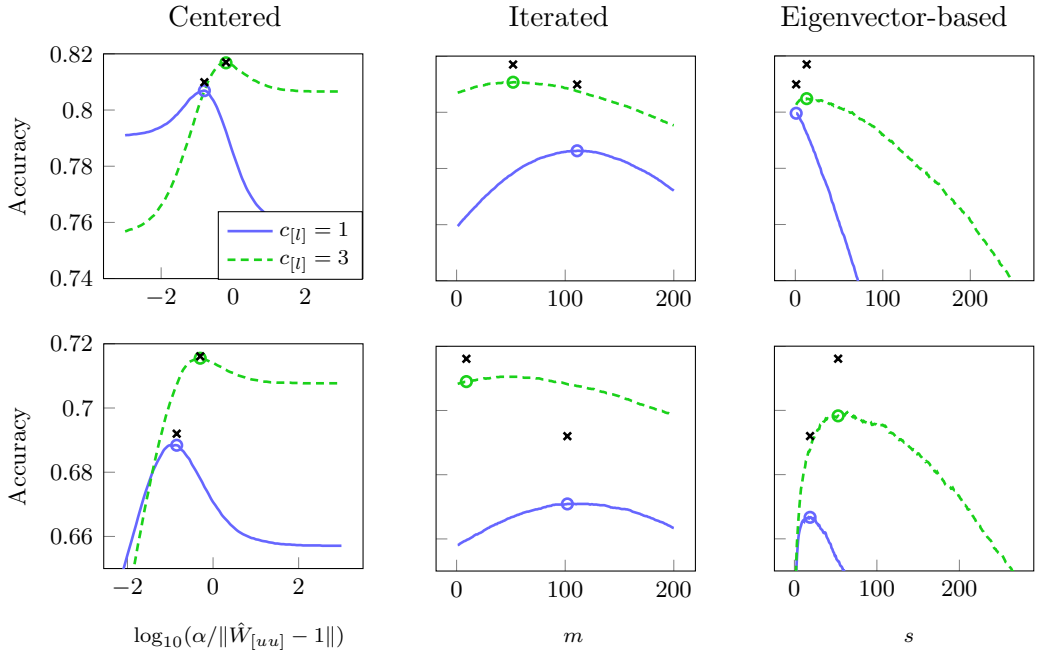

}

\subsection{Application of Centered Regularization on Sparse Graphs}
\label{sec:sparse graphs}

{\BLUE Constructing sparse graphs with good local geometry has been the focus of many research works in graph-based learning. In these graphs, a data point is only connected with non-zero weight to a small portion of other points, and connections within the same affinity group should be more frequent than between different groups. Sparse graphs are also natural objects in problems of community detection \citep{fortunato2010community}.  As our proposed algorithm involves a centering operation on the weight matrix \(W\), it disrupts the sparsity of the weight matrix, as well as the traditional concept of local geometry in sparse graphs. We may wonder about the computational efficiency (which benefits from the structure of sparse matrices) and the performance of the proposed method on sparse graphs, in comparison to the original Laplacian approach. 
	
In terms of computational efficiency, note that, even though the centered weight matrix \(\hat W\) is not sparse, it can be written as a sum of \(W\) and a matrix of rank two:
	
\begin{align*}
		\hat W= W+\begin{bmatrix}
		1_n &v
		\end{bmatrix}A\begin{bmatrix}
		1_n^\trans \\v^\trans
		\end{bmatrix}
\end{align*}
where \(v=W1_n\) and \(A=\begin{bmatrix}
(1_n^\trans W1_n)/n^2&-1/n\\
-1/n&0
\end{bmatrix}\). Using Woodbury's inversion formula, we can then decompose the inverse of \(\alpha I_{n_{[u]}}-\hat{W}_{[uu]}\) as the inverse of \(\alpha I_{n_{[u]}}-W_{[uu]}\) plus a matrix of rank two as:
\begin{align*}
	\left(\alpha I_{n_{[u]}}-\hat{W}_{[uu]}\right)^{-1}=Q-Q\begin{bmatrix}
	1_{n_{[u]}} &v_{[u]}
	\end{bmatrix}\left(A^{-1}+\begin{bmatrix}
	1_{n_{[u]}}^\trans \\v_{[u]}^\trans
	\end{bmatrix}Q\begin{bmatrix}
	1_{n_{[u]}} &v_{[u]}
	\end{bmatrix}\right)^{-1}\begin{bmatrix}
	1_{n_{[u]}}^\trans \\v_{[u]}^\trans
	\end{bmatrix}Q
\end{align*}
where \(Q=(\alpha I_{n_{[u]}}-W_{[uu]})^{-1}\). Therefore, the complexity of computing the solution of centered regularization can be reduced to that of computing \(QW_{[ul]}f_{[l]}\), which benefits from the sparsity of \(W\).

There remains the question of the learning performance on sparse graphs. Recall first from Subsection~\ref{sec:interpretation and properties} that the solution of centered regularization can be viewed as a stationary point of a label propagation through \(W\) with a centering operation on the input and output score vectors at each iteration. Naturally, the label propagation should be able to exploit the local geometry of \(W\), and we thus expect the centered regularization method to perform well on sparse graphs. To verify this claim, we test the centered regularization method, along with other graph-base SSL algorithms, over sparse graphs generated from stochastic block models (SBMs). SBMs are standard models for simply characterizing an underlying local geometry, where a pair of points \(x_i,x_j\) are connected (i.e., \(w_{ij}=1\)) with a probability of \(q_{\text{in}}\) if they belong to the same class and \(q_{\text{out}}\) otherwise. To account for heterogeneous degrees, the Degree-Corrected SBMs investigated in \citep{coja2010finding,gulikers2017spectral} propose to modify the probability of \(x_i,x_j\) being connected as $r_ir_jq_{\text{in}}$ for\(x_i,x_j\) in the same class and $r_ir_jq_{\text{out}}$ for \(x_i,x_j\) in different classes, with \(r_i\) reflecting the intrinsic connectivity of node \(i\). The results reported in Table~\ref{table:sparse graphs} show again a significant advantage of centered regularization over other methods across various ratios of labelled points, suggesting a highly competitive performance of the proposed method even on sparse graphs. We also observe that the centered regularization method tends to be more robust to heterogeneous degrees than other methods.
	
%There may be  a few data points with no connection to others in graphs of SBM, in this case, we replace them with new generations.

	\begin{table}
		\centering

		{	\small	\begin{tabular}{cccc}
			\hlinewd{1pt}
			$n_{[l]}/n$& $1/20$& $1/10$& $1/5$\\
			\hline \noalign{\smallskip}
			&\multicolumn{3}{c}{Case 1}
			 \\
			\noalign{\smallskip}\hline \noalign{\smallskip}
			Laplacian &$64.2\pm 2.6$ &$68.6\pm 2.2$  &$73.8\pm 2.1$ \\
			
			Centered & $\mathbf{70.7\pm 3.0}$ & $\mathbf{73.7\pm 2.4}$ &  $\mathbf{77.6\pm 2.0}$ \\
			
			Iterated Laplacian &$68.6\pm 3.0$ & $72.0\pm 2.6$ & $76.0\pm 2.3$\\
			
			Eigenvector-based  &$69.8\pm 3.3$ &$72.3\pm 2.6$ &$75.5\pm 2.2$  \\
			\noalign{\smallskip}\hline\noalign{\smallskip}
			&\multicolumn{3}{c}{Case 2}\\
			\noalign{\smallskip}\hline\noalign{\smallskip}
			Laplacian &$68.9\pm 2.9$ &$73.5\pm 2.2$  &$79.0\pm 1.7$ \\
			
			Centered &$\mathbf{80.7\pm 1.9}$ &$\mathbf{82.0\pm 1.8}$ &$\mathbf{83.9\pm 1.5}$ \\
			
			Iterated Laplacian &$77.2\pm 2.7$ & $79.2\pm 2.2$ & $82.3\pm 1.6$\\
			
			Eigenvector-based  &$78.4\pm3.1$ &$80.1\pm2.4$ &$82.3\pm 1.8$ \\
		\noalign{\smallskip}	\hlinewd{1pt}
			
		\end{tabular}}

	\caption{Accuracy of graph-based SSL algorithms on sparse graphs of SBMs. Averaged over \(1000\) realizations. Case 1: \(n=1000\), \(q_{\text{in}}=14/n\), \(q_{\text{out}}=7/n\), homogeneous degrees with \(r_i=1\)  for all \(i\in\{1,\ldots,n\}\). Case 2: \(n=1000\),  \(q_{\text{in}}=35/n\), \(q_{\text{out}}=15/n\), heterogeneous degrees with \(\mathbb{P}(r_i=0.3)=\frac{1}{4}\), \(\mathbb{P}(r_i=0.5)=\frac{1}{2}\) and \(\mathbb{P}(r_i=1)=\frac{1}{4}\).}
	\label{table:sparse graphs}
	\end{table}
	
}

\section{Concluding Remarks}
\label{sec:con}
The key to the proposed semi-supervised learning method lies in the replacement of conventional Laplacian regularizations by a centering operation on similarities. The motivation behind this operation is rooted in the large dimensional concentration of pairwise-data distances and thus likely to extend beyond the present graph-based semi-supervised learning schemes. It would in particular be interesting to know whether other advanced learning models involving Laplacian regularizations benefit from the same update. A specific example is Laplacian support vector machines (Laplacian SVMs) \citep{belkin2006manifold}, which is another widespread semi-supervised learning algorithm. Answering this question about Laplacian SVMs is however not a straightforward extension of the present analysis. Unlike the outcomes of Laplacian regularization, Laplacian SVMs are learned through an optimization problem without an explicit solution; additional technical tools, such as those recently devised in the work of \cite{el2013robust}, to deal with implicit objects are required for analyzing their performance. 

\medskip

As already anticipated by the theoretical results, it is not surprising that the proposed centered similarities regularization empirically produces large performance gains over the standard Laplacian regularization method when the aforementioned distance concentration problem is severe on the experimented data. However, it is quite illuminating to observe that even on datasets with weak distance concentration, for which the standard Laplacian approach exhibits a clear performance growth with respect to unlabelled data, the advantage of the proposed algorithm is still preserved. This attests to the general potential of such high dimensional studies for improving machine learning algorithms by identifying and settling some underlying issues compromising their learning performance, which would be difficult to spot if not through high dimensional analyses. 
%{\BLUE This observation also motivates further works to better understand the general advantage of this centering approach, and search for its possible limitations. Such investigations can be initiated, for instance, by checking the compatibility of the centering approach with other regularization techniques (e.g., the iterated Laplacian method\footnote{We refer interested readers to Appendix~\ref{app: iterated laplacian} for an extended discussion.}).}

\section{Acknowledgements}
Couillet's work is supported by the IDEX GSTATS and the MIAI ``GSTATS'' chairs at University Grenoble Alpes, as well as by the HUAWEI LarDist project.

\appendix
\section{Generalization of Theorem~\ref{th:statistics of fu for centered method} and Proof}
\subsection{Generalized Theorem}

We first present an extended version of Theorem~\ref{th:statistics of fu for centered method} for the general setting where $C_1$ may differ from $C_2$. {\BLUE The functions \(m(\xi)\), \(\sigma^2(\xi)\) defined in \eqref{eq:m(xi)} and \eqref{eq:sigma2(xi)} for describing the statistical distribution of unlabelled scores in the case of \(C_1=C_2\) need be adapted as follows:
\begin{align}
&m(\xi)=\frac{2c_{[l]}\theta(\xi)}{c_{[u]}\big(1-\theta(\xi)\big)}\label{eq:m(xi) general}\\
&\sigma^2(\xi)=\frac{\rho_1\rho_2(2c_{[l]}+m(\xi)c_{[u]})^2s(\xi)+\rho_1\rho_2(4c_{{l}}+m(\xi)^2c_{[u]})q(\xi)}{c_{[u]}\left(c_{[u]}-q(\xi)\right)},\quad k\in\{1,2\}\label{eq:sigma2(xi) general}
\end{align}
where
\begin{align}
\theta(\xi)&=\rho_1\rho_2\xi(\nu_1-\nu_2)^\trans\left(I_p-\xi \bar \Sigma\right)^{-1}(\nu_1-\nu_2)\nonumber\\
q(\xi)&=\xi^2p^{-1}\tr\left[\left(I_p-\xi \bar \Sigma\right)^{-1}\bar \Sigma\right]^2\nonumber\\
s(\xi)&=\rho_1\rho_2\xi^2(\nu_1-\nu_2)^\trans \left(I_p-\xi \bar \Sigma\right)^{-1}\bar\Sigma\left(I_p-\xi \bar \Sigma\right)^{-1}(\nu_1-\nu_2),\label{eq:omeage(xi),q(xi),s(xi)}
\end{align}
with 
\begin{align*}
\nu_k&=\begin{bmatrix}\sqrt{-2h'(\tau)}\mu_k^\trans &\sqrt{h''(\tau)}\tr{C_k}/\sqrt{p}\end{bmatrix}^\trans\\
\Sigma_k&=\begin{bmatrix}-2h'(\tau)C_k&0_{p\times1}\\ 0_{1\times p}&2h''(\tau)\tr{C_k}^2/p\end{bmatrix}
\end{align*}
and \(\bar \Sigma=\rho_1\Sigma_1+\rho_2\Sigma_2\). 

Notice that the adaptation is made here through the redefinitions of \(\theta(\xi)\), \(q(\xi\) and \(s(\xi)\); the expressions of \(m(\xi)\) and \(\sigma^2(\xi)\) are kept identical. As in the case of \(C_1=C_2\), the positive functions \(m(\xi)\) and \(\sigma^2(\xi)\) are defined respectively on the domains \((0,\xi_m)\) and \((0,\xi_{\sigma^2})\) with \(\xi_m,\xi_{\sigma^2}>0\) uniquely given  by \(\theta(\xi_m)=1\) and \(q(\xi_{\sigma^2})=c_{[u]}\). We define \(\xi_{\sup}\) as
\begin{equation}
\xi_{\sup}=\min\{\xi_m,\xi_{\sigma^2}\} \label{eq:xi_sup general}
\end{equation}

With these adapted notations,  we present the generalized results in the theorem below.

\begin{theorem}
	\label{th:statistics of fu for centered method (generalized)}
	Let Assumption~\ref{ass:data model} hold, the function \(h\) of \eqref{eq:definition W} be three-times continuously differentiable in a neighborhood of \(\tau\), \(f_{[u]}\) be the solution of \eqref{eq:centered kernel ssl} with fixed norm \(n_{[u]}e^2\), and with the notations of \(m(\xi)\), \(\sigma^2(\xi)\), \(\xi_{\sup}\) given in \eqref{eq:m(xi) general}, \eqref{eq:sigma2(xi) general}, \eqref{eq:xi_sup general}. Then, for \(n_{[l]}+1\leq i\leq n\) (i.e., \(x_i\) unlabelled) and \(x_i\in\mathcal{C}_k\),
	
	\begin{equation*}
	f_i\overset{\mathcal{L}}{\to}\mathcal{N}\left((-1)^k(1-\rho_k)\hat m,\hat\sigma_k^2\right)
	\end{equation*}
	where
	\begin{align*}
	&\hat m=m(\xi_e)\\
	 &\hat\sigma_k^2=c_{[u]}^{-2}\rho_1\rho_2\left[(2c_{[l]}+m(\xi_e)c_{[u]})^2s_k(\xi_e)+(4c_{{l}}+m(\xi_e)^2c_{[u]}+\sigma^2(\xi_e)c_{[u]})q(\xi_e)\right]
	\end{align*}
	with 
	\begin{align*}
	s_a(\xi_e)=\rho_1\rho_2\xi_e^2(\nu_1-\nu_2)^\trans \left(I_p-\xi_e \bar \Sigma\right)^{-1}\Sigma_k\left(I_p-\xi_e \bar \Sigma\right)^{-1}(\nu_1-\nu_2),\quad a\in\{1,2\},
	\end{align*}
	and \(\xi_e\in(0,\xi_{\sup})\) uniquely given by 
	\begin{align*}
		\rho_1\rho_2m(\xi_e)^2+\sigma^2(\xi_e)=e^2.
	\end{align*}
\end{theorem}}

\subsection{Proof of Generalized Theorem}
{\RED The proof of Theorem~\ref{th:statistics of fu for centered method (generalized)} relies on a leave-one-out approach, in the spirit of \citet{el2013robust}, along with arguments from previous related analyses \citep{COU16,mai2018random} based on random matrix theory .

\subsubsection{Main Idea} 

The main idea of the proof is to first demonstrate that for unlabelled data scores \(f_i\) (i.e., with \(i>n_{[l]}\)),
    \begin{align}
    \label{eq:main idea 1}
        f_i=\gamma\beta^{(i)\trans}\phi_c(x_i)+o_P(1)
    \end{align}
    where \(\gamma\) is a finite constant, \(\phi_c\) a certain mapping from the data space that we shall define, and \(\beta^{(i)}\) a random vector independent of \(\phi_c(x_i)\). Additionally, we shall show that
   {\BLUE \begin{align}
    \label{eq:main idea 2}
        \beta^{(i)}=\frac{1}{p}\sum_{j=1}^n f_j\phi_c(x_j)+\epsilon
    \end{align}}
     with \(\Vert \epsilon\Vert/\Vert \beta^{(i)}\Vert =o_P(1)\).

As a consequence of \eqref{eq:main idea 1}, the statistical behavior of the unlabelled data scores can be understood through that of \(\beta^{(i)}\), which itself depends on the unlabelled data scores as described by \eqref{eq:main idea 2}. By combining \eqref{eq:main idea 1} and \eqref{eq:main idea 2}, we thus establish the equations ruling the asymptotic statistical behavior (i.e., mean and variance) of the unlabelled data scores $f_i$.
}
\bigskip

\subsubsection{Detailed Arguments} 

In addition to the notations given in the end of the introduction (Section~\ref{sec:intro}), we specify that when multidimensional objects are concerned, $O(u_n)$  is understood entry-wise. The notation \(O_{\Vert\cdot\Vert}\) is understood as follows: for a vector $v$, $v=O_{\Vert\cdot\Vert}(u_n)$ means its Euclidean norm is $O(u_n)$ and for a square matrix $M$, $M=O_{\Vert\cdot\Vert}(u_n)$ means that the operator norm of $M$ is $O(u_n)$. 

\smallskip

First note that, as \(w_{ij}=h(\Vert x_i-x_j\Vert^2/p)=h(\tau)+O(p^{-\frac{1}{2}})\), Taylor-expanding \(w_{ij}\) around \(h(\tau)\) gives {\RED (see Appendix~\ref{app: hat W} for a detailed proof)} \( \hat{W}=O_{\Vert\cdot\Vert}(1)\) and
\begin{equation}\label{eq:hat W}
    \hat{W}=\frac{1}{p}\hat{\Phi}^\trans\hat{\Phi}+[h(0)-h(\tau)+\tau h'(\tau)]P_n+O_{\Vert\cdot\Vert}(p^{-\frac{1}{2}})\\
\end{equation}
where \(P_n=I_n-\frac{1}{n}1_n1_n^\trans\), and \(\hat{\Phi}=[\hat{\phi}(x_1),\ldots,\hat{\phi}(x_n)]=[\phi(x_1),\ldots,\phi(x_n)]P_n\) with
\begin{align*}
    \phi(x_i)=\begin{bmatrix}\sqrt{-2h'(\tau)}x_i^\trans& \sqrt{h''(\tau)}\Vert x_i\Vert^2/\sqrt{p}\end{bmatrix}^\trans.
\end{align*}

{\RED Define \(\nu_k=\mathbb{E}\{\phi(x_i)\}\), \(\Sigma_k={\rm cov}\{\phi(x_i)\}\) for \(x_i\in\mathcal{C}_k\), \(k\in\{1,2\}\), and let \(Z=[z_1,\ldots,z_n]\) with \(z_i=\phi(x_i)-\nu_k\) (i.e., \(\mathbb{E}\{z_i\}=0\)). We also write the labelled versus unlabelled divisions \(\Phi=\begin{bmatrix}\Phi_{[l]}&\Phi_{[u]}\end{bmatrix}\), \(Z=\begin{bmatrix}Z_{[l]}&Z_{[u]}\end{bmatrix}\) and \(\hat{\Phi}=\begin{bmatrix}\hat{\Phi}_{[l]}&\hat{\Phi}_{[u]}\end{bmatrix}\).}

Recall that \(f_{[u]}=\left(\alpha I_{n_{[u]}}-\hat{W}_{[uu]}\right)^{-1}\hat{W}_{[ul]}f_{[l]}\). {\RED To proceed, we need to show that \(\frac{1}{n}1_{n_{[u]}}^\trans f_{[u]}=O(p^{-\frac{1}{2}})\). Applying \eqref{eq:hat W}, we can express \(f_{[u]}\) as
\begin{align*}
    f_{[u]}=\left(\tilde \alpha I_{n_{[u]}}-\frac{1}{p}\hat{\Phi}_{[u]}^\trans\hat{\Phi}_{[u]}+\frac{r}{n}1_{n_{[u]}}1_{n_{[u]}}^\trans\right)^{-1}\left(\frac{1}{p}\hat{\Phi}_{[u]}^\trans\hat{\Phi}_{[l]}-\frac{r}{n}1_{n_{[u]}}1_{n_{[l]}}^\trans\right)f_{[l]}+O(p^{-\frac{1}{2}})
\end{align*}
where \(\tilde \alpha=\alpha-h(0)+h(\tau)-\tau h'(\tau)\), \(r=h(0)-h(\tau)+\tau h'(\tau)\). Since \(1_{[l]}^\trans f_{[l]}=0\) from its definition given in \eqref{eq:balanced f_l},
\begin{align}
\label{eq:fu}
    f_{[u]}=\left(\tilde \alpha I_{n_{[u]}}-\frac{1}{p}\hat{\Phi}_{[u]}^\trans\hat{\Phi}_{[u]}+\frac{r}{n}1_{n_{[u]}}1_{n_{[u]}}^\trans\right)^{-1}\frac{1}{p}\hat{\Phi}_{[u]}^\trans\Phi_{[l]}f_{[l]}+O(p^{-\frac{1}{2}}).
\end{align}
Write \(\hat{\Phi}_{[u]}=\mathbb{E}\{\hat{\Phi}_{[u]}\}+Z_{[u]}-(Z1_n/n)1_{n_{[u]}}^\trans\). Evidently, \(\mathbb{E}\{\hat{\Phi}_{[u]}\}=(\nu_1-\nu_2)s^\trans\) where \(s\in\mathbb{R}^{n_{[u]}}\) with \(s_i=(-1)^k(n-n_k)/n\) for \(x_i\in\mathcal{C}_k\), \(k\in\{1,2\}\). By the large number law, \(s=\zeta+O(p^{-\frac{1}{2}})\) where \(\zeta\in\mathbb{R}^{n_{[u]}}\) with \(\zeta_i=(-1)^k(1-\rho_k)\) for \(x_i\in\mathcal{C}_k\), therefore
\begin{align*}
    \frac{1}{p}\hat{\Phi}_{[u]}^\trans\hat{\Phi}_{[u]}=&\frac{1}{p}\bigg\{\Vert\nu_1-\nu_2\Vert^2 \zeta\zeta^\trans+Z_{[u]}^\trans Z_{[u]}+(1_n^\trans Z^\trans Z1_n/n^2)1_{n_{[u]}}1_{n_{[u]}}^\trans+[Z_{[u]}^\trans (\nu_1-\nu_2)]\zeta^\trans\\
    &+\zeta[Z_{[u]}^\trans (\nu_1-\nu_2)]^\trans-(Z_{[u]}^\trans Z1_n/n)1_{n_{[u]}}^\trans-1_{n_{[u]}}(Z_{[u]}^\trans Z1_n/n)^\trans\bigg\}+O_{\Vert\cdot\Vert}(p^{-\frac{1}{2}}).
\end{align*}
Invoking Woodbury’s identity \citep{woodbury1950inverting} expressed as
{\BLUE $$\left(R-UNU^\trans\right)^{-1}=R+RU(N^{-1}-U^\trans RU)^{-1}U^\trans R,$$
 we get \begin{align}
    \left(\tilde \alpha I_{n_{[u]}}-\frac{1}{p}\hat{\Phi}_{[u]}^\trans\hat{\Phi}_{[u]}+\frac{r}{n}1_{n_{[u]}}1_{n_{[u]}}^\trans\right)^{-1}&= \left(\tilde \alpha I_{n_{[u]}}-\frac{1}{p}Z_{[u]}^\trans Z_{[u]}-UNU^\trans\right)^{-1}+O_{\Vert\cdot\Vert}(p^{-\frac{1}{2}})\nonumber\\
    &=R+RU(N^{-1}-U^\trans RU)^{-1}U^\trans R+O_{\Vert\cdot\Vert}(p^{-\frac{1}{2}})\label{eq:resolvant}
\end{align}
by letting \(R=\left(\tilde \alpha I_{n_{[u]}}-\frac{1}{p}Z_{[u]}^\trans Z_{[u]}\right)^{-1}\) and \begin{align}
    &U=\frac{1}{\sqrt{p}}\begin{bmatrix}\zeta&Z_{[u]}^\trans (\nu_1-\nu_2)&1_{n_{[u]}}&Z_{[u]}^\trans Z1_n/n\end{bmatrix}\nonumber\\
    &N=\begin{bmatrix}\Vert\nu_1-\nu_2\Vert^2&1&0&0\\
    1&0&0&0\\
    0&0&(1_n^\trans Z^\trans Z1_n/n^2)-\frac{r}{c_0}&-1\\
    0&0&-1&0
    \end{bmatrix}.\label{eq:N}
\end{align} } Note also that \begin{align}
\frac{1}{p}\hat{\Phi}_{[u]}^\trans\Phi_{[l]}f_{[l]}= \sqrt{p}U \begin{bmatrix}(\nu_2-\nu_1)^\trans\frac{1}{p}\Phi_{[l]}f_{[l]}\\2c_{[l]}\rho_1\rho_2\\0\\0\end{bmatrix}+\frac{1}{p}Z_{[u]}^\trans Z_{[l]}f_{[l]}+O(p^{-\frac{1}{2}}).
\label{eq:right of resolvant}
\end{align}
{\BLUE Now we want to prove that  \(U^\trans RU\) is of the form
\begin{align}
\label{eq:form UtransRU}
    U^\trans RU=\begin{bmatrix}A&0_{2\times 2}\\
0_{2\times 2}&B \end{bmatrix}+O(p^{-\frac{1}{2}}),
\end{align}
for some matrices $A,B\in\RR^{2\times 2}$ with elements of $O(1)$. First it should be pointed out that \(z_i\) is a Gaussian vector if the last element is ignored. Since ignoring the last element of \(z_i\) will not change the concentration results given subsequently to prove the form of  \(U^\trans RU\), we shall treat \(z_i\) as Gaussian vectors for simplicity. As  there exists a deterministic matrix \(\bar R\) of the form \(c I_{n_{[u]}}\) such that 
\begin{align*}
	a^\trans R b-a^\trans \bar R b=O(p^{-\frac{1}{2}})
\end{align*}
for any \(a,b=O_{\Vert\cdot\Vert}(1)\) independent of \(R\) \citep[Proposition 5]{BEN15}, we get immediately that 
\begin{align*}
	U_{\cdot 1}^\trans R U_{\cdot 3}=\frac{1}{p} \zeta^\trans R 1_{n_{[u]}}=\frac{1}{p} \zeta^\trans \bar  R 1_{n_{[u]}}+O(p^{-\frac{1}{2}})=O(p^{-\frac{1}{2}}).
\end{align*}
In order to prove the rest, we begin by showing that
\begin{align}
\label{eq:aZuRb}
 \frac{1}{\sqrt{p}}a^\trans Z_{[u]}R b= O(p^{-\frac{1}{2}})
\end{align}
for any \(a,b=O_{\Vert\cdot\Vert}(1)\) independent of \(Z_{[u]}\). First let us set \(a'=\cov \{z_i\}^{\frac{1}{2}}a\) and denote by \(P_{a'}\) the projection matrix orthogonal to \(a'\).  We write then 
\begin{align*}
z_i&=\cov \{z_i\}^{\frac{1}{2}}P_{a'}\cov \{z_i\}^{-\frac{1}{2}}z_i+\cov \{z_i\}^{\frac{1}{2}}\frac{a'a^{\prime\trans}}{\Vert a'\Vert^2}\cov \{z_i\}^{-\frac{1}{2}}z_i\\
&=\tilde{z}_i+\frac{a^{\trans} z_i}{\Vert a'\Vert^2}\cov \{z_i\}a
\end{align*}
where \(\tilde{z}_i=\cov \{z_i\}^{\frac{1}{2}}P_{a'}\cov \{z_i\}^{-\frac{1}{2}}z_i\). Note that in this decomposition of \(z_i\), the two terms are independent. Indeed, since 
\begin{align*}
	\cov\{\tilde z_i,a^\trans z_i\}=\E\{\cov \{z_i\}^{\frac{1}{2}}P_{a'}\cov \{z_i\}^{-\frac{1}{2}}z_iz_i^\trans a\}=\cov \{z_i\}^{\frac{1}{2}}P_{a'}\cov \{z_i\}^{\frac{1}{2}} a=0_p,
\end{align*}
\(a^\trans z_i\) and \(\tilde z_i\) are uncorrelated, and thus independent by the property that uncorrelated jointly Gaussian variables are independent. Applying this decomposition of \(z_i\), we have, by letting \(\tilde Z=[\tilde z_1,\ldots,\tilde z_n]\) and \(q=[a^{\trans} z_1\Vert \cov \{z_1\}a\Vert/ \Vert a'\Vert^2, \ldots, a^{\trans} z_n\Vert \cov \{z_n\}a\Vert/ \Vert a'\Vert^2]\) , that 
\begin{align*}
	Z_{[u]}^\trans Z_{[u]}=\tilde Z_{[u]}^\trans \tilde Z_{[u]}+qq^\trans.
\end{align*}
Then with the help of Sherman-Morrison's formula \citep{sherman1950adjustment}, we get
\begin{align*}
	R=\tilde R -\frac{\tilde R qq^\trans \tilde R/p}{1+q^\trans \tilde R q/p}.
\end{align*}
Similarly to \(R\), we have also for \(\tilde R\)  a deterministic equivalent \(\bar{\tilde R}=\tilde c I_{n_{[u]}}\) with some constant \(\tilde c\) such that 
\begin{align*}
u^\trans \tilde R v-u^\trans \bar {\tilde R} v=O(p^{-\frac{1}{2}})
\end{align*}
for any \(u,v=O_{\Vert\cdot\Vert}(1)\) independent of \(\tilde R\) \citep[Proposition 5]{BEN15}. Since \( Z_{[u]}^\trans a\) and \(q\) are independent of \(\tilde R\), we prove \(\frac{1}{\sqrt{p}}a^\trans Z_{[u]}R b= O(p^{-\frac{1}{2}})\) with
\begin{align*}
	\frac{1}{\sqrt{p}}a^\trans Z_{[u]}R b&=\frac{1}{\sqrt{p}}a^\trans Z_{[u]}\tilde R b-\frac{\frac{1}{\sqrt{p}}a^\trans Z_{[u]}\tilde R qq^\trans \tilde Rb}{1+q^\trans \tilde R q}\\
	&=\frac{1}{\sqrt{p}}\tilde c a^\trans Z_{[u]} b-\frac{\frac{1}{\sqrt{p}}\tilde c^2 a^\trans Z_{[u]}qq^\trans b}{1+\tilde c \Vert q\Vert^2}+O(p^{-\frac{1}{2}})\\
	&=O(p^{-\frac{1}{2}}).
\end{align*} This leads directly to 
\begin{align*}
	U_{\cdot 2}^\trans R U_{\cdot 3}=\frac{1}{\sqrt{p}}(\nu_1-\nu_2)Z_{[u]}R 1_{n_{[u]}}/\sqrt{p}=O(p^{-\frac{1}{2}}).
\end{align*} 
With the same argument, we have also 
\begin{align*}
U_{\cdot 1}^\trans R U_{\cdot 4}&=\frac{1}{p}\zeta^\trans R \left(Z_{[u]}^\trans Z_{[u]} 1_{n_{[u]}}/n+Z_{[u]}^\trans Z_{[l]} 1_{n_{[l]}}/n\right)\\
&=\zeta^\trans(\tilde\alpha R-I_{n_{[u]}})1_{n_{[u]}}/n+\frac{1}{p}\zeta^\trans RZ_{[u]}^\trans \left(Z_{[l]} 1_{n_{[l]}}/n\right)=O(p^{-\frac{1}{2}});
\end{align*}
and
\begin{align*}
	U_{\cdot 2}^\trans R U_{\cdot 4}=&\frac{1}{p}(\nu_1-\nu_2)^\trans Z_{[u]} R \left(Z_{[u]}^\trans Z_{[u]} 1_{n_{[u]}}/n+Z_{[u]}^\trans Z_{[l]} 1_{n_{[l]}}/n\right)\\
	=&\tilde\alpha(\nu_1-\nu_2)^\trans Z_{[u]} R1_{n_{[u]}}/n-(\nu_1-\nu_2)^\trans Z_{[u]}1_{n_{[u]}}/n\\
     &+(\nu_1-\nu_2)^\trans\left(\frac{1}{p} Z_{[u]}RZ_{[u]}^\trans\right) \left(Z_{[l]} 1_{n_{[l]}}/n\right)    =O(p^{-\frac{1}{2}}).
\end{align*}
We conclude thus that \(U^\trans RU\) is of the form \eqref{eq:form UtransRU}.

Substituting \eqref{eq:resolvant} and \eqref{eq:right of resolvant} into \eqref{eq:fu} and using the fact that \(p^{-\frac{3}{2}}\Vert U^\trans RZ_{[u]}^\trans Z_{[l]}f_{[l]}\Vert=O(p^{-\frac{1}{2}})\) derived by similar reasoning to the above, we obtain 
\begin{align}
\label{eq:1trans fu}
    \frac{1}{n}1_{n_{[u]}}^\trans f_{[u]}=c_0^{-1}\begin{bmatrix}0&0&1&0\end{bmatrix}K\begin{bmatrix}(\nu_2-\nu_1)^\trans\frac{1}{p}\Phi_{[l]}f_{[l]}\\2c_{[l]}\rho_1\rho_2\\0\\0\end{bmatrix}+O(p^{-\frac{1}{2}})
\end{align}
with 
\begin{align*}
    K=U^\trans RU+U^\trans RU(N^{-1}-U^\trans RU)^{-1}U^\trans RU.
\end{align*}
Since \(U^\trans RU\) is of the form \eqref{eq:form UtransRU}, we find from classical algebraic arguments that \(K\) is also of the same diagonal block matrix form. We thus finally get from \eqref{eq:1trans fu} that 
\begin{align*}
	\frac{1}{n}1_{n_{[u]}}^\trans f_{[u]}=O(p^{-\frac{1}{2}}).
\end{align*}

}

Now that we have shown that \(\frac{1}{n}1_{n_{[u]}}^\trans f_{[u]}=O(p^{-\frac{1}{2}})\), multiplying both sides of \eqref{eq:fu} with \(\tilde \alpha I_{n_{[u]}}-\frac{1}{p}\hat{\Phi}_{[u]}^\trans\hat{\Phi}_{[u]}+\frac{r}{n}1_{n_{[u]}}1_{n_{[u]}}^\trans\) from the left gives
\begin{align*}
    \tilde \alpha f_{[u]}&=\frac{1}{p}\hat{\Phi}_{[u]}^\trans\hat{\Phi}_{[u]}f_{[u]}+\frac{1}{p}\hat{\Phi}_{[u]}^\trans\hat{\Phi}_{[l]}f_{[l]}+O(p^{-\frac{1}{2}}).
\end{align*}

Decomposing this equation for any \(i>n_{[l]}\) (i.e., \(x_i\) unlabelled) leads to 
\begin{align}
    \tilde \alpha f_{i}&=\frac{1}{p}\hat{\phi}(x_i)^\trans\hat{\Phi}f+O(p^{-\frac{1}{2}})\label{eq:fi-def} \\
    \tilde \alpha f_{[u]}^{\{i\}}&=\frac{1}{p}\hat{\Phi}_{[u]}^{\{i\}\trans}\hat{\phi}(x_i)f_i+\frac{1}{p}\hat{\Phi}_{[u]}^{\{i\}\trans}\hat{\Phi}_{[u]}^{\{i\}}f_{[u]}^{\{i\}}+\frac{1}{p}\hat{\Phi}_{[u]}^{\{i\}\trans}\hat{\Phi}_{[l]}f_{[l]}+O(p^{-\frac{1}{2}})\label{eq:f-i}
\end{align}
with \(f_{[u]}^{\{i\}}\) standing for the vector obtained by removing \(f_i\) from \(f_{[u]}\), \(\hat{\Phi}_{[u]}^{\{i\}}\) for the matrix obtained by removing \(\hat{\phi}(x_i)\) from \(\hat{\Phi}_{[u]}\).

\bigskip

Our objective is to compare the behavior of the vector $f_{[u]}$ decomposed as $\{f_i,f_{[u]}^{\{i\}}\}$ to the ``leave-$x_i$-out'' version $f_{[u]}^{(i)}$ to be introduced next. To this end, define the leave-one-out dataset \(X^{(i)}=\{x_1,\ldots,x_{i-1},x_{i+1},\ldots,x_n\}\in\mathbb{R}^{(n-1)\times p}\) for any \(i>n_{[l]}\) (i.e., \(x_i\) unlabelled), and \(\hat{W}^{(i)}\in\mathbb{R}^{(n-1)\times(n-1)}\) the corresponding centered similarity matrix, for which we have, similarly to \(\hat{W}\),
\begin{equation}
\label{eq:W(i)}
    \hat{W}^{(i)}=\frac{1}{p}\hat{\Phi}^{(i)\trans}\hat{\Phi}^{(i)}+[h(0)-h(\tau)+\tau h'(\tau)]P_{n-1}+O_{\Vert\cdot\Vert}(p^{-\frac{1}{2}})
\end{equation}
where \(\hat{\Phi}^{(i)}=[\hat{\phi}^{(i)}(x_1),\ldots,\hat{\phi}^{(i)}(x_{i-1}),\hat{\phi}^{(i)}(x_{i+1}),\ldots,\hat{\phi}^{(i)}(x_n)]=[\phi(x_1),\ldots,\phi(x_{i-1}),\phi(x_{i+1}),\)
\(\ldots,\phi(x_n)]P_{n-1}\).
{\RED Denote by \(f_{[u]}^{(i)}\) the solution of the centered similarities regularization on the ``leave-one-out'' dataset \(X_{(i)}\), i.e.,
\begin{align}
\label{eq:f(i)-def}
f_{[u]}^{(i)}=\left(\alpha I_{n_{[u]}-1}-\hat{W}_{[uu]}^{(i)}\right)^{-1}\hat{W}_{[ul]}^{(i)}f_{[l]}.
\end{align}
Substituting \eqref{eq:W(i)} into \eqref{eq:f(i)-def} leads to}
\begin{align}
\tilde \alpha f_{[u]}^{(i)}=&\frac{1}{p}\hat{\Phi}_{[u]}^{(i)\trans}\hat{\Phi}_{[u]}^{(i)}f_{[u]}^{(i)}+\frac{1}{p}\hat{\Phi}_{[u]}^{(i)\trans}\hat{\Phi}_{[l]}f_{[l]}+O(p^{-\frac{1}{2}})\label{eq:f(i)}
\end{align}
where \(\hat{\Phi}^{(i)}=\begin{bmatrix}\hat{\Phi}_{[l]}^{(i)}&\hat{\Phi}_{[u]}^{(i)}\end{bmatrix}\). {\BLUE From the definitions of $\hat\Phi_{[u]}^{(i)}$ and $\hat{\Phi}_{[u]}^{\{i\}}$, which essentially differ by the addition of the $O(1/\sqrt{p})$-norm term $\phi(x_i)/n$ to every column, we easily have
\begin{equation*}
\frac{1}{\sqrt{p}}\hat{\Phi}_{[u]}^{(i)}- \frac{1}{\sqrt{p}}\hat{\Phi}_{[u]}^{\{i\}}=O_{\|\cdot\|}(p^{-1}),
\end{equation*}
which entails
\begin{equation}
\label{eq:phi-i-phi(i)}
\frac{1}{p}\hat{\Phi}_{[u]}^{(i)\trans}\hat{\Phi}_{[u]}^{(i)}- \frac{1}{p}\hat{\Phi}_{[u]}^{\{i\}\trans}\hat{\Phi}_{[u]}^{\{i\}}=O_{\|\cdot\|}(p^{-1}),
\end{equation}}

Thus, subtracting \eqref{eq:f(i)} from \eqref{eq:f-i} gives
\begin{equation}
\label{eq:f-i-f(i)}
    M^{(i)}\left(f_{[u]}^{\{i\}}-f_{[u]}^{(i)}\right)=\frac{1}{p}\hat{\Phi}_{[u]}^{(i)\trans}\hat{\phi}(x_i)f_i+O(p^{-\frac{1}{2}})
\end{equation}
with \begin{align*}
    M^{(i)}=\tilde\alpha I_{(n_{[u]}-1)}-\frac{1}{p}\hat{\Phi}_{[u]}^{(i)\trans}\hat{\Phi}_{[u]}^{(i)}.
\end{align*}
Set \(\beta=\frac{1}{p}\hat{\Phi}f=O_{\Vert\cdot\Vert}(1)\), the unlabelled data ``regression vector'' {\BLUE which gives unlabelled data scores by \(f_i= \tilde\alpha^{-1}\beta^\trans \hat\phi(x_i)\)}, and its ``leave-one-out'' version \(\beta^{(i)}=\frac{1}{p}\hat{\Phi}^{(i)}f^{(i)}\) with \(f^{(i)}=\begin{bmatrix}f_{[l]}&f_{[u]}^{(i)}\end{bmatrix}\). {\RED Applying \eqref{eq:phi-i-phi(i)}
and \eqref{eq:f-i-f(i)}, we get that
\begin{align}
\label{eq:beta-beta(i)}
    \beta - \beta^{(i)}=\left(I_p+\frac{1}{p}\hat{\Phi}_{[u]}^{(i)}\left(M^{(i)}\right)^{-1}\hat{\Phi}_{[u]}^{(i)\trans}\right)\frac{1}{p}f_i\hat{\phi}(x_i)+O_{\Vert\cdot\Vert}(p^{-1})=O_{\Vert\cdot\Vert}(p^{-\frac{1}{2}}).
\end{align}
By the above result, Equation \eqref{eq:fi-def} can be expanded as
\begin{align}
\label{eq:f_i(1)}
    \tilde \alpha f_i=&\beta^{(i)\trans}\hat{\phi}(x_i)+\frac{1}{p}\hat{\phi}(x_i)^\trans\left(I_p+\frac{1}{p}\hat{\Phi}_{[u]}^{(i)}\left(M^{(i)}\right)^{-1}\hat{\Phi}_{[u]}^{(i)\trans}\right)\hat{\phi}(x_i)f_i+O(p^{-\frac{1}{2}}).
\end{align}
To go further in the development of \eqref{eq:f_i(1)}, we first need to evaluate the quadratic form
\begin{align*}
    \kappa_i\equiv \frac{1}{p}\hat{\phi}(x_i)^\trans T^{(i)}\hat{\phi}(x_i)
\end{align*}
where \begin{align*}
    T^{(i)}=I_p+\frac{1}{p}\hat{\Phi}_{[u]}^{(i)}\left(M^{(i)}\right)^{-1}\hat{\Phi}_{[u]}^{(i)\trans}.
\end{align*}
{\BLUE Since \(\frac{1}{p}\hat{\Phi}_{[u]}^{(i)\trans}\hat{\Phi}_{[u]}^{(i)}=O_{\Vert \cdot\Vert}(1)\), it is easy to see that \(T^{(i)}=O_{\Vert \cdot\Vert}(1)\). As \(\hat{\phi}(x_i)\) is independent of \(T^{(i)}\), it unfolds from the ``trace lemma'' \citep[Theorem 3.4]{couillet2011random} that 
\begin{align*}
	\kappa_i-\frac{1}{p}\tr\Sigma_kT^{(i)}\asto 0.
\end{align*}

Notice that
\begin{align*}
	T^{(i)}&=\tilde \alpha \left(\tilde\alpha I_p-\frac{1}{p}\hat{\Phi}_{[u]}^{(i)}\hat{\Phi}_{[u]}^{(i)\trans}\right)^{-1}=\tilde \alpha \left(\tilde\alpha I_p-\frac{1}{p}\hat{\Phi}_{[u]}^{\{i\}}\hat{\Phi}_{[u]}^{\{i\}\trans}\right)^{-1}+O_{\|\cdot\|}(p^{-1})\\
	&=T-\frac{\frac{\tilde\alpha}{p}T^{(i)}\hat{\phi}(x_i)\hat{\phi}(x_i)^\trans T^{(i)}}{1-\frac{1}{p}\frac{\kappa_i}{\tilde\alpha}}+O_{\|\cdot\|}(p^{-1})
\end{align*}
where 
\begin{align*}
	T=\tilde \alpha \left(\tilde\alpha I_p-\frac{1}{p}\hat{\Phi}_{[u]}\hat{\Phi}_{[u]}^{\trans}\right)^{-1}=T^{(i)}+\frac{\frac{\tilde\alpha}{p}T^{(i)}\hat{\phi}(x_i)\hat{\phi}(x_i)^\trans T^{(i)}}{1-\frac{1}{p}\frac{\kappa_i}{\tilde\alpha}}
\end{align*}
by Sherman-Morrison's formula  \citep{sherman1950adjustment}.  We get consequently
\begin{align*}
	\frac{1}{p}\tr\Sigma_kT^{(i)}=\frac{1}{p}\tr\Sigma_kT+O(p^{-1}),
\end{align*}
$\kappa_i$ converges thus  to a deterministic limit \(\kappa\) independent of $i$ at large \(n,p\). }

Equation \eqref{eq:f_i(1)} then becomes
\begin{align}
\label{eq:f_i(2)}
    f_i=\gamma\beta^{(i)\trans}\hat{\phi}(x_i)+O(p^{-\frac{1}{2}}).
\end{align}
where \(\gamma=(\tilde\alpha-\kappa)^{-1}\).}

{\RED We focus now on the term \(\beta^{(i)\trans}\hat{\phi}(x_i)\) in \eqref{eq:f_i(2)}. To discard the ``weak'' dependence between \(\beta^{(i)\trans}\) and \(\hat{\phi}(x_i)\),  let us define
\begin{align*}
    \phi_c(x_i)=(-1)^k(1-\rho_k)(\nu_2-\nu_1)+z_i.
\end{align*}
As \(n_k/n=\rho_k+O(n^{-\frac{1}{2}})\), by the law of large numbers, \(\mathbb{E}\{\hat{\phi}(x_i)\}=(-1)^k[(n-n_k)/n](\nu_2-\nu_1)=\mathbb{E}\{\phi_c(x_i)\}+O_{\Vert \cdot\Vert}(n^{-\frac{1}{2}})\). Remark that, unlike \(\hat{\phi}(x_i)\), \(\phi_c(x_i)\) is independent of all \(x_j\) with \(j\neq i\), and therefore independent of \(\beta^{(i)}\). We thus now have \begin{align*}
    \beta^{(i)\trans}\hat{\phi}(x_i)&=\beta^{(i)\trans}\bigg(\mathbb{E}\{\hat{\phi}(x_i)\}+z_i-\frac{1}{n}\sum_{m=1}^nz_m\bigg)=\beta^{(i)\trans}\phi_c(x_i)+\frac{1}{n}\beta^{\trans} Z1_{n}+O(p^{-\frac{1}{2}}).
\end{align*}
We get from \eqref{eq:beta-beta(i)} that \(\frac{1}{n}\beta^{(i)\trans} Z1_{n}=\frac{1}{n}\beta^{\trans} Z1_{n}+O(p^{-\frac{1}{2}})\), leading to
\begin{align}
\label{eq:f_i(3)}
f_i=\gamma \beta^{(i)\trans}\phi_c(x_i)+\frac{1}{n}\beta^{\trans} Z1_{n}+O(p^{-\frac{1}{2}}).
\end{align}
Since \(\phi_c(x_i)\) is independent of \(\beta^{(i)}\), according to the central limit theorem, \(\beta^{(i)\trans}\phi_c(x_i)\) asymptotically follows a Gaussian distribution. 

To demonstrate that \(\frac1n\beta^\trans Z1_n\) is negligibly small, notice fist that, by summing \eqref{eq:f_i(3)} for all \(i>n_{[u]}\), we have 
\begin{align*}
    \frac{1}{n}1_{n_{[u]}}^\trans f_{[u]}=\frac{1}{n}\sum_{i=n_{[l]}+1}^n \beta^{(i)\trans}\phi_c(x_i)+c_{[u]}(\beta^{(i)\trans} Z1_{n}/n)+O(p^{-\frac{1}{2}}).
\end{align*}
Since \(\frac{1}{n}1_{n_{[u]}}^\trans f_{[u]}=O(p^{-\frac{1}{2}})\), it suffices to prove \(\frac{1}{n}\sum_{i=n_{[l]}+1}^n \beta^{(i)\trans}\phi_c(x_i)=O(p^{-\frac{1}{2}})\) to consequently show that \(\frac1n\beta^\trans Z1_n=O(p^{-\frac{1}{2}})\) from the above equation. To this end, we shall examine the correlation between \(\beta^{(i)\trans}\phi_c(x_i)\) and \(\beta^{(j)\trans}\phi_c(x_j)\) for \(i\neq j>n_{[l]}\). Consider \(\beta^{(ij)}, \hat{\Phi}_{[u]}^{(ij)},M^{(ij)}\) obtained in the same way as \(\beta^{(i)}, \hat{\Phi}_{[u]}^{(i)},M^{(i)}\), but this time by leaving out the two unlabelled samples \(x_i,x_j\). Similarly to \eqref{eq:beta-beta(i)}, we have 
\begin{align}
\label{eq:beta(i)-beta(ij)}
    \beta^{(i)} - \beta^{(ij)}=\left(I_p+\frac{1}{p}\hat{\Phi}_{[u]}^{(ij)}\left(M^{(ij)}\right)^{-1}\hat{\Phi}_{[u]}^{(ij)\trans}\right)\frac{1}{p}f_j\hat{\phi}(x_j)+O_{\Vert\cdot\Vert}(p^{-1})=O_{\Vert\cdot\Vert}(p^{-\frac{1}{2}}). 
\end{align} It follows from the above equation that, for \(i\neq j>n_{[l]}\),
\begin{align}
\label{eq:var gigj}
    &\quad\cov\{\beta^{(i)\trans}\phi_c(x_i),\beta^{(i)\trans}\phi_c(x_j)\}\nonumber\\
    &=\mathbb{E}\{\beta^{(i)\trans}\phi_c(x_i)\beta^{(i)\trans}\phi_c(x_j)\}-\mathbb{E}\{\beta^{(i)\trans}\phi_c(x_i)\}\mathbb{E}\{\beta^{(j)\trans}\phi_c(x_j)\}\nonumber\\
    &=\mathbb{E}\{\beta^{(ij)\trans}\phi_c(x_i)\beta^{(ij)\trans}\phi_c(x_j)\}-\mathbb{E}\{\beta^{(i)\trans}\phi_c(x_i)\}\mathbb{E}\{\beta^{(j)\trans}\phi_c(x_j)\}+O(p^{-1})\nonumber\\
    &=\mathbb{E}\{\beta^{(ij)\trans}\phi_c(x_i)\}\mathbb{E}\{\beta^{(ij)\trans}\phi_c(x_j)\}-\mathbb{E}\{\beta^{(i)\trans}\phi_c(x_i)\}\mathbb{E}\{\beta^{(j)\trans}\phi_c(x_j)\}+O(p^{-1})\nonumber\\
    &=O(p^{-1}),
\end{align}
leading to the conclusion that \(\frac{1}{n_{[u]}}\sum_{i=n_{[l]}+1}^n \beta^{(i)\trans}\phi_c(x_i)=\frac{1}{n_{[u]}}\sum_{i=n_{[l]}+1}^n \mathbb{E}\{\beta^{(i)\trans}\phi_c(x_i)\}+O(p^{-\frac{1}{2}})=O(p^{-\frac{1}{2}})\). Hence, \(\frac{1}{n}\beta^{\trans} Z1_{n}=O(p^{-\frac{1}{2}})\). Finally, we have that, for \(i>n_{[l]}\),
\begin{align}
\label{eq:f_i}
    f_i=\gamma \beta^{(i)\trans}\phi_c(x_i)+O(p^{-\frac{1}{2}}),
\end{align}
indicating that, up to the constant $\gamma$, \(f_i\) asymptotically follows the same Gaussian distribution as \(\beta^{(i)\trans}\phi_c(x_i)\).
}

Moreover, taking the expectation and the variance of the both sides of \eqref{eq:f_i} for \(x_i\in\mathcal{C}_k\) yields
\begin{align*}
    \mathbb{E}\{f_i\vert i>n_{[l]},x\in\mathcal{C}_k\}&=\gamma\mathbb{E}\{\beta^{(i)\trans}\}(-1)^k(1-\rho_k)(\nu_2-\nu_1)+O(p^{-\frac{1}{2}})\\
    {\rm var}\{f_i\vert i>n_{[l]},x\in\mathcal{C}_k\}&=\gamma^2{\rm tr}\big[{\rm cov}\{\beta^{(i)}\}\Sigma_k\big]+\gamma^2 \mathbb{E}\{\beta^{(i)}\}^\trans\Sigma_k\mathbb{E}\{\beta^{(i)}\}+O(p^{-\frac{1}{2}}).
\end{align*}
{\RED Since \(\beta-\beta^{(i)}=O_{\Vert\cdot\Vert}(p^{-\frac{1}{2}})\)  as per \eqref{eq:beta-beta(i)}, we obtain
\begin{align}
    \mathbb{E}\{f_i\vert i>n_{[l]},x\in\mathcal{C}_k\}&=\gamma\mathbb{E}\{\beta^\trans\}(-1)^k(1-\rho_k)(\nu_2-\nu_1)+O(p^{-\frac{1}{2}})\label{eq:E fi}\\
    {\rm var}\{f_i\vert i>n_{[l]},x\in\mathcal{C}_k\}&=\gamma^2{\rm tr}\big[{\rm cov}\{\beta\}\Sigma_k\big]+\gamma^2 \mathbb{E}\{\beta\}^\trans\Sigma_k\mathbb{E}\{\beta\}+O(p^{-\frac{1}{2}})\label{eq:var fi}.
\end{align}}

\smallskip

After linking the distribution parameters of unlabelled scores to those of \(\beta\) with Equation \eqref{eq:E fi} and Equation \eqref{eq:var fi}, we now turn our attention to the statistical behaviour of \(\beta\). Substituting \eqref{eq:f_i} into \(\beta=\frac{1}{p}\hat{\Phi}f\) yields
\begin{align}
    \beta&=\frac{1}{p}\sum_{i=1}^{n_{[l]}}f_i\hat{\phi}(x_i)+\frac{1}{p}\sum_{i=n_{[l]}+1}^n\gamma\beta^{(i)\trans}\phi_c(x_i)\hat{\phi}(x_i)+O_{\Vert\cdot\Vert}(p^{-\frac{1}{2}})\nonumber\\
    &=\frac{1}{p}\sum_{i=1}^{n_{[l]}}f_i\phi_c(x_i)+\frac{1}{p}\sum_{i=n_{[l]}+1}^n\gamma\beta^{(i)\trans}\phi_c(x_i)\phi_c(x_i)+O_{\Vert\cdot\Vert}(p^{-\frac{1}{2}})\label{eq:beta(1)}.
\end{align}
{\RED For \(i>n_{[l]}\) and \(x_i\in\mathcal{C}_k\), we decompose \(\phi_c(x_i)\) as
\begin{align}
\label{eq:decomposition phi_c}
\phi_c(x_i)=\mathbb{E}\{\phi_c(x_i)\}+\frac{\Sigma_k\beta^{(i)}}{\beta^{(i)\trans} z_i}+\tilde{z}_i
\end{align}
where 
\begin{align*}
    \tilde{z}_i= z_i-\frac{\Sigma_k\beta^{(i)}}{\beta^{(i)\trans} z_i}.
\end{align*}
By substituting the expression \eqref{eq:decomposition phi_c} of \(\phi_c(x_i)\) into \eqref{eq:beta(1)} and using the fact that \(\beta-\beta^{(i)}=O_{\Vert\cdot\Vert}(p^{-\frac{1}{2}})\), we obtain
\begin{align}
\label{eq:beta}
    \bigg(I_p-\gamma c_{[u]}\sum_{a=1}^2\rho_a\Sigma_a\bigg)\beta&=\frac{1}{p}\sum_{i=1}^{n_{[l]}}f_i\mathbb{E}\{\phi_c(x_i)\}+\frac{1}{p}\sum_{i=n_{[l]}+1}^n\gamma\beta^{(i)\trans} \phi_c(x_i)\mathbb{E}\{\phi_c(x_i)\}\nonumber\\
    &+\frac{1}{p}\sum_{i=1}^{n_{[l]}}f_iz_i+\frac{1}{p}\sum_{i=n_{[l]}+1}^n\gamma\beta^{(i)\trans} \phi_c(x_i)\tilde{z}_i +O_{\Vert\cdot\Vert}(p^{-\frac{1}{2}}).
\end{align}

Recall that \(f_{[l]}\) is a deterministic vector (given in \eqref{eq:balanced f_l}) and note that 
\begin{align*}
    \mathbb{E}\{\beta^{(i)\trans}\phi_c(x_i)\tilde{z}_i\}=\mathbb{E}\{\beta^{(i)\trans} z_i [z_i-\Sigma_k\beta^{(i)}/(\beta^{(i)\trans} z_i)]\}=\mathbb{E}\{\beta^{(i)\trans} z_i z_i\}-\Sigma_k\mathbb{E}\{\beta^{(i)}\}=0.
\end{align*}
Taking the expectation of both sides of \eqref{eq:beta} thus gives 
\begin{align}
    &\quad\bigg(I_p-\gamma c_{[u]}\sum_{a=1}^2\rho_a\Sigma_a\bigg)\mathbb{E}\{\beta\}\nonumber\\
    &=\frac{1}{p}\sum_{i=1}^{n_{[l]}}f_i\mathbb{E}\{\phi_c(x_i)\}+\frac{1}{p}\sum_{i=n_{[l]}+1}^n\gamma\mathbb{E}\{\beta^{(i)}\}^\trans \mathbb{E}\{\phi_c(x_i)\}\mathbb{E}\{\phi_c(x_i)\}+O_{\Vert\cdot\Vert}(p^{-\frac{1}{2}})\nonumber\\
    &=\frac{1}{p}\sum_{i=1}^{n_{[l]}}f_i\mathbb{E}\{\phi_c(x_i)\}+\frac{1}{p}\sum_{i=n_{[l]}+1}^n\gamma\mathbb{E}\{\beta\}^\trans
     \mathbb{E}\{\phi_c(x_i)\}\mathbb{E}\{\phi_c(x_i)\}+O_{\Vert\cdot\Vert}(p^{-\frac{1}{2}}).\label{eq:E beta}
\end{align}
Let \(Q=I_p-\gamma c_{[u]}\bar \Sigma\) with \(\bar \Sigma=\rho_1\Sigma_1+\rho_2\Sigma_2\) and denote \(\hat m\equiv \gamma(\nu_2-\nu_1)^\trans\mathbb{E}\{\beta\}\). With these notations, we get directly from the above equation that
\begin{align}
    \hat m=\gamma\rho_1\rho_2(2c_{[l]} + mc_{[u]})(\nu_2-\nu_1)^\trans Q^{-1}(\nu_2-\nu_1)+o_P(1).\label{eq:m}
\end{align}
With the notation $m$, \eqref{eq:E fi} notably becomes
\begin{align*}
    \mathbb{E}\{f_i\vert i>n_{[l]},x\in\mathcal{C}_k\}=(-1)^k(1-\rho_k)\hat m+O(p^{-\frac{1}{2}}).
\end{align*}
In addition, we get from \eqref{eq:E beta} that 
\begin{align}
  \gamma^2 \mathbb{E}\{\beta\}^\trans\Sigma_k\mathbb{E}\{\beta\}= \left[\gamma\rho_1\rho_2(2c_{[l]} + \hat mc_{[u]})\right]^2(\nu_2-\nu_1)^\trans Q^{-1} \Sigma_k Q^{-1} (\nu_2-\nu_1).\label{eq:E beta trans Sigma E beta}
\end{align}
Furthermore, we have from \eqref{eq:beta} and \eqref{eq:E beta} 
\begin{align*}
    {\rm tr}[{\rm cov}\{\beta\}\Sigma_k]&=\mathbb{E}\left\{(\beta-\mathbb{E}\{\beta\})^\trans\Sigma_k(\beta-\mathbb{E}\{\beta\})\right\}\\
    &=\frac{1}{p^2}\sum_{i=1}^{n_{[l]}}f_i^2\mathbb{E}\{z_i^\trans Q^{-1}\Sigma_kQ^{-1}z_i\}+\frac{1}{p^2}\sum_{i=n_{[l]}+1}^{n}\gamma^2\mathbb{E}\{(\beta^{(i)\trans} \phi_c(x_i))^2\tilde z_i^\trans Q^{-1}\Sigma_kQ^{-1}\tilde z_i\}\\
    &+O(p^{-\frac{1}{2}}).
\end{align*}
Since \(\frac{1}{p}z_i^\trans Q^{-1}\Sigma_kQ^{-1}z_i=\frac{1}{p}{\rm tr}(Q^{-1}\bar \Sigma)^2+O(p^{-\frac{1}{2}})\) and \(\frac{1}{p}\tilde z_i^\trans Q^{-1}\Sigma_kQ^{-1}\tilde z_i=\frac{1}{p}{\rm tr}(Q^{-1}\bar \Sigma)^2+O(p^{-\frac{1}{2}})\), by the trace lemma \citep[Theorem 3.4]{couillet2011random} and Assumption~\ref{ass:data model}, 
\begin{align}
    \gamma^2{\rm tr}[{\rm cov}\{\beta\}\Sigma_k]=&\gamma^2\big[\rho_1\rho_2(4c_{[l]}+\hat m^2c_{[u]})+c_{[u]}\sum_{a=1}^2\rho_a{\rm var}\{f_i\vert i>n_{[l]},x\in\mathcal{C}_a\}\big]\frac1p{\rm tr}(Q^{-1}\bar \Sigma)^2\nonumber\\
    &+O(p^{-\frac{1}{2}}).\label{eq:tr cov beta Sigma}
\end{align}
Using the shortcut notation \(\hat\sigma^2_k\equiv {\rm var}\{f_i\vert i>n_{[l]},x\in\mathcal{C}_k\}\)  for \(k\in\{1,2\}\), we get by substituting \eqref{eq:E beta trans Sigma E beta} and \eqref{eq:tr cov beta Sigma} into \eqref{eq:var fi} that
\begin{align}
    \hat\sigma^2_k=& \left[\gamma\rho_1\rho_2(2c_{[l]} + \hat mc_{[u]})\right]^2(\nu_2-\nu_1)^\trans Q^{-1} \Sigma_k Q^{-1} (\nu_2-\nu_1)\nonumber\\
    &+\gamma^2\big[\rho_1\rho_2(4c_{[l]}+\hat m^2c_{[u]})+c_{[u]}\sum_{a=1}^2\rho_a\hat \sigma^2_a\big]\frac1p{\rm tr}(Q^{-1}\bar \Sigma)^2+o_P(1).\label{eq:sigma_k}
\end{align}

{\BLUE Letting \(\xi\equiv c_{[u]}\gamma\), we get by multiplying the both sides of \eqref{eq:m} with \(c_{[u]}\) that
\begin{align*}
	c_{[u]} \hat m=\xi\rho_1\rho_2(2c_{[l]} + \hat mc_{[u]})(\nu_2-\nu_1)^\trans \left(I_p-\gamma c_{[u]}\bar \Sigma\right)^{-1}(\nu_2-\nu_1)+o_P(1).
\end{align*}
And multiplying the both sides of \eqref{eq:sigma_k} with \(c_{[u]}^2\) leads to
\begin{align}
	 c_{[u]}^2\hat\sigma^2_k=& \left[\rho_1\rho_2(2c_{[l]} + \hat mc_{[u]})\right]^2\xi^2(\nu_2-\nu_1)^\trans Q^{-1} \Sigma_k Q^{-1} (\nu_2-\nu_1)\nonumber\\
	&+\big[\rho_1\rho_2(4c_{[l]}+\hat m^2c_{[u]})+c_{[u]}\sum_{a=1}^2\rho_a\hat\sigma^2_a\big]\xi^2p^{-1}{\rm tr}(Q^{-1}\bar \Sigma)^2+o_P(1).\label{eq:sigma_k function of m and sigma}
\end{align} 
Set \(\hat\sigma^2=\sum_{a=1}^2\rho_a\hat\sigma^2_a\), we obtain
\begin{align*}
	 c_{[u]}^2\hat\sigma^2=& \left[\rho_1\rho_2(2c_{[l]} + \hat mc_{[u]})\right]^2\xi^2(\nu_2-\nu_1)^\trans Q^{-1} \bar\Sigma Q^{-1} (\nu_2-\nu_1)\nonumber\\
&+\big[\rho_1\rho_2(4c_{[l]}+\hat m^2c_{[u]})+c_{[u]}\hat\sigma^2\big]\xi^2p^{-1}{\rm tr}(Q^{-1}\bar \Sigma)^2+o_P(1).
\end{align*}
It is derived from the above equations that there exists a \(\xi\in\RR\) such that \((\hat m,\hat \sigma^2)=(m(\xi),\sigma^2(\xi))\) with \(m(\xi),\sigma^2(\xi)\) as given in  \eqref{eq:m(xi) general} and \eqref{eq:sigma2(xi) general}. Let us denote by \(\xi_e\) the value of \(\xi\) that allows us to access \(\hat m,\hat\sigma^2\) at some given value of the hyperparameter \(e>0\) (which, as we recall, was introduced in \eqref{eq:centered kernel ssl}). Notice that, as a direct consequence of \eqref{eq:var gigj} and \eqref{eq:f_i}, we have
\begin{align*}
	\cov\{f_i,f_j\}=O(p^{-1})
\end{align*}
for \(i,j>n_{[l]}\). With the same arguments, we get easily 
\begin{align*}
\cov\{f_i^2,f_j^2\}=O(p^{-1}),
\end{align*}
which entails 
\begin{align*}
	\frac{1}{n_{[u]}}\Vert f_{[u]}\Vert^2=\frac{1}{n_{[u]}}\sum_{i=n_{[l]}+1}^{n}f_i^2=\frac{1}{n_{[u]}}\sum_{i=n_{[l]}+1}^{n}\E\{f_i^2\}+O(p^{-\frac{1}{2}})=\rho_1\rho_2m^2+\sigma^2+O(p^{-\frac{1}{2}}).
\end{align*}
Therefore, the value \(\xi_e\) should satisfy, up to some asymptotically negligible terms, the equation
\begin{align*}
	\rho_1\rho_2m(\xi_e)^2+\sigma^2(\xi_e)=e^2.
\end{align*}
Note that the above equation does not give an unique \(\xi_e\) if \(\xi_e\) is allowed to take any value in \(\RR\). We need thus to further specify the admissible range of \(\xi_e\) as \(e\) goes from zero to infinity. We start by showing that \(m\) has always a positive value. With small adjustment to \eqref{eq:1trans fu}, we have
\begin{align*}
\frac{1}{n}\zeta^\trans f_{[u]}=c_0^{-1}\begin{bmatrix}1&0&0&0\end{bmatrix}K\begin{bmatrix}(\nu_2-\nu_1)^\trans\frac{1}{p}\Phi_{[l]}f_{[l]}\\2c_{[l]}\rho_1\rho_2\\0\\0\end{bmatrix}+O(p^{-\frac{1}{2}})
\end{align*}
with 
\begin{align*}
K=U^\trans RU+U^\trans RU(N^{-1}-U^\trans RU)^{-1}U^\trans RU.
\end{align*}
We recall  \(U^\trans RU\) is of the form \eqref{eq:form UtransRU}, and further remark that the matrix \(A\) in \eqref{eq:form UtransRU} is of the form \(A=\begin{bmatrix} a_{11}&0\\
0& a_{22}
\end{bmatrix}\) as we have \(U_{\cdot1}^\trans RU_{\cdot2}\) by applying \eqref{eq:aZuRb}. As indicated in Section~\ref{sec:proposition of the method}, for any \(e>0\), \(\alpha\) has a value greater than which is determined by \eqref{eq:relation alpha and e}. The matrix \(\left(\alpha I_{n_{[u]}}-\hat W_{[uu]}\right)^{-1}\)  is thus definite positive. Since
\begin{align*}
\left(\alpha I_{n_{[u]}}-\hat W_{[uu]}\right)^{-1}=&\left(\tilde \alpha I_{n_{[u]}}-\frac{1}{p}\hat{\Phi}_{[u]}^\trans\hat{\Phi}_{[u]}+\frac{r}{n}1_{n_{[u]}}1_{n_{[u]}}^\trans\right)^{-1}\\
=&R+RU(N^{-1}-U^\trans RU)^{-1}U^\trans R+O_{\Vert\cdot\Vert}(p^{-\frac{1}{2}}),
\end{align*} \(K\) is definite positive with high probability. Notice also that
\begin{align*}
K=U^\trans RU+U^\trans RU(N^{-1}-U^\trans RU)^{-1}U^\trans RU=\left[\left(U^\trans RU\right)^{-1}-N\right]^{-1},
\end{align*}
meaning that
\begin{align*}
	K_{12}=\frac{N_{12}}{{\rm det} \left\{\left(U^\trans RU\right)^{-1}-N\right\}}=\frac{1}{{\rm det} \left\{\left(U^\trans RU\right)^{-1}-N\right\}}.
\end{align*}
We get thus \(K_{12}>0\) since \({\rm det} \left\{\left(U^\trans RU\right)^{-1}-N\right\}={\rm det} \left\{K^{-1}\right\}>0\) due to the definite positiveness of \(K\), which implies that all the eigenvalues of \(K\) are positive. The fact that \(K\) is definite positive implies also \(K_{11}>0\), otherwise we would have \(\begin{bmatrix}
1&0
\end{bmatrix}K\begin{bmatrix}
1\\0
\end{bmatrix}=K_{11}\leq 0\). Since
\begin{align*}
	\frac{1}{n}\zeta^\trans f_{[u]}&=c_0^{-1}\left(K_{11}(\nu_2-\nu_1)^\trans\frac{1}{p}\Phi_{[l]}f_{[l]}+K_{12}2c_{[l]}\rho_1\rho_2\right)+O(p^{-\frac{1}{2}})\\
	&=2\rho_1\rho_2(K_{11}\Vert\nu_2-\nu_1\Vert^2+K_{12}ln_{[l]}/n)+O(p^{-\frac{1}{2}}),
\end{align*}
we get \(\frac{1}{n}\zeta^\trans f_{[u]}>0\) at large \(p\). As 
\begin{align*}
	\frac{1}{n}\zeta^\trans f_{[u]}=\frac{1}{n}\zeta^\trans \E\{f_{[u]}\}+O(p^{-\frac{1}{2}})=\rho_1\rho_2\hat \hat m+O(p^{-\frac{1}{2}})
\end{align*}
as a result of \(\cov\{f_i,f_j\}=O(p^{-1})\). We remark thus that \(\hat m>0\) holds asymptotically for any \(e>0\). Since \(\sigma^2>0\) by definition, we have necessarily \(\xi_e\in(0,\xi_{\sup})\) for any \(e\), as at least one of \(m(\xi_e),\sigma^2(\xi_e)\) is negative (or not well defined) outside this range. It can also be observed from the expressions  \eqref{eq:m(xi) general}--\eqref{eq:sigma2(xi) general} of \(m(\xi_e)\) and \(\sigma^2(\xi_e)\) that \(\rho_1\rho_2m^2(\xi)+\sigma^2(\xi)\) monotonously increases from zero to infinity as \(\xi\) increases from zero to \(\xi_{\sup}\). Therefore, \(\xi_e\in(0,\xi_{\sup})\) is uniquely given by
\begin{align*}
\rho_1\rho_2m(\xi_e)^2+\sigma^2(\xi_e)=e^2.
\end{align*}
In summary, for any \(e\in(0,+\infty)\), we have that \(\hat m=m(\xi_e)\), \(\hat \sigma^2=\sigma^2(\xi_e)\) with functions \(m(\xi),\sigma^2(\xi)\) as defined in  \eqref{eq:m(xi) general}--\eqref{eq:sigma2(xi) general} and \(\xi_e\in(0,\xi_{\sup})\) the unique solution of \(\rho_1\rho_2m(\xi_e)^2+\sigma^2(\xi_e)=e^2\); we get also from \eqref{eq:sigma_k function of m and sigma} the value of \(\hat \sigma_k^2\) as
\begin{align*}
	 \hat\sigma^2_k=& c_{[u]}^{-2}\left[\rho_1\rho_2(2c_{[l]} + m(\xi_e)c_{[u]})\right]^2\xi_e^2(\nu_2-\nu_1)^\trans Q^{-1} \Sigma_k Q^{-1} (\nu_2-\nu_1)\nonumber\\
	&+c_{[u]}^{-2}\big[\rho_1\rho_2(4c_{[l]}+ m(\xi_e)^2c_{[u]})+c_{[u]}\sum_{a=1}^2\rho_a\sigma^2(\xi_e)_a\big]\xi_e^2p^{-1}{\rm tr}(Q^{-1}\bar \Sigma)^2
\end{align*}
The proof of theorem~\ref{th:statistics of fu for centered method (generalized)} is thus concluded.

}

%\begin{align*}
%    \mathbb{E}\{\beta\}=2c_{[l]}\rho_1\rho_2\left[I_p-\xi\left(\sum_{a=1}^2\rho_a\Sigma_a+\rho_1\rho_2(\nu_2-\nu_1)(\nu_2-\nu_1)^\trans\right)\right](\nu_2-\nu_1)+O(p^{-\frac{1}{2}}),
%\end{align*}
%leading directly to
%\begin{align}
%    \theta=&\xi\rho_1\rho_2(\nu_2-\nu_1)^\trans\left[I_p-\xi\left(\sum_{a=1}^2\rho_a\Sigma_a+\rho_1\rho_2(\nu_2-\nu_1)(\nu_2-\nu_1)^\trans\right)\right](\nu_2-\nu_1)+o_P(1)\label{eq:theta}
%\end{align}
%where \(\theta=c_{[u]}m/2c_{[l]}\).
%Finally, the equations of Theorem~\ref{th:statistics of fu for centered method (generalized)} are retrieved from the two equation above with the notations of gathering \eqref{eq:m}, \eqref{eq:sigma_k} and \eqref{eq:theta} and by ignoring the vanishing terms. This completes the proof.

\section{Proof of Proposition~\ref{prop:spectral clustering}}
As the eigenvector of \(L_s\) associated with the smallest eigenvalue is \(D^{\frac{1}{2}}1_n\), we consider
\begin{align*}
L_s'=nD^{-\frac{1}{2}}WD^{-\frac{1}{2}}-n\frac{D^{\frac{1}{2}}1_n1_n^\trans D^{\frac{1}{2}}}{1_n^\trans D 1_n}.
\end{align*}
{\RED Note that \(\Vert L_s'\Vert =O(1)\) according to \citep[Theorem 1]{COU16}, and if \(v\) is an eigenvector of \(L_s\) associated with the eigenvalue \(u\), then it is also an eigenvector of \(L_s'\) associated with the eigenvalue \(-u+1\), except for the eigenvalue-eigenvector pair \((n, D^{\frac{1}{2}}1_n)\) of \(L_s\) turned into \((0, D^{\frac{1}{2}}1_n)\) for \(L_s'\). The second smallest eigenvector \(v_{{\rm Lap}}\) of \(L_s\) is the same as the largest eigenvector of \(L_s'\).

{\RED From the random matrix equivalent of \(L_s'\) given by  \citet[Theorem 1]{COU16} and  that of \(\hat W\) expressed in \eqref{eq:hat W}}, we have
\begin{align*}
    \hat W=h(\tau)L_s'+\frac{5h'(\tau)^2}{4}\psi\psi^\trans+O(p^{-\frac{1}{2}})
\end{align*}
where \(\psi=[\psi_1,\ldots,\psi_n]^\trans\) with \(\psi_i=\Vert x_i\Vert^2-\mathbb{E}[\Vert x_i\Vert^2]\).

%For \(k\in\{1,2\}\), define \(j_k\in\mathbb{R}^n\) the indicator vector of class \(k\) with \([j_k]_i=1\) if \(x_i\in\mathcal{C}_k\), otherwise \([j_k]_i=0\). Then, we have
%\begin{align*}
%    &d_{{\rm inter}}(v)=\vert j_1^\trans v/n_1-j_2^\trans v/n_2\vert\\
%    &d_{{\rm intra}}(v)=\Vert v-(j_1^\trans v/n_1)j_1-(j_2^\trans v/n_2)j_2 \Vert/\sqrt{n}
%\end{align*}
%for some \(v\in\mathbb{R}^n\).  

Recall that
\begin{align*}
&d_{{\rm inter}}(v)=\vert j_1^\trans v/n_1-j_2^\trans v/n_2\vert\\
&d_{{\rm intra}}(v)=\Vert v-(j_1^\trans v/n_1)j_1-(j_2^\trans v/n_2)j_2 \Vert/\sqrt{n}
\end{align*}
for some \(v\in\mathbb{R}^n\), and \(j_k\in\mathbb{R}^n\) with \(k\in\{1,2\}\) the indicator vector of class \(k\) with \([j_k]_i=1\) if \(x_i\in\mathcal{C}_k\), otherwise \([j_k]_i=0\). 

Denote by \(\lambda_{{\rm Lap}}\) the eigenvalue of \(h(\tau)L_s'\)  associated with \(v_{{\rm Lap}}\), and \(\lambda_{{\rm ctr}}\) the eigenvalue of \(\hat W\)  associated with \(v_{{\rm ctr}}\). {\RED Under the condition of non-trivial clustering upon \(v_{{\rm Lap}}\) with \(d_{{\rm inter}}(v_{{\rm Lap}})/d_{{\rm intra}}(v_{{\rm Lap}})=O(1)\), we have \(j_k^\trans v_{{\rm Lap}}/\sqrt{n_k}=O(1)\) from the above expressions of \(d_{{\rm inter}}(v)\) and \(d_{{\rm intra}}(v)\). The fact that \(j_k^\trans v_{{\rm Lap}}/\sqrt{n_k}=O(1)\) implies that the eigenvalue \(\lambda_{{\rm Lap}}\) of \(h(\tau)L_s'\) remains at a non vanishing distance from other eigenvalues of \(h(\tau)L_s'\) \citep[Theorem 4]{COU16}. %, in the sense that the distances between \(\lambda_{{\rm Lap}}\) and other eigenvalues of \(h(\tau)L_s'\) are of \(O(1)\).
The same can be said about \(\hat W\) and its eigenvalue \(\lambda_{{\rm ctr}}\).}

Let \(\gamma\) be a positively oriented complex closed path circling only around \(\lambda_{{\rm Lap}}\) and \(\lambda_{{\rm ctr}}\). {\RED Since there can be only one eigenvector of \(L_s'\) (\(\hat W\), resp.) whose limiting scalar product with \(j_k\) for \(k\in\{1,2\}\) is bounded away from zero \citep[Theorem 4]{COU16}, which is \(v_{{\rm Lap}}\) (resp., \(v_{{\rm ctr}}\)), we have, by Cauchy's formula \citep[Theorem 10.15]{walter1987real},} 
\begin{align*}
    &\frac{1}{n_k}(j_k^\trans v_{{\rm Lap}})^2=-\frac{1}{2\pi i}\oint_\gamma  \frac{1}{n_k}j_k^\trans(h(\tau)L_s'-zI_n)^{-1}j_kdz+o_P(1)\\
    &\frac{1}{n_k}(j_k^\trans v_{{\rm ctr}})^2=-\frac{1}{2\pi i}\oint_\gamma  \frac{1}{n_k}j_k^\trans(\hat W-zI_n)^{-1}j_kdz+o_P(1)
\end{align*}
for \(k\in\{1,2\}\). Since $\hat W$ is a low-rank perturbation of $\hat L$, invoking Sherman-Morrison's formula {\RED \citep{sherman1950adjustment}}, we further have
\begin{align*}
    j_k^\trans(\hat W-zI_n)^{-1}j_k= j_k^\trans(h(\tau)L_s'-zI_n)^{-1}j_k-\frac{(5h'(\tau)^2/4)\left(j_k^\trans(h(\tau)L_s'-zI_n)^{-1}\psi\right)^2}{1+(5h'(\tau)^2/4)\psi^\trans(h(\tau)L_s'-zI_n)^{-1}\psi}+o_P(n_k).
\end{align*}
As \(\frac1{\sqrt{n_k}}j_k^\trans(h(\tau)L_s'-zI_n)^{-1}\psi=o_P(1)\) \citep[Equation 7.6]{COU16}, we get
\begin{align*}
 \frac1{n_k}j_k^\trans(\hat W-zI_n)^{-1}j_k=\frac1{n_k} j_k^\trans(h(\tau)L_s'-zI_n)^{-1}j_k+o_P(1),
\end{align*}
and thus
\begin{align*}
    \frac{1}{n_k}(j_k^\trans v_{{\rm Lap}})^2=\frac{1}{n_k}(j_k^\trans v_{{\rm ctr}})^2+o_P(1),
\end{align*}
which concludes the proof of Proposition~\ref{prop:spectral clustering}.

{\RED

\section{Asymptotic Matrix Equivalent for \(\hat W\)}
\label{app: hat W}
The objective of this section is to prove the asymptotic matrix equivalent for \(\hat W\) expressed in \eqref{eq:hat W}. Some additional notations that will be useful in the proof:
\begin{itemize}
	\item for $x_{i}\in\mathcal{C}_{k}$, \(k\in\{1,2\}\), $\theta_{i}\equiv x_{i}-\mu_{k}$, and $\theta\equiv [\theta_{1},\cdots,\theta_{n}]^\trans$;
	\item \(\mu^{\circ}_k=\mu_k-\frac{1}{n}\sum_{k'=1}^2n_{k'}\mu_{k'}\), \(t_k=\left({\rm tr}C_k-\frac{1}{n}\sum_{k'=1}^2n_{k'}{\rm tr}C_{k'}\right)/\sqrt{p}\);
	\item $j_{k}\in\mathbb{R}^{n}$ is the canonical vector of $\mathcal{C}_{k}$, i.e., $[j_k]_i=1$ if $x_{i}\in\mathcal{C}_{k}$ and $[j_k]_i=0$ otherwise;
	\item $\psi_{i}\equiv\left(\Vert \theta_{i} \Vert^{2}-{\rm E}[\Vert \theta_{i} \Vert^{2}\right)/\sqrt{p}$, $\psi\equiv [\psi_{1},\cdots,\psi_{n}]^\trans$ and $(\psi)^{2}\equiv [(\psi_{1})^{2},\cdots,(\psi_{n})^{2}]^\trans$.
\end{itemize}
As \(w_{ij}=h(\Vert x_i-x_j\Vert^2/p=h(\tau)+O(p^{-\frac{1}{2}})\) for all $i\neq j$, we can Taylor-expand $w_{ij}=h(\| x_{i}-x_{j}\|^2/p$ around $h(\tau)$ to obtain the following expansion for $W$, which can be found in \citep{COU16}:
\begin{align*}
    W &= h(\tau)1_{n}1_{n}^\trans+\frac{h'(\tau)}{\sqrt{p}}\left[\psi 1_{n}^\trans+1_{n}\psi^\trans+\sum_{b=1}^{2}t_{b}j_{b}1_{n}^\trans+1_{n}\sum_{a=1}^{2}t_{a}j_{a}^\trans\right]\\
     &+\frac{h'(\tau)}{p}\Bigg[\sum_{a,b=1}^{2}\|\muctra-\muctrb\|^2j_{b}j_{a}^\trans-2\theta \sum_{a=1}^{2}\muctra j_{a}^\trans+2\sum_{b=1}^{2}{\rm diag}(j_{b})\theta \muctrb 1_n^\trans\nonumber\\
	  &-2\sum_{b=1}^{2}j_{b}\mu_{b}^{\circ \trans}\theta^\trans+21_{n}\sum_{a=1}^{2}{\muctra}^\trans\theta^\trans {\rm diag}(j_{a})-2\theta\theta^\trans\Bigg]\nonumber\\
	&+\frac{h''(\tau)}{2p}\bigg[(\psi)^21_{n}^\trans+1_{n}[(\psi)^2]^\trans+\sum_{b=1}^{2}t_b^2j_{b} 1_n^\trans+1_{n}\sum_{a=1}^{2}t_a^2j_{a}^\trans \nonumber\\
	&+2\sum_{a,b=1}^{2}t_{a}t_{b}j_{b}j_{a}^\trans+2\sum_{b=1}^{2}{\rm diag}(j_{b})t_{b}\psi 1_{n}^\trans+2\sum_{b=1}^{2}t_{b}j_{b}\psi^\trans+2\sum_{a=1}^{2}1_{n}\psi^\trans {\rm diag}(j_{a})t_{a}\nonumber\\
    &+2\psi \sum_{a=1}^{2}t_{a}j_{a}^\trans+2\psi\psi^\trans \bigg]+(h(0)-h(\tau)+\tau h'(\tau))I_{n}+O_{\Vert \cdot\Vert}(p^{-\frac{1}{2}}).
\end{align*}
Applying \(P_n=\left(I_n-\frac{1}{n}1_n1_n^\trans\right)\) on both sides of the above equation, we get
\begin{align*}
    \hat W&= P_nWP_n\\
    &=\frac{-2h'(\tau)}{p}\Bigg[\sum_{a,b=1}^{2}(\mu_{a}^{\circ \trans} \muctrb) j_{b}j_{a}^\trans+P_n\theta \sum_{a=1}^{2}\muctra j_{a}^\trans+\sum_{b=1}^{2}j_{b}\mu_{b}^{\circ \trans}\theta^\trans P_n+P_n\theta\theta^\trans P_n\Bigg]\\
    &+\frac{h''(\tau)}{p}\Bigg[\sum_{a,b=1}^{2}t_{a}t_{b}j_{b}j_{a}^\trans+\sum_{b=1}^{2}t_{b}j_{b}\psi^\trans P_n+P_n\psi \sum_{a=1}^{2}t_{a}j_{a}^\trans+P_n\psi\psi^\trans P_n\Bigg]\\
    &+(h(0)-h'(\tau)+\tau h''(\tau))P_n+O(p^{-\frac{1}{2}})\\
    &=\frac{1}{p}\hat \Phi^\trans\hat \Phi+(h(0)-h(\tau)+\tau h'(\tau))P_n+O_{\Vert \cdot\Vert}(p^{-\frac{1}{2}})
\end{align*}
where the last equality is justified by 
\begin{align*}
    \frac{1}{p}\hat \Phi^\trans\hat \Phi&=\frac{-2h'(\tau)}{p}\Bigg[\sum_{a,b=1}^{2}(\mu_{a}^{\circ \trans} \muctrb) j_{b}j_{a}^\trans+P_n\theta \sum_{a=1}^{2}\muctra j_{a}^\trans+\sum_{b=1}^{2}j_{b}\mu_{b}^{\circ \trans}\theta^\trans P_n+P_n\theta\theta^\trans P_n\Bigg]\\
    &+\frac{h''(\tau)}{p}\Bigg[\sum_{a,b=1}^{2}t_{a}t_{b}j_{b}j_{a}^\trans+\sum_{b=1}^{2}t_{b}j_{b}\psi^\trans P_n+P_n\psi \sum_{a=1}^{2}t_{a}j_{a}^\trans+P_n\psi\psi^\trans P_n\Bigg].
\end{align*}
Equation~\eqref{eq:hat W} is thus proved.
}

\section{Guarantee for approaching the optimal performance on isotropic Gaussian data}
\label{app:optimal performance on isotropic data}
{\BLUE
	
The purpose of this section is to provide some general guarantee for the proposed centered regularization method to approach the best achievable performance on isotropic high dimensional Gaussian data, which was characterized in the recent work of \citet{lelarge2019asymptotic}. In this work, the considered isotropic data model is a special case of our analytical framework, in which \(-\mu_1=\mu_2=\mu\), \(C_1=C_2=I_p\)  and \(\rho_1=\rho_2\). Reorganizing the results of \citep{lelarge2019asymptotic}, the optimally achievable classification accuracy in the limit of large \(p\) is equal to
\begin{align*}
1-Q(\sqrt{q_*})
\end{align*}
with \(q_*>0\) satisfying the fixed point equation
\begin{align}
%	q_2^*=\frac{n_{[l]}}{n}+\frac{n_{[u]}\mathbb{E}\{{\rm tanh}( f^*)\}}{n}\\
%	1/(1+q_1^*)=\frac{n_{[l]}+n_{[u]}\mathbb{E}\{{\rm tanh}( f^*)\}}{p+n_{[l]}+n_{[u]}\mathbb{E}\{{\rm tanh}( f^*)}
q_*=\Vert\mu\Vert^2-\frac{p\Vert\mu\Vert^2}{p+\Vert\mu\Vert^2\left(n_{[l]}+\mathbb{E}_{z\sim\mathcal{N}(q_*,q_*)}\{{\rm tanh}(z)\}n_{[u]}\right)}.\label{eq:q optimal}
\end{align}
It is easy to see that the optimal accuracy is higher with greater \(q_*\).
%\begin{align*}
%	\mathbb{P}\{s_*>0\}\text{ for } s_*\sim\mathcal{N}\left(\frac{\Vert\mu\Vert^2}{1+q_*},\frac{\Vert\mu\Vert^2}{1+q_*}\right)
%\end{align*}
%with \(q_*>0\) satisfying the fixed point equation
%\begin{align}
%%	q_2^*=\frac{n_{[l]}}{n}+\frac{n_{[u]}\mathbb{E}\{{\rm tanh}( f^*)\}}{n}\\
%%	1/(1+q_1^*)=\frac{n_{[l]}+n_{[u]}\mathbb{E}\{{\rm tanh}( f^*)\}}{p+n_{[l]}+n_{[u]}\mathbb{E}\{{\rm tanh}( f^*)}
%	q_*=\frac{p}{\Vert\mu\Vert^2\left(n_{[l]}+\mathbb{E}\{{\rm tanh}( s_*)\}n_{[u]}\right)}.\label{eq:q optimal}
%\end{align}
In parallel, reformulating the results of Corollary~\ref{cor: r_ctr} for some value of the hyperparameter \(e>0\) such that
\begin{align}
\label{eq:condition e}
	m(\xi_e)=\frac{m(\xi_e)^2}{m(\xi_e)^2+\sigma^2(\xi_e)},
\end{align}
the high dimensional classification accuracy achieved by the centered regularization method is asymptotically equal to
\begin{align*}
1-Q(\sqrt{q_c})
\end{align*}
with \(q_c>0\) satisfying the fixed point equation
\begin{align}
q_c=\Vert\mu\Vert^2-\frac{p\Vert\mu\Vert^2}{p+\Vert\mu\Vert^2\left(n_{[l]}+\frac{q_c}{q_c+1}n_{[u]}\right)}\label{eq:q centered}
\end{align}

%\begin{align*}
%\mathbb{P}\{s_c>0\}\text{ for } s_c\sim\frac{1+q_c}{1+q_c+\Vert\mu\Vert^2}\mathcal{N}\left(\frac{\Vert\mu\Vert^2}{1+q_c},\frac{\Vert\mu\Vert^2}{1+q_c}\right)
%\end{align*}
%with \(q_c>0\) satisfying the fixed point equation
%\begin{align}
%q_c=\frac{p}{\Vert\mu\Vert^2\left(n_{[l]}+\frac{\mathbb{E}\{s_c\}^2}{\mathbb{E}\{s_c^2\}} n_{[u]}\right)}.\label{eq:q centered}
%\end{align}

Obviously, the fixed-point equations \eqref{eq:q optimal}--\eqref{eq:q centered} are identical at \(n_{[u]}=0\), meaning that the centered regularization method achieves the optimal performance on fully labelled sets. For partially labelled sets,  the difference between \eqref{eq:q optimal} and \eqref{eq:q centered} resides in the multiplying factors before \(n_{[u]}\). This means that, for a best achievable accuracy of \(1-Q(\sqrt{q_*})\) at some \(n_{[l]}\) and \(n_{[u]}\), the centered regularization method achieves, with the hyperparameter \(e\) set to satisfy \eqref{eq:condition e}, the same level of accuracy with the same amount of labelled samples and \(g(q_*)n_{[u]}\) unlabelled ones where \(g(q_*)=\mathbb{E}_{z\sim\mathcal{N}(q_*,q_*)}\{{\rm tanh}(z)\}(q_*+1)/q_*\).
The ratio function
\begin{align*}
	g(q)=\frac{\mathbb{E}_{z\sim\mathcal{N}(q,q)}\{{\rm tanh}(z)\}(q+1)}{q}
\end{align*}
 is plotted in Figure~\ref{fig:g(q)}. We remark also that \(\lim_{q\to 0^+} g(q)=1\) and \(\lim_{q\to +\infty}g(q)=0\). Although the value of \(g(q)\) can get up to \(1.168\), the number of unlabelled samples required to reach the optimal performance can be reduced with an optimally chosen \(e\) (which generally does not satisfy \eqref{eq:condition e}). In fact, even with the same numbers of labelled and unlabelled data, the performance of centered regularization method at an optimally set \(e\) is often very close to the best achievable one, as shown in Figures \ref{fig:optimal performance on isotropic data}--\ref{fig: graph-based SSL algorithms on isotropic Gaussian mixture data}.}

\begin{figure}
	\centering
	\begin{tikzpicture}[font=\footnotesize]
	\renewcommand{\axisdefaulttryminticks}{4} 
	\tikzstyle{every major grid}+=[style=densely dashed ] \tikzstyle{every axis y label}+=[yshift=-10pt] 
	\tikzstyle{every axis x label}+=[yshift=5pt]
	\tikzstyle{every axis legend}+=[cells={anchor=west},fill=white,
	at={(1,0.2)}, anchor=north east, font=\small]
	\begin{axis}[
	%ybar,
	width=.65\linewidth,
	height=.4\linewidth,
	%xmode=log,
	%log basis x={10},
	grid=major,
	ymajorgrids=false,
	scaled ticks=true,
	ylabel={\(g(q)\)},
	xlabel={\(q\)},
	]
	\addplot[only marks,mark=*, blue, mark size=1pt] plot coordinates{
(0.2,1.015)(0.4,1.038)(0.6,1.062)(0.8,1.083)(1.0,1.101)(1.2,1.116)(1.4,1.129)(1.6,1.139)(1.8,1.147)(2.0,1.153)(2.2,1.158)(2.4,1.162)(2.6,1.165)(2.8,1.167)(3.0,1.168)(3.2,1.168)(3.4,1.168)(3.6,1.167)(3.8,1.166)(4.0,1.164)(4.2,1.163)(4.4,1.161)(4.6,1.158)(4.8,1.156)(5.0,1.154)(5.2,1.151)(5.4,1.149)(5.6,1.146)(5.8,1.144)(6.0,1.141)(6.2,1.139)(6.4,1.136)(6.6,1.134)(6.8,1.131)(7.0,1.129)(7.2,1.126)(7.4,1.124)(7.6,1.121)(7.8,1.119)(8.0,1.117)
	};

	\end{axis}
	\end{tikzpicture} 
	\caption{Values of \(g(q)\) at various \(q\). }
	\label{fig:g(q)}
\end{figure}
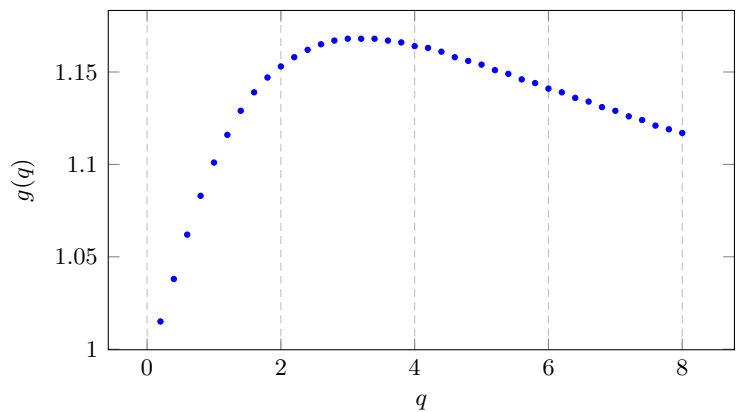

\bibliography{biblio}

\end{document}